\documentclass[11pt]{article}

\usepackage[final]{acl}

\usepackage{times}
\usepackage{latexsym}

\usepackage[T1]{fontenc}

\usepackage[utf8]{inputenc}

\usepackage{microtype}

\usepackage{inconsolata}

\usepackage{graphicx}

\usepackage{hyperref}
\usepackage{url}
\usepackage{comment}
\usepackage{amsmath}
\usepackage{amssymb}
\usepackage{multirow}
\usepackage{multicol}
\usepackage{booktabs}
\usepackage{adjustbox}
\usepackage{xcolor}
\usepackage[table]{xcolor}
\usepackage{enumitem}
\usepackage{pifont}

\definecolor{Mycolor1}{HTML}{EF9A9A}
\definecolor{Mycolor2}{HTML}{90CAF9}
\definecolor{Mycolor3}{HTML}{AED581}
\definecolor{mygray}{gray}{0.9}

\title{Two Pathways to Truthfulness: On the Intrinsic \\ Encoding of LLM Hallucinations}

\author{Wen Luo\textsuperscript{\rm $\heartsuit$},
Guangyue Peng\textsuperscript{\rm $\heartsuit$},
Wei Li\textsuperscript{\rm $\heartsuit$},
Shaohang Wei\textsuperscript{\rm $\heartsuit$},
Feifan Song\textsuperscript{\rm $\heartsuit$},\\
\textbf{Liang Wang\textsuperscript{\ding{171}},
Nan Yang\textsuperscript{\ding{171}},
Xingxing Zhang\textsuperscript{\ding{171}},
Jing Jin\textsuperscript{\rm $\heartsuit$},
Furu Wei\textsuperscript{\ding{171}},
Houfeng Wang\textsuperscript{\rm $\heartsuit$}}\thanks{Corresponding author}\\
\textsuperscript{\rm $\heartsuit$} State Key Laboratory of Multimedia Information Processing,\\
School of Computer Science, Peking University\\
\textsuperscript{\ding{171}} Microsoft Research Asia\\
\texttt{rowan.wenluo@gmail.com, wanghf@pku.edu.cn}
}

\begin{document}
\maketitle
\begin{abstract}
Despite their impressive capabilities, large language models (LLMs) frequently generate hallucinations. Previous work shows that their internal states encode rich signals of truthfulness, yet the origins and mechanisms of these signals remain unclear.
In this paper, we demonstrate that truthfulness cues arise from two distinct information pathways: (1) a Question-Anchored pathway that depends on question-answer information flow, and (2) an Answer-Anchored pathway that derives self-contained evidence from the generated answer itself.
First, we validate and disentangle these pathways through attention knockout and token patching.
Afterwards, we uncover notable and intriguing properties of these two mechanisms.
Further experiments reveal that (1) the two mechanisms are closely associated with LLM knowledge boundaries; and (2) internal representations are aware of their distinctions.
Finally, building on these insightful findings, two applications are proposed to enhance hallucination detection performance.
Overall, our work provides new insight into how LLMs internally encode truthfulness, offering directions for more reliable and self-aware generative systems.\footnote{Code is publicly available at \url{https://github.com/RowanWenLuo/llm-truthfulness-pathways}}
\end{abstract}

\section{Introduction} 

Despite their remarkable capabilities in natural language understanding and generation, large language models (LLMs) often produce \emph{hallucinations}—outputs that appear plausible but are factually incorrect. This phenomenon poses a critical challenge for deploying LLMs in real-world applications where reliability and trustworthiness are paramount \citep{DBLP:conf/acl/Shi00GRCR24,DBLP:conf/acl/BaiLBHLZLSG0O24}.
One line of research tackles hallucination detection from an extrinsic perspective \citep{DBLP:conf/emnlp/MinKLLYKIZH23,DBLP:conf/acl/HuZJZW025,DBLP:conf/acl/HuangFMFFGYZZWW25}, evaluating only the model's outputs while disregarding its internal dynamics.
Although such approaches can identify surface-level textual inconsistencies, their extrinsic focus limits the insight they offer into the underlying causes of hallucinations.
Complementing these efforts, another line of work investigates the intrinsic properties of LLMs, revealing that their internal representations encode rich truthfulness signals \citep{DBLP:conf/iclr/BurnsYKS23,DBLP:conf/nips/0002PVPW23,DBLP:conf/iclr/0026L0GWTFY24,DBLP:conf/iclr/OrgadTGRSKB25,niu2025robust}. 
These internal truthfulness signals can be exploited to detect an LLM's own generative hallucinations by training a linear classifier (i.e., a probe) on its hidden representations. 
However, while prior work establishes the presence of such cues, the mechanisms by which they arise and operate remain largely unexplored.
Recent studies indicate well-established mechanisms in LLMs that underpin complex capabilities such as in-context learning \citep{wang2023label}, long-context retrieval \citep{DBLP:conf/iclr/WuWX0F25}, and reasoning \citep{qian2025demystifying}.
This observation naturally leads to a key question: \emph{how do truthfulness cues arise and function within LLMs?}

In this paper, we uncover that truthfulness signals in LLMs arise from \textbf{two distinct information pathways}:  
(1) a \textbf{Question-Anchored (Q-Anchored) pathway}, which depends on the flow of information from the input question to the generated answer, and  
(2) an \textbf{Answer-Anchored (A-Anchored) pathway}, which derives self-contained evidence directly from the model’s own outputs.  
We begin with a preliminary study using saliency analysis to quantify information flow potentially relevant to hallucination detection. Results reveal a bimodal distribution of dependency on question-answer interactions, suggesting heterogeneous truthfulness encoding mechanisms.  
To validate this hypothesis, we design two experiments across 4 diverse datasets using 12 models that vary in both architecture and scale, including base, instruction-tuned, and reasoning-oriented models.
By (i) blocking critical question-answer information flow through attention knockout \citep{DBLP:conf/emnlp/GevaBFG23,DBLP:conf/acl/FierroFES25} and (ii) injecting hallucinatory cues into questions via token patching \citep{DBLP:conf/icml/GhandehariounCP24,DBLP:conf/iclr/ToddLSMWB24}, we disentangle these truthfulness pathways. Our analyses confirm that Q-Anchored signals rely heavily on question-derived cues, whereas A-Anchored signals are robust to their removal and primarily originate from the generated answer itself. 

Building on this foundation, we further investigate emergent properties of these truthfulness pathways through large-scale experiments. Our findings highlight two intriguing characteristics:  
\textbf{(1) Association with knowledge boundaries:} Q-anchored encoding predominates for well-established facts that fall within the knowledge boundary, whereas A-anchored encoding is favored in long-tail cases.  
\textbf{(2) Self-awareness:} LLM internal states can distinguish which mechanism is being employed, suggesting intrinsic awareness of pathway distinctions.  

Finally, these analyses not only deepen our mechanistic understanding of hallucinations but also enable practical applications. Specifically, by leveraging the fundamentally different dependencies of the truthfulness pathways and the model’s intrinsic awareness, we propose two pathway-aware strategies to enhance hallucination detection.
\textbf{(1) Mixture-of-Probes (MoP):} Motivated by the specialization of internal pathways, MoP employs a set of expert probing classifiers, each tailored to capture distinct truthfulness encoding mechanisms.
\textbf{(2) Pathway Reweighting (PR):} From the perspective of selectively emphasizing pathway-relevant internal cues, PR modulates information intensity to amplify signals that are most informative for hallucination detection, aligning internal activations with pathway-specific evidence.
Experiments demonstrate that our proposed methods consistently outperform competing approaches, achieving up to a 10\% AUC gain across various datasets and models.

Overall, our key contributions are summarized as follows:
\begin{itemize}[left=0cm]
\item \textbf{(Mechanism)} We conduct a systematic investigation into how internal truthfulness signals emerge and operate within LLMs, revealing two distinct information pathways: a \emph{Question-Anchored} pathway that relies on question-answer information flow, and an \emph{Answer-Anchored} pathway that derives self-contained evidence from the generated output.
\item \textbf{(Discovery)} Through large-scale experiments across multiple datasets and model families, we identify two key properties of these mechanisms: (i) association with knowledge boundaries, and (ii) intrinsic self-awareness of pathway distinctions.
\item \textbf{(Application)} Building on these findings, we propose two pathway-aware detection methods that exploit the complementary nature of the two mechanisms to enhance hallucination detection, providing new insights for building more reliable generative systems.
\end{itemize}

\section{Background}

\subsection{Hallucination Detection}
 
Given an LLM $f$, we denote the dataset as $D=\{(q_i, \hat{y}^{f}_{i}, z^{f}_{i})\}_{i=1}^{N}$,  
where $q_i$ is the question, $\hat{y}^{f}_{i}$ the model’s answer in open-ended generation,  
and $z^{f}_{i}\in\{0,1\}$ indicates whether the answer is hallucinatory.  
The task is to predict $z^{f}_{i}$ given the input $x^f_i=[q_i,\hat{y}^{f}_{i}]$ for each instance.  
Cases in which the model refuses to answer are excluded, as they are not genuine hallucinations and can be trivially classified.  
Methods based on internal signals assume access to the model’s hidden representations but no external resources (e.g., retrieval systems or fact-checking APIs) \citep{xue-etal-2025-ualign}.
Within this paradigm, probing trains a lightweight linear classifier on hidden activations to discriminate between hallucinatory and factual outputs, and has been shown to be among the most effective approaches in this class of internal-signal-based methods \citep{DBLP:conf/iclr/OrgadTGRSKB25}.

\subsection{Exact Question and Answer Tokens}

To analyze the origins and mechanisms of truthfulness signals in LLMs, we primarily focus on \textbf{exact tokens} in question-answer pairs.  
Not all tokens contribute equally to detecting factual errors: some carry core information essential to the meaning of the question or answer, while others provide peripheral details.  
We draw on \emph{semantic frame} theory \citep{baker1998berkeley,pagnoni2021understanding}, which represents a situation or event along with its participants and their roles. In the theory, frame elements are categorized as:  
(1) \emph{Core frame elements}, which define the situation itself, and  
(2) \emph{Non-core elements}, which provide additional, non-essential context.  

As shown in Table \ref{tab:exact_tokens_example}, we define:  
(1) \textbf{Exact question tokens:} core frame elements in the question, typically including \textbf{the exact subject and property tokens} (i.e., \textit{South Carolina} and \textit{capital}). 
(2) \textbf{Exact answer tokens:} core frame elements in the answer that convey the critical information required to respond correctly (i.e., \textit{Columbia}).
Humans tend to rely more on core elements when detecting errors, as these tokens carry the most precise information. Consistent with this intuition, recent work \citep{DBLP:conf/iclr/OrgadTGRSKB25} shows that probing activations on the exact answer tokens offers the strongest signal for hallucination detection, outperforming all other token choices.
Motivated by these findings, our analysis mainly centers on exact tokens to probe truthfulness signals in LLMs. Moreover, to validate the robustness of our conclusions, we also conduct comprehensive experiments using alternative, non-exact-token configurations (see Appendix \ref{sec:appendix Probing Implementation Details}).

\begin{table}[!htb]
\centering
\small
\begin{tabular}{p{0.92\columnwidth}}
\toprule
\textbf{Question:} What is the \colorbox{Mycolor1!80}{capital} of \colorbox{Mycolor2!80}{South Carolina}? \\
\textbf{Answer:} It is \colorbox{Mycolor3!80}{Columbia}, a hub for government, culture, and education that houses the South Carolina State House and the University of South Carolina. \\
\bottomrule
\end{tabular}
\caption{Example of exact question and answer tokens. Colors indicate token types: 
\textcolor{Mycolor1}{$\vcenter{\hbox{\rule{0.7em}{1.4ex}}}$}  - exact property, 
\textcolor{Mycolor2}{$\vcenter{\hbox{\rule{0.7em}{1.4ex}}}$}  - exact subject, and 
\textcolor{Mycolor3}{$\vcenter{\hbox{\rule{0.7em}{1.4ex}}}$}  - exact answer tokens.}
\label{tab:exact_tokens_example}
\end{table}

\section{Two Internal Truthfulness Pathways}

We begin with a preliminary analysis using metrics based on saliency scores (\S \ref{sec:preliminary}).
The quantitative results reveal \textbf{two distinct information pathways for truthfulness encoding:}
(1) a \textbf{Question-Anchored (Q-Anchored) Pathway}, which relies heavily on exact question tokens (i.e., the questions), and
(2) an \textbf{Answer-Anchored (A-Anchored) Pathway}, in which the truthfulness signal is largely independent of the question-to-answer information flow. 
Section \ref{sec:Disentangling Information Mechanisms} presents experiments validating this hypothesis. In particular, we show that Q-Anchored Pathway depends critically on information flowing from the question to the answer, whereas the signals along the A-Anchored Pathway are primarily derived from the LLM-generated answer itself.

\subsection{Saliency-Driven Preliminary Study}
\label{sec:preliminary}

This section investigates the intrinsic characteristics of LLM attention interactions and their potential role in truthfulness encoding.
We employ saliency analysis \citep{DBLP:journals/corr/SimonyanVZ13}, a widely used interpretability method, to reveal how attention among tokens influences probe decisions.
Following common practice \citep{michel2019sixteen,wang2023label}, we compute the saliency score as:
\begin{equation}
S^l(i,j) = \left| A^l(i,j) \frac{\partial \mathcal{L}(x)}{\partial A^l(i,j)} \right|,
\end{equation}
where $S^l$ denotes the saliency score matrix of the $l$-th layer, $A^l$ represents the attention weights of that layer, and $\mathcal{L}$ is the loss function for hallucination detection (i.e., the binary cross-entropy loss).
Scores are averaged over all attention heads within each layer.
In particular, $S^l(i,j)$ quantifies the saliency of attention from query $i$ to key $j$, capturing how strongly the information flow from $j$ to $i$ contributes to the detection.
We study two types of information flow:  
(1) $S_{E_Q \rightarrow E_A}$, the saliency of direct information flow from the exact question tokens to the exact answer tokens, and  
(2) $S_{E_Q \rightarrow *}$, the saliency of the total information disseminated by the exact question tokens. 

\paragraph{Results}

\begin{figure}[!htb]
\centering
\includegraphics[width=\columnwidth]{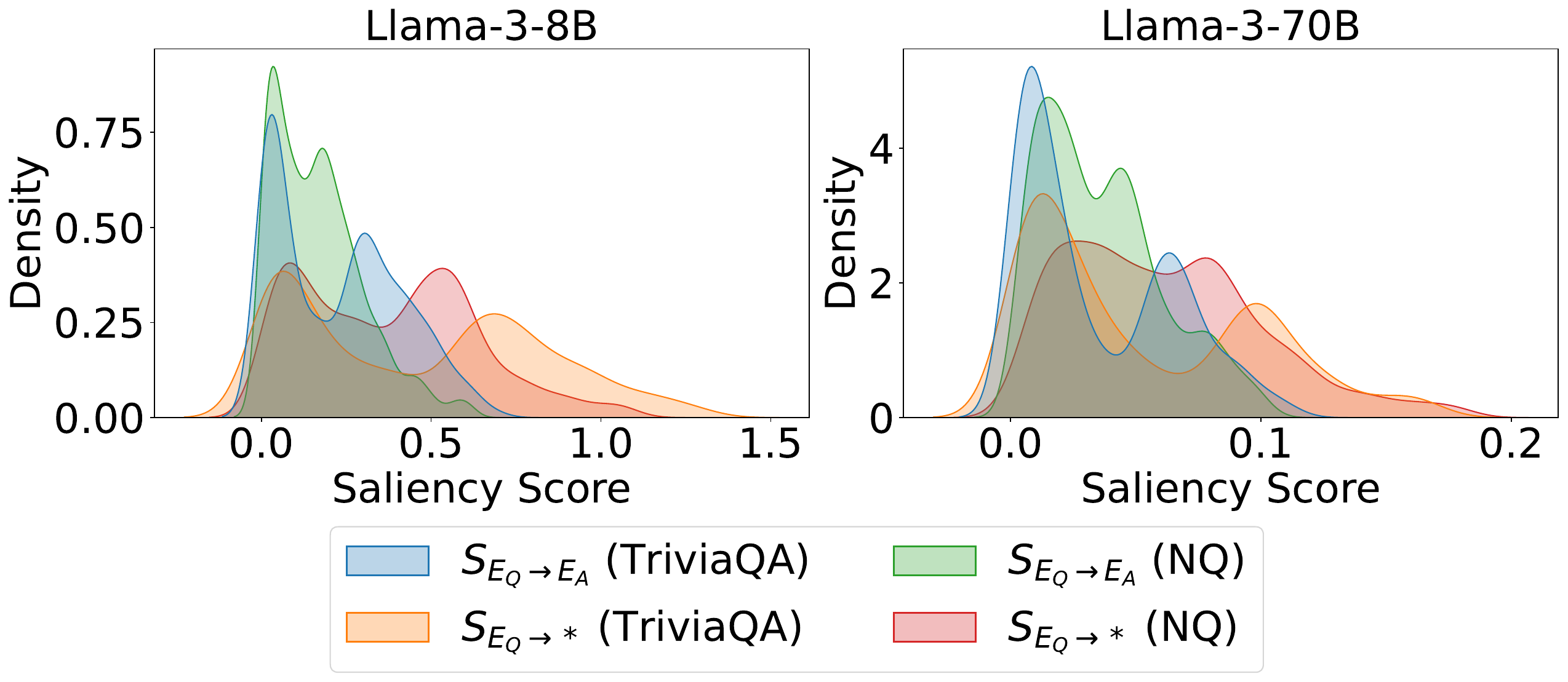}
\caption{Kernel density estimates of saliency‐score distributions for critical question-to-answer information flows. The bimodal pattern suggests two distinct information mechanisms.}
\label{fig:preliminary}
\end{figure}

We demonstrate Kernel Density Estimation results of the saliency scores on TriviaQA \citep{joshi-etal-2017-triviaqa} and Natural Questions \citep{kwiatkowski-etal-2019-natural} datasets.
As shown in Figure \ref{fig:preliminary}, probability densities reveal a clear bimodal distribution: for all examined information types originating from the question, the probability mass concentrates around two peaks, one near zero saliency and another at a substantially higher value.
The near-zero peak suggests that, for a substantial subset of samples, the question-to-answer information flow contributes minimally to hallucination detection, whereas the higher peak reflects strong dependence on such flow.

\begin{figure*}[!htb]
\centering
\includegraphics[width=0.95\textwidth]{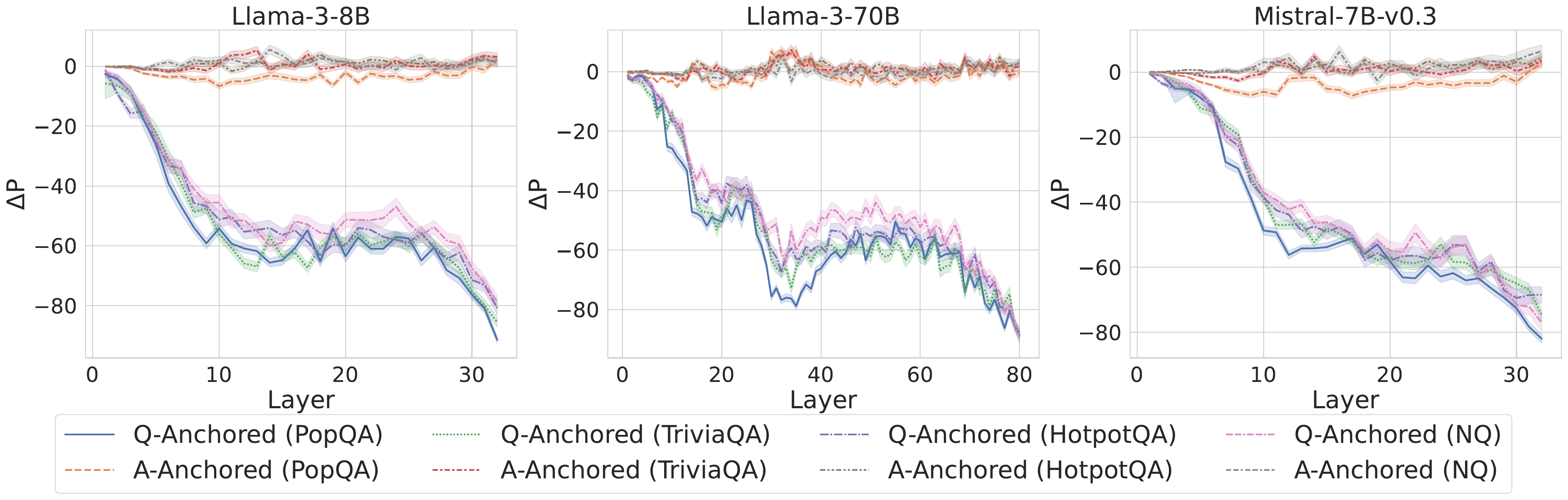}
\caption{$\Delta \mathrm{P}$ under attention knockout. The layer axis indicates the Transformer layer on which the probe is trained. Shaded regions indicate 95\% confidence intervals. Full results in Appendix \ref{sec:appendix_attn_knockout}.}
\label{fig:Disentangling Information Mechanisms attn knockout}
\end{figure*}

\paragraph{Hypothesis}

These observations lead to the hypothesis that there are \textbf{two distinct mechanisms of internal truthfulness encoding} for hallucination detection:  
(1) one characterized by strong reliance on the key question-to-answer information from the exact question tokens, and  
(2) one in which truthfulness encoding is largely independent of the question.  
We validate the proposed hypothesis through further experiments in the next section.

\subsection{Disentangling Information Mechanisms}
\label{sec:Disentangling Information Mechanisms}

We hypothesize that the internal truthfulness encoding operates through two distinct information flow mechanisms, driven by the attention modules within Transformer blocks. To validate the hypothesis, we first block information flows associated with the exact question tokens and analyze the resulting changes in the probe's predictions. Subsequently, we apply a complementary technique, called token patching, to further substantiate the existence of these two mechanisms. Finally, we demonstrate that the self-contained information from the LLM-generated answer itself drives the truthfulness encoding for the A-Anchored type.

\subsubsection{Experimental Setup}
\label{sec:3.2Experimental Setup}

Our analysis covers a diverse collection of 12 LLMs that vary in both scale and architectural design. Specifically, we consider three categories: (1) \emph{base models}, including Llama-3.2-1B \citep{grattafiori2024llama3herdmodels}, Llama-3.2-3B, Llama-3-8B, Llama-3-70B, Mistral-7B-v0.1 \citep{jiang2023mistral7b}, and Mistral-7B-v0.3; (2) \emph{instruction-tuned models}, including Llama-3.2-3B-Instruct, Llama-3-8B-Instruct, Mistral-7B-Instruct-v0.1, and Mistral-7B-Instruct-v0.3; and (3) \emph{reasoning-oriented models}, namely Qwen3-8B \citep{yang2025qwen3technicalreport} and Qwen3-32B.
We conduct experiments on 4 widely used question-answering datasets: PopQA \citep{mallen-etal-2023-trust}, TriviaQA \citep{joshi-etal-2017-triviaqa}, HotpotQA \citep{yang-etal-2018-hotpotqa}, and Natural Questions \citep{kwiatkowski-etal-2019-natural}. Additional implementation details are provided in Appendix \ref{sec:appendix all Implementation Details}.

\subsubsection{Identifying Anchored Modes via Attention Knockout}

\paragraph{Experiment} 

To investigate whether internal truthfulness encoding operates via distinct information mechanisms, we perform an attention knockout experiment targeting the exact question tokens. Specifically, for a probe trained on representations from the $k$-th layer, we set $A_l(i, E_Q) = 0$ for layers $l \in \{1, \dots, k\}$ and positions $i > E_Q$.  
This procedure blocks the information flow from question tokens to subsequent positions in the representation.  
We then examine how the probe’s predictions respond to this intervention. To provide a clearer picture, instances are categorized according to whether their prediction $\hat{z}$ changes after the attention knockout: 
\begin{equation}
    \text{Mode}(x) =
\begin{cases}
\text{Q-Anchored}, & \text{if } \hat{z} \neq \tilde{\hat{z}}   \\
\text{A-Anchored}, &  \text{otherwise}
\end{cases}
\end{equation}
where $\hat{z}$ and $\tilde{\hat{z}}$ denote predictions before and after the attention knockout, respectively. 

\begin{figure*}[!htb]
\centering
\includegraphics[width=0.9\textwidth]{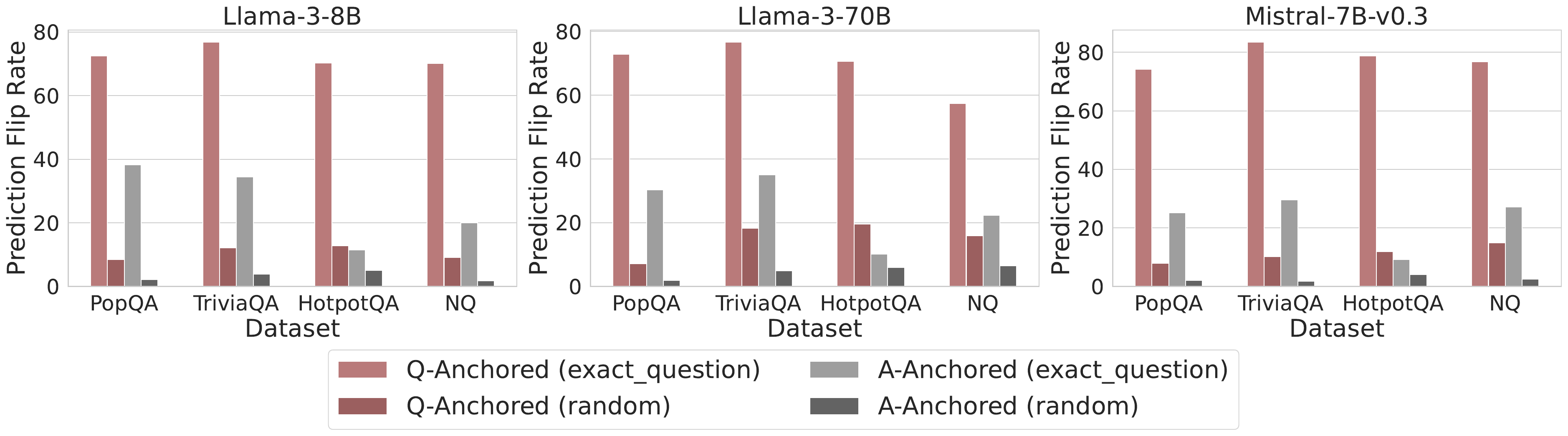}
\caption{Prediction flip rate under token patching. Q-Anchored samples demonstrate significantly higher sensitivity than the counterparts when hallucinatory cues are injected into exact questions. Full results in Appendix \ref{sec:appendix_token_patching}.}
\label{fig:Disentangling Information Mechanisms critical token patching}
\end{figure*}

\paragraph{Results}

The results in Figure \ref{fig:Disentangling Information Mechanisms attn knockout} and Appendix \ref{sec:appendix_attn_knockout} reveal a clear bifurcation of behaviors: for one subset of instances, probabilities shift substantially, while for another subset, probabilities remain nearly unchanged across all layers. 
Shaded regions indicate 95\% confidence intervals, confirming that this qualitative separation is statistically robust. 
This sharp divergence supports the hypothesis that internal truthfulness encoding operates via two distinct mechanisms with respect to question-answer information. In Appendix \ref{sec:appendix_attn_knockout}, we conduct a comprehensive analysis of alternative configurations for token selection, activation extraction, and various instruction- or reasoning-oriented models, and observe consistent patterns across all settings. 
Moreover, Figure \ref{fig:appendix_attention_knockout_base_randomblocking_mlpact_exactans} in Appendix \ref{sec:appendix_attn_knockout} shows that blocking information from randomly selected question tokens yields negligible changes, in contrast to blocking exact question tokens, underscoring the nontrivial nature of the identified mechanisms.

\subsubsection{Further Validation via Token Patching}

\paragraph{Experiment} 

To further validate our findings, we employ a critical token patching technique to investigate how the internal representations of the LLM respond to hallucinatory signals originating from exact question tokens under the two proposed mechanisms.  
Given a context sample $d_c$, we randomly select a patch sample $d_p$ and replace the original question tokens $E_Q^c$ in $d_c$ with the exact question tokens $E_Q^p$ from $d_p$.  
This operation introduces hallucinatory cues into the context sample, allowing us to assess whether the LLM's internal states appropriately reflect the injected changes.  
We restrict our analysis to context instances where the original LLM answers are factual, ensuring that any observed changes can be attributed solely to the injected hallucinatory cues.

\paragraph{Results}

We measure the sensitivity of the truthfulness signals using the prediction flip rate, defined as the frequency with which the probe’s prediction changes after hallucinatory cues are introduced.
Figure \ref{fig:Disentangling Information Mechanisms critical token patching} and Appendix \ref{sec:appendix_token_patching} present the results of the best-performing layer of each model on four datasets when patching the exact subject tokens.
Across models and datasets, Q-Anchored mode exhibits significantly higher sensitivity compared to A-Anchored mode when exposed to hallucination cues from the questions. Furthermore, within each pathway, the flip rates where exact question tokens are patched are substantially higher than those observed when random tokens are patched, ruling out the possibility that the observed effects are mainly due to general semantic disruption from token replacement.
These consistent results provide further support for our hypothesis regarding distinct mechanisms of information pathways.

\subsubsection{What Drives A-Anchored Encoding?}

\paragraph{Experiment} 

Since the A-Anchored mode operates largely independently of the question-to-answer information flow, it is important to investigate the source of information it uses to identify hallucinations.  
To this end, we remove the questions entirely from each sample and perform a separate forward pass using only the LLM-generated answers. This procedure yields answer-only hidden states, which are subsequently provided as input to the probe.
We then evaluate how the probe's predictions change under this “answer-only” condition.
This setup enables us to assess whether A-Anchored predictions rely primarily on the generated answer itself rather than on the original question.

\paragraph{Results}

\begin{figure*}[!htb]
\centering
\includegraphics[width=0.85\textwidth]{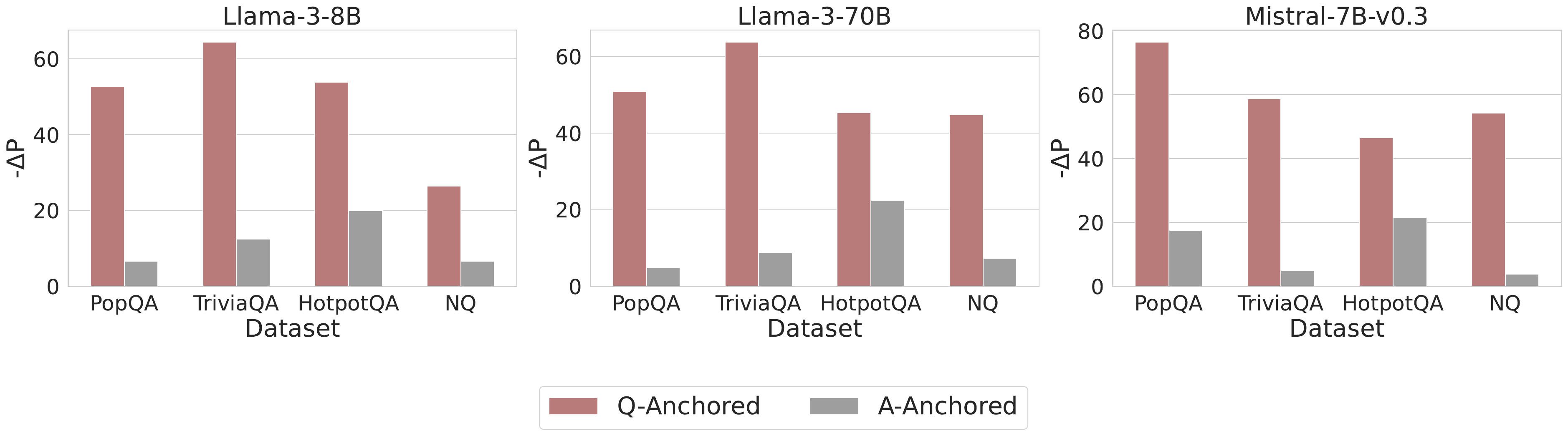}
\caption{$-\Delta \mathrm{P}$ with only the LLM-generated answer. Q-Anchored instances exhibit substantial shifts, whereas A-Anchored instances remain stable, confirming that A-Anchored truthfulness encoding relies on information in the LLM-generated answer itself. Full results in Appendix \ref{sec:appendix_answer_only_input}.}
\label{fig:Disentangling Information Mechanisms answer-only}
\end{figure*}

As shown in Figure \ref{fig:Disentangling Information Mechanisms answer-only} and Appendix \ref{sec:appendix_answer_only_input}, Q-Anchored instances exhibit substantial changes in prediction probability when the question is removed, reflecting their dependence on question-to-answer information.
In contrast, A-Anchored instances remain largely invariant, indicating that the probe continues to detect hallucinations using information encoded within the LLM-generated answer itself.
These findings suggest that the A-Anchored mechanism primarily leverages self-contained answer information to build signals about truthfulness.

\section{Properties of Truthfulness Pathways}

This section examines notable properties and distinct behaviors of intrinsic truthfulness encoding: 
\textbf{(1) Associations with knowledge boundaries:} samples within the LLM's knowledge boundary tend to encode truthfulness via the Q-Anchored pathway, whereas samples beyond the boundary often rely on the A-Anchored signal;
\textbf{(2) Self-awareness:} internal representations can be used to predict which mechanism is being employed, suggesting that LLMs possess intrinsic awareness of pathway distinctions.

\subsection{Associations with Knowledge Boundaries}

We find that distinct patterns of truthfulness encoding are closely associated with the knowledge boundaries of LLMs.
To characterize these boundaries, three complementary metrics are employed:
\textbf{(1) Answer accuracy}, the most direct indicator of an LLM’s factual competence;
\textbf{(2) I-don’t-know rate} (shown in Appendix \ref{sec:appendix_idk}), which reflects the model’s ability to recognize and express its own knowledge limitations;
\textbf{(3) Entity popularity}, which is widely used to distinguish between common and long-tail factual knowledge \citep{mallen-etal-2023-trust}.

\begin{figure*}[!htb]
\centering
\includegraphics[width=\textwidth]{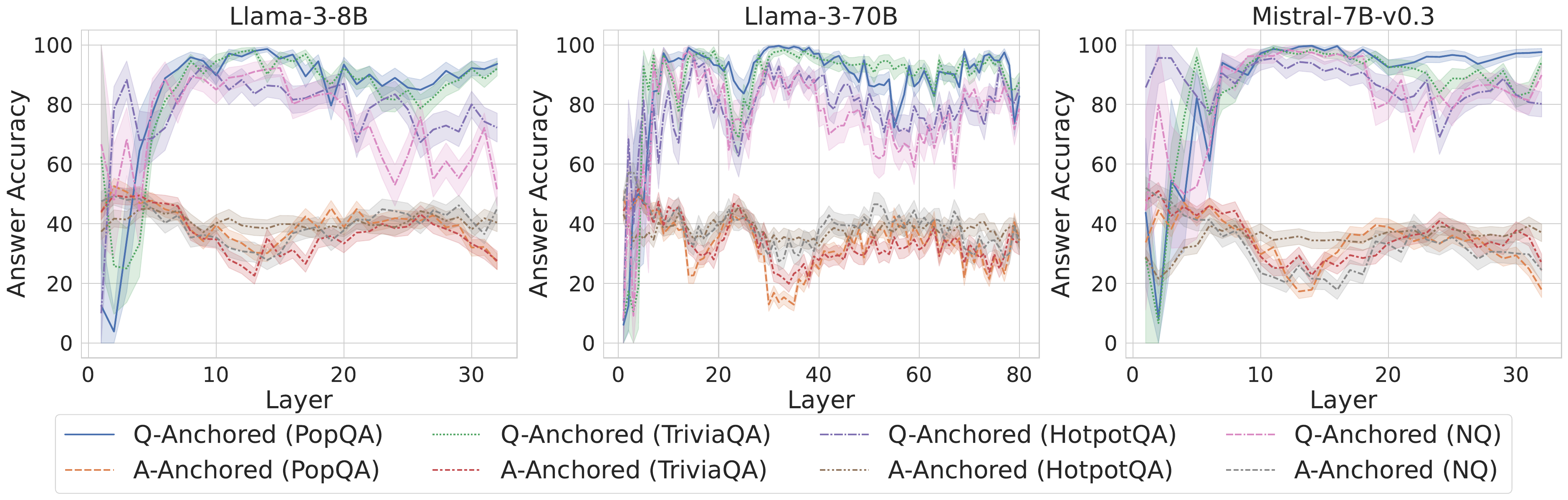}
\caption{Comparisons of answer accuracy between pathways. Q-Anchored samples show higher accuracy than A-Anchored ones, highlighting the association between truthfulness encoding and LLM knowledge boundaries. Full results in Appendix \ref{sec:appendix_answer_accuracy} and \ref{sec:appendix_idk}.}
\label{fig:Associations with Knowledge Boundaries Acc}
\end{figure*}

\begin{figure}[!htb]
\centering
\includegraphics[width=0.9\columnwidth]{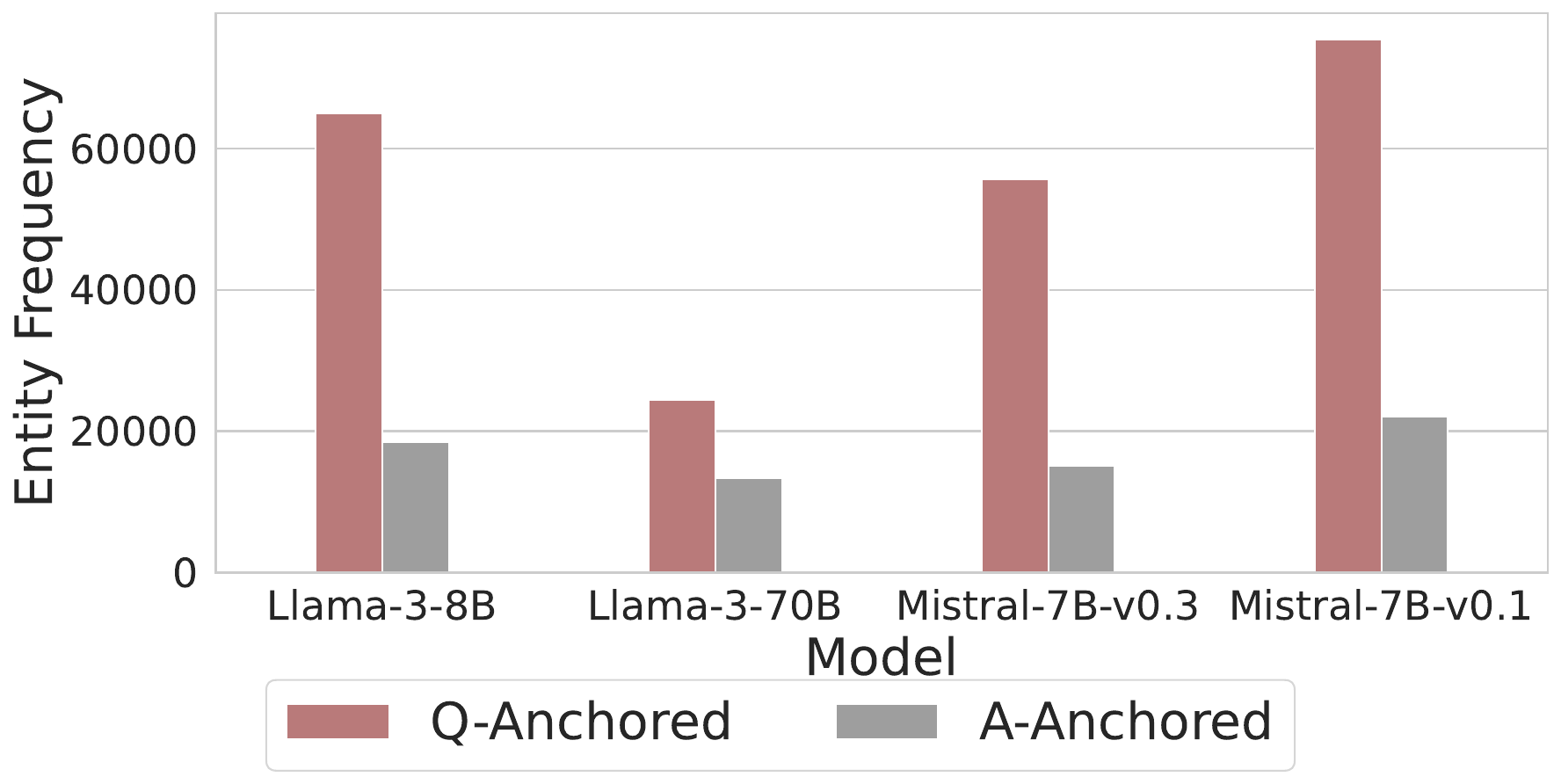}
\caption{Entity frequency distributions for both pathways on PopQA. Q-Anchored samples concentrate on more popular entities, whereas A-Anchored samples skew toward long-tail entities.}
\label{fig:Associations with Knowledge Boundaries frequency}
\end{figure}

As shown in Figure \ref{fig:Associations with Knowledge Boundaries Acc} and Appendix \ref{sec:appendix_answer_accuracy}, Q-Anchored samples achieve significantly higher accuracy than those driven by the A-Anchored pathway.
The results for the I-don’t-know rate, reported in Appendix \ref{sec:appendix_idk}, exhibit trends consistent with answer accuracy, further indicating stronger knowledge handling in Q-Anchored samples.
Moreover, entity popularity, shown in Figure \ref{fig:Associations with Knowledge Boundaries frequency}, provides a more fine-grained perspective on knowledge boundaries.
Specifically, Q-Anchored samples tend to involve more popular entities, whereas A-Anchored samples are more frequently associated with less popular, long-tail factual knowledge.
These findings suggest that truthfulness encoding is strongly aligned with the availability of stored knowledge:
when LLMs possess the requisite knowledge, they predominantly rely on question-answer information flow (Q-Anchored);
when knowledge is unavailable, they instead draw upon internal patterns within their own generated outputs (A-Anchored).

\subsection{Self-Awareness of Pathway Distinctions}
\label{sec:Intrinsic Awareness of Encoding Pathway Selection}

Given that LLMs encode truthfulness via two distinct mechanisms, this section investigates whether their internal representations contain discriminative information that can be used to distinguish between these mechanisms. 
To this end, we train probing classifiers on the models' original internal states (i.e., without knockout interventions) to predict which mechanism is being utilized.

Table \ref{tab:Intrinsic Awareness of Encoding Pathway Selection} reports the pathway classification results of the best-performing layers in hallucination detection across different models. Our findings demonstrate that different mechanisms can be reliably inferred from internal representations, suggesting that, in addition to encoding truthfulness, LLMs exhibit intrinsic awareness of pathway distinctions. 
These findings highlight a potential avenue for fine-grained improvements targeting specific truthfulness encoding mechanisms.

\begin{table}[!htb]
\centering
\begin{adjustbox}{valign=c, width=\columnwidth}
\begin{tabular}{l|ccc}
\toprule
Datasets & Llama-3-8B & Llama-3-70B & Mistral-7B-v0.3\\
\midrule
PopQA & 87.80 & 92.66 & 87.64\\
TriviaQA & 75.10 & 83.91 & 85.87\\
HotpotQA & 86.31 & 87.34 & 92.13\\
NQ & 78.31 & 84.14 & 84.83\\
\bottomrule
\end{tabular}
\end{adjustbox}
\caption{AUCs for encoding pathway classification. The predictability from internal representations indicates that LLMs possess intrinsic awareness of pathway distinctions.}
\label{tab:Intrinsic Awareness of Encoding Pathway Selection}
\end{table}

\section{Pathway-Aware Detection}

Building on the intriguing findings, we explore how the discovered pathway distinctions can be leveraged to improve hallucination detection. 
Specifically, two simple yet effective pathway-aware strategies are proposed: 
\textbf{(1) Mixture-of-Probes (MoP)} (\S \ref{sec:Mixture-of-Probes}), which allows expert probes to specialize in Q-Anchored and A-Anchored pathways respectively, and
\textbf{(2) Pathway Reweighting (PR)} (\S \ref{sec:Pathway Reweighting}), a plug-and-play approach that amplifies pathway-relevant cues salient for detection.

\subsection{Mixture-of-Probes}
\label{sec:Mixture-of-Probes}

Motivated by the fundamentally different dependencies of the two encoding pathways and the LLMs' intrinsic awareness of them, we propose a \textbf{Mixture-of-Probes (MoP)} framework that explicitly captures this heterogeneity.  
Rather than training a single probe to handle all inputs, MoP employs two pathway-specialized experts and leverages the self-awareness probe (\S \ref{sec:Intrinsic Awareness of Encoding Pathway Selection}) as a gating network to combine their predictions.
Let $\mathbf{h}^{l^*}(x)\!\in\!\mathbb{R}^d$ be the token hidden state from the best detection layer $l^*$.  
Two expert probes $p_Q(\cdot)$ and $p_A(\cdot)$ are trained separately for two pathway samples, and the self-awareness probe provides a gating coefficient $\pi(\mathbf{h}^{l^*}(x))\!\in\![0,1]$.
The final prediction is a convex combination, \textbf{requiring no extra training}:
\begin{equation}
\small
\begin{aligned}
p_{\text{MoP}}(z\!=\!1\mid \mathbf{h}^{l^*}(x))
&= \pi_Q\, p_Q(z\!=\!1\mid \mathbf{h}^{l^*}(x))\\
&\quad + (1-\pi_Q)\, p_A(z\!=\!1\mid \mathbf{h}^{l^*}(x)).
\end{aligned}
\end{equation}

\subsection{Pathway Reweighting}
\label{sec:Pathway Reweighting}

From the perspective of emphasizing pathway-relevant internal cues, we introduce a plug-and-play \textbf{Pathway Reweighting (PR)} method that directly modulates the question-answer information flow.  
The key idea is to adjust the attention from exact answer to question tokens according to the predicted pathway, amplifying the signals most salient for hallucination detection.
For each layer $l \leq l^*$, two learnable scalars $\alpha_Q^{l},\alpha_A^{l}>0$ are introduced.  
Given self-awareness probability $\pi(\mathbf{h}^{l^*}(x))$, we rescale attention edges $i\!\in\!E_A$, $j\!\in\!E_Q$ to construct representations tailored for detection:
\begin{equation}
\small
\tilde{A}^{l}(i,j)=
\begin{cases}
\bigl[1 + s(\mathbf{h}^{l^*}(x))\bigr] A^{l}(i,j), & i\!\in\!E_A, j\!\in\!E_Q,\\
A^{l}(i,j), & \text{otherwise},
\end{cases}
\end{equation}
where
\begin{equation}
\small
s(\mathbf{h}^{l^*}(x)) = \pi_Q\,\alpha_Q^{l} - (1-\pi_Q)\,\alpha_A^{l}.
\end{equation}
The extra parameters serve as a lightweight adapter, used only during detection to guide salient truthfulness cues and omitted during generation, leaving the generation capacity unaffected.

\begin{table*}[!htb]
\centering
\begin{adjustbox}{valign=c, width=0.85\textwidth}
\begin{tabular}{lcccccccc}
\toprule
\multirow{2}{*}{Method} & \multicolumn{4}{c}{Llama-3-8B} & \multicolumn{4}{c}{Mistral-7B-v0.3} \\
\cmidrule(lr){2-5} \cmidrule(lr){6-9}
 & PopQA & TriviaQA & HotpotQA  & NQ & PopQA & TriviaQA & HotpotQA  & NQ\\
\midrule
P(True) & 55.85 & 49.92 & 52.14 & 53.27 & 45.49 & 47.61 & 57.87 & 52.79\\
Logits-mean & 74.52 & 60.39 & 51.94 & 52.63 & 69.52 & 66.76 & 55.45 & 57.88\\
Logits-min & 85.36 & 70.89 & 61.28 & 56.50 & 87.05 & 77.33 & 68.08 & 54.40\\
Probing Baseline & 88.71 & 77.58 & 82.23 & 70.20 & 87.39 & 81.74 & 83.19 & 73.60\\
\rowcolor{mygray} MoP-RandomGate & 75.52 & 69.17 & 79.88 & 66.56 & 79.81 & 70.88 & 72.23 & 61.19 \\
\rowcolor{mygray} MoP-VanillaExperts & 89.11 & 78.73 & 84.57 & 71.21 & 88.53 & 80.93 & 82.93 & 73.77 \\
\rowcolor{mygray} MoP & 92.11 & 81.18 & 85.45 & 74.64 & 91.66 & 83.57 & 85.82 & 76.87\\
\rowcolor{mygray} PR & \textbf{94.01} & \textbf{83.13} & \textbf{87.81} & \textbf{79.10} & \textbf{93.09} & \textbf{84.36} & \textbf{89.03} & \textbf{79.09} \\
\bottomrule
\end{tabular}
\end{adjustbox}
\caption{Comparison of hallucination detection performance (AUC). Full results in Appendix \ref{sec:appendix_applications}.}
\label{tab:Applications: Pathway-Aware Hallucination Detection}
\end{table*}

\subsection{Experiments}

\paragraph{Setup}
The experimental setup follows Section \ref{sec:3.2Experimental Setup}. 
We compare our method against several internal-based baselines, including 
(1) P(True) \citep{Kadavath2022LanguageM},
(2) uncertainty-based metrics \citep{aichberger2024semantically,xue-etal-2025-ualign}, and
(3) probing classifiers \citep{DBLP:conf/iclr/0026L0GWTFY24,DBLP:conf/iclr/OrgadTGRSKB25}. 
Results are averaged over three random seeds.
Additional implementation details are provided in Appendix \ref{sec:appendix Baseline Implementation Details} and \ref{sec:appendix Our Implementation Details}.

\paragraph{Results}

As shown in Table \ref{tab:Applications: Pathway-Aware Hallucination Detection} and Appendix \ref{sec:appendix_applications}, both MoP and PR consistently outperform competing approaches across different datasets and model scales. 
Specifically, for MoP, we further examine two ablated variants: (1) MoP-RandomGate, which randomly routes the two pathway experts without leveraging the self-awareness probe; and (2) MoP-VanillaExperts, which replaces the expert probes with two vanilla probes to serve as a simple ensemble strategy. 
Both ablated variants exhibit substantially degraded performance compared to MoP, underscoring the roles of pathway specialization and self-awareness gating. 
For PR, the method proves particularly effective in improving performance by dynamically adjusting the focus on salient truthfulness cues. 
These results demonstrate that explicitly modeling truthfulness encoding heterogeneity can effectively translate the insights of our analysis into practical gains for hallucination detection.

\section{Related Work}

Hallucination detection in LLMs has received increasing attention because of its critical role in building reliable and trustworthy generative systems \citep{DBLP:conf/acl/TianGSZ024,DBLP:conf/acl/Shi00GRCR24,DBLP:conf/acl/BaiLBHLZLSG0O24}.  
Existing approaches can be broadly grouped by whether they rely on external resources (e.g., retrieval systems or fact-checking APIs). 
Externally assisted methods cross-verify output texts against external knowledge bases \citep{DBLP:conf/emnlp/MinKLLYKIZH23,DBLP:conf/acl/HuZJZW025,DBLP:conf/acl/HuangFMFFGYZZWW25} or specialized LLM judges \citep{luo2024halludiallargescalebenchmarkautomatic,bouchard2025uncertainty,zhang2025siren}.  
Resource-free methods avoid external data and instead exploit the model’s own intermediate computations.  
Some leverage the model’s self-awareness of knowledge boundaries \citep{Kadavath2022LanguageM,luo-etal-2025-odysseus}, while others use uncertainty-based measures \citep{aichberger2024semantically,xue-etal-2025-ualign}, treating confidence as a proxy for truthfulness.  
These techniques analyze output distributions (e.g., logits) \citep{aichberger2024semantically}, variance across multiple samples (e.g., consistency) \citep{DBLP:conf/emnlp/MinKLLYKIZH23,DBLP:conf/iclr/AichbergerSIH25}, or other statistical indicators of prediction uncertainty \citep{xue2025verify}.  
Another line of work trains linear probing classifiers on hidden representations to capture intrinsic truthfulness signals.  
Prior work \citep{DBLP:conf/iclr/BurnsYKS23,DBLP:conf/nips/0002PVPW23,DBLP:conf/iclr/0026L0GWTFY24,DBLP:conf/iclr/OrgadTGRSKB25} shows that LLMs encode rich latent features correlated with factual accuracy, enabling efficient detection with minimal overhead.  
Yet the mechanisms behind these internal truthfulness encoding remain poorly understood.  
Compared to previous approaches, our work addresses this gap by dissecting how such intrinsic signals emerge and operate, revealing distinct information pathways that not only yield explanatory insights but also enhance detection performance.

\section{Conclusion}

We investigate how LLMs encode truthfulness, revealing two complementary pathways: a \textbf{Question-Anchored} pathway relying on question-answer flow, and an \textbf{Answer-Anchored} pathway extracting self-contained evidence from generated outputs. Analyses across datasets and models highlight their ties to knowledge boundaries and intrinsic self-awareness. Building on these insights, we further propose two applications to improve hallucination detection. Overall, our findings not only advance mechanistic understanding of intrinsic truthfulness encoding but also offer practical applications for building more reliable generative systems.

\section*{Limitations}

While this work provides a systematic analysis of intrinsic truthfulness encoding mechanisms in LLMs and demonstrates their utility for hallucination detection, one limitation is that, similar to prior work on mechanistic interpretability, our analyses and pathway-aware applications assume access to internal model representations. Such access may not always be available in strictly black-box settings. In these scenarios, additional engineering or alternative approximations may be required for practical deployment, which we leave for future work.

\section*{Ethics Statement}

Our work presents minimal potential for negative societal impact, primarily due to the use of publicly available datasets and models. This accessibility inherently reduces the risk of adverse effects on individuals or society.

\section*{Acknowledgments}

This work was supported by Beijing Natural Science Foundation (L253020)  \&  the Academic Research Projects of Beijing Union University (NO.ZK10202405).

\bibliography{custom}

\begin{thebibliography}{38}
\providecommand{\natexlab}[1]{#1}

\bibitem[{Aichberger et~al.(2024)Aichberger, Schweighofer, Ielanskyi, and Hochreiter}]{aichberger2024semantically}
Lukas Aichberger, Kajetan Schweighofer, Mykyta Ielanskyi, and Sepp Hochreiter. 2024.
\newblock Semantically diverse language generation for uncertainty estimation in language models.
\newblock \emph{arXiv preprint arXiv:2406.04306}.

\bibitem[{Aichberger et~al.(2025)Aichberger, Schweighofer, Ielanskyi, and Hochreiter}]{DBLP:conf/iclr/AichbergerSIH25}
Lukas Aichberger, Kajetan Schweighofer, Mykyta Ielanskyi, and Sepp Hochreiter. 2025.
\newblock \href {https://openreview.net/forum?id=HSi4VetQLj} {Improving uncertainty estimation through semantically diverse language generation}.
\newblock In \emph{The Thirteenth International Conference on Learning Representations, {ICLR} 2025, Singapore, April 24-28, 2025}. OpenReview.net.

\bibitem[{Bai et~al.(2024)Bai, Liu, Bu, He, Liu, Zhou, Lin, Su, Ge, Zheng, and Ouyang}]{DBLP:conf/acl/BaiLBHLZLSG0O24}
Ge~Bai, Jie Liu, Xingyuan Bu, Yancheng He, Jiaheng Liu, Zhanhui Zhou, Zhuoran Lin, Wenbo Su, Tiezheng Ge, Bo~Zheng, and Wanli Ouyang. 2024.
\newblock \href {https://doi.org/10.18653/V1/2024.ACL-LONG.401} {Mt-bench-101: {A} fine-grained benchmark for evaluating large language models in multi-turn dialogues}.
\newblock In \emph{Proceedings of the 62nd Annual Meeting of the Association for Computational Linguistics (Volume 1: Long Papers), {ACL} 2024, Bangkok, Thailand, August 11-16, 2024}, pages 7421--7454. Association for Computational Linguistics.

\bibitem[{Baker et~al.(1998)Baker, Fillmore, and Lowe}]{baker1998berkeley}
Collin~F Baker, Charles~J Fillmore, and John~B Lowe. 1998.
\newblock The berkeley framenet project.
\newblock In \emph{36th Annual Meeting of the Association for Computational Linguistics and 17th International Conference on Computational Linguistics, Volume 1}, pages 86--90.

\bibitem[{Bouchard and Chauhan(2025)}]{bouchard2025uncertainty}
Dylan Bouchard and Mohit~Singh Chauhan. 2025.
\newblock Uncertainty quantification for language models: A suite of black-box, white-box, llm judge, and ensemble scorers.
\newblock \emph{arXiv preprint arXiv:2504.19254}.

\bibitem[{Burns et~al.(2023)Burns, Ye, Klein, and Steinhardt}]{DBLP:conf/iclr/BurnsYKS23}
Collin Burns, Haotian Ye, Dan Klein, and Jacob Steinhardt. 2023.
\newblock \href {https://openreview.net/forum?id=ETKGuby0hcs} {Discovering latent knowledge in language models without supervision}.
\newblock In \emph{The Eleventh International Conference on Learning Representations, {ICLR} 2023, Kigali, Rwanda, May 1-5, 2023}. OpenReview.net.

\bibitem[{Chen et~al.(2024)Chen, Liu, Chen, Gu, Wu, Tao, Fu, and Ye}]{DBLP:conf/iclr/0026L0GWTFY24}
Chao Chen, Kai Liu, Ze~Chen, Yi~Gu, Yue Wu, Mingyuan Tao, Zhihang Fu, and Jieping Ye. 2024.
\newblock \href {https://openreview.net/forum?id=Zj12nzlQbz} {{INSIDE:} llms' internal states retain the power of hallucination detection}.
\newblock In \emph{The Twelfth International Conference on Learning Representations, {ICLR} 2024, Vienna, Austria, May 7-11, 2024}. OpenReview.net.

\bibitem[{Fierro et~al.(2025)Fierro, Foroutan, Elliott, and S{\o}gaard}]{DBLP:conf/acl/FierroFES25}
Constanza Fierro, Negar Foroutan, Desmond Elliott, and Anders S{\o}gaard. 2025.
\newblock \href {https://aclanthology.org/2025.findings-acl.827/} {How do multilingual language models remember facts?}
\newblock In \emph{Findings of the Association for Computational Linguistics, {ACL} 2025, Vienna, Austria, July 27 - August 1, 2025}, pages 16052--16106. Association for Computational Linguistics.

\bibitem[{Geva et~al.(2023)Geva, Bastings, Filippova, and Globerson}]{DBLP:conf/emnlp/GevaBFG23}
Mor Geva, Jasmijn Bastings, Katja Filippova, and Amir Globerson. 2023.
\newblock \href {https://doi.org/10.18653/V1/2023.EMNLP-MAIN.751} {Dissecting recall of factual associations in auto-regressive language models}.
\newblock In \emph{Proceedings of the 2023 Conference on Empirical Methods in Natural Language Processing, {EMNLP} 2023, Singapore, December 6-10, 2023}, pages 12216--12235. Association for Computational Linguistics.

\bibitem[{Ghandeharioun et~al.(2024)Ghandeharioun, Caciularu, Pearce, Dixon, and Geva}]{DBLP:conf/icml/GhandehariounCP24}
Asma Ghandeharioun, Avi Caciularu, Adam Pearce, Lucas Dixon, and Mor Geva. 2024.
\newblock \href {https://openreview.net/forum?id=5uwBzcn885} {Patchscopes: {A} unifying framework for inspecting hidden representations of language models}.
\newblock In \emph{Forty-first International Conference on Machine Learning, {ICML} 2024, Vienna, Austria, July 21-27, 2024}. OpenReview.net.

\bibitem[{Grattafiori et~al.(2024)Grattafiori, Dubey, Jauhri, Pandey, Kadian, Al-Dahle, Letman, Mathur, Schelten, Vaughan, Yang, Fan, Goyal, Hartshorn, Yang, Mitra, Sravankumar, Korenev, Hinsvark, Rao, Zhang, Rodriguez, Gregerson, Spataru, Roziere, Biron, Tang, Chern, Caucheteux, Nayak, Bi, Marra, McConnell, Keller, Touret, Wu, Wong, Ferrer, Nikolaidis, Allonsius, Song, Pintz, Livshits, Wyatt, Esiobu, Choudhary, Mahajan, Garcia-Olano, Perino, Hupkes, Lakomkin, AlBadawy, Lobanova, Dinan, Smith, Radenovic, Guzmán, Zhang, Synnaeve, Lee, Anderson, Thattai, Nail, Mialon, Pang, Cucurell, Nguyen, Korevaar, Xu, Touvron, Zarov, Ibarra, Kloumann, Misra, Evtimov, Zhang, Copet, Lee, Geffert, Vranes, Park, Mahadeokar, Shah, van~der Linde, Billock, Hong, Lee, Fu, Chi, Huang, Liu, Wang, Yu, Bitton, Spisak, Park, Rocca, Johnstun, Saxe, Jia, Alwala, Prasad, Upasani, Plawiak, Li, Heafield, Stone, El-Arini, Iyer, Malik, Chiu, Bhalla, Lakhotia, Rantala-Yeary, van~der Maaten, Chen, Tan, Jenkins, Martin, Madaan, Malo, Blecher,
  Landzaat, de~Oliveira, Muzzi, Pasupuleti, Singh, Paluri, Kardas, Tsimpoukelli, Oldham, Rita, Pavlova, Kambadur, Lewis, Si, Singh, Hassan, Goyal, Torabi, Bashlykov, Bogoychev, Chatterji, Zhang, Duchenne, Çelebi, Alrassy, Zhang, Li, Vasic, Weng, Bhargava, Dubal, Krishnan, Koura, Xu, He, Dong, Srinivasan, Ganapathy, Calderer, Cabral, Stojnic, Raileanu, Maheswari, Girdhar, Patel, Sauvestre, Polidoro, Sumbaly, Taylor, Silva, Hou, Wang, Hosseini, Chennabasappa, Singh, Bell, Kim, Edunov, Nie, Narang, Raparthy, Shen, Wan, Bhosale, Zhang, Vandenhende, Batra, Whitman, Sootla, Collot, Gururangan, Borodinsky, Herman, Fowler, Sheasha, Georgiou, Scialom, Speckbacher, Mihaylov, Xiao, Karn, Goswami, Gupta, Ramanathan, Kerkez, Gonguet, Do, Vogeti, Albiero, Petrovic, Chu, Xiong, Fu, Meers, Martinet, Wang, Wang, Tan, Xia, Xie, Jia, Wang, Goldschlag, Gaur, Babaei, Wen, Song, Zhang, Li, Mao, Coudert, Yan, Chen, Papakipos, Singh, Srivastava, Jain, Kelsey, Shajnfeld, Gangidi, Victoria, Goldstand, Menon, Sharma, Boesenberg,
  Baevski, Feinstein, Kallet, Sangani, Teo, Yunus, Lupu, Alvarado, Caples, Gu, Ho, Poulton, Ryan, Ramchandani, Dong, Franco, Goyal, Saraf, Chowdhury, Gabriel, Bharambe, Eisenman, Yazdan, James, Maurer, Leonhardi, Huang, Loyd, Paola, Paranjape, Liu, Wu, Ni, Hancock, Wasti, Spence, Stojkovic, Gamido, Montalvo, Parker, Burton, Mejia, Liu, Wang, Kim, Zhou, Hu, Chu, Cai, Tindal, Feichtenhofer, Gao, Civin, Beaty, Kreymer, Li, Adkins, Xu, Testuggine, David, Parikh, Liskovich, Foss, Wang, Le, Holland, Dowling, Jamil, Montgomery, Presani, Hahn, Wood, Le, Brinkman, Arcaute, Dunbar, Smothers, Sun, Kreuk, Tian, Kokkinos, Ozgenel, Caggioni, Kanayet, Seide, Florez, Schwarz, Badeer, Swee, Halpern, Herman, Sizov, Guangyi, Zhang, Lakshminarayanan, Inan, Shojanazeri, Zou, Wang, Zha, Habeeb, Rudolph, Suk, Aspegren, Goldman, Zhan, Damlaj, Molybog, Tufanov, Leontiadis, Veliche, Gat, Weissman, Geboski, Kohli, Lam, Asher, Gaya, Marcus, Tang, Chan, Zhen, Reizenstein, Teboul, Zhong, Jin, Yang, Cummings, Carvill, Shepard, McPhie,
  Torres, Ginsburg, Wang, Wu, U, Saxena, Khandelwal, Zand, Matosich, Veeraraghavan, Michelena, Li, Jagadeesh, Huang, Chawla, Huang, Chen, Garg, A, Silva, Bell, Zhang, Guo, Yu, Moshkovich, Wehrstedt, Khabsa, Avalani, Bhatt, Mankus, Hasson, Lennie, Reso, Groshev, Naumov, Lathi, Keneally, Liu, Seltzer, Valko, Restrepo, Patel, Vyatskov, Samvelyan, Clark, Macey, Wang, Hermoso, Metanat, Rastegari, Bansal, Santhanam, Parks, White, Bawa, Singhal, Egebo, Usunier, Mehta, Laptev, Dong, Cheng, Chernoguz, Hart, Salpekar, Kalinli, Kent, Parekh, Saab, Balaji, Rittner, Bontrager, Roux, Dollar, Zvyagina, Ratanchandani, Yuvraj, Liang, Alao, Rodriguez, Ayub, Murthy, Nayani, Mitra, Parthasarathy, Li, Hogan, Battey, Wang, Howes, Rinott, Mehta, Siby, Bondu, Datta, Chugh, Hunt, Dhillon, Sidorov, Pan, Mahajan, Verma, Yamamoto, Ramaswamy, Lindsay, Lindsay, Feng, Lin, Zha, Patil, Shankar, Zhang, Zhang, Wang, Agarwal, Sajuyigbe, Chintala, Max, Chen, Kehoe, Satterfield, Govindaprasad, Gupta, Deng, Cho, Virk, Subramanian, Choudhury,
  Goldman, Remez, Glaser, Best, Koehler, Robinson, Li, Zhang, Matthews, Chou, Shaked, Vontimitta, Ajayi, Montanez, Mohan, Kumar, Mangla, Ionescu, Poenaru, Mihailescu, Ivanov, Li, Wang, Jiang, Bouaziz, Constable, Tang, Wu, Wang, Wu, Gao, Kleinman, Chen, Hu, Jia, Qi, Li, Zhang, Zhang, Adi, Nam, Yu, Wang, Zhao, Hao, Qian, Li, He, Rait, DeVito, Rosnbrick, Wen, Yang, Zhao, and Ma}]{grattafiori2024llama3herdmodels}
Aaron Grattafiori, Abhimanyu Dubey, Abhinav Jauhri, Abhinav Pandey, Abhishek Kadian, Ahmad Al-Dahle, Aiesha Letman, Akhil Mathur, Alan Schelten, Alex Vaughan, Amy Yang, Angela Fan, Anirudh Goyal, Anthony Hartshorn, Aobo Yang, Archi Mitra, Archie Sravankumar, Artem Korenev, Arthur Hinsvark, and 542 others. 2024.
\newblock \href {https://arxiv.org/abs/2407.21783} {The llama 3 herd of models}.
\newblock \emph{Preprint}, arXiv:2407.21783.

\bibitem[{Hu et~al.(2025)Hu, Zhang, Jiang, Zhang, Wei, and Li}]{DBLP:conf/acl/HuZJZW025}
Wentao Hu, Wengyu Zhang, Yiyang Jiang, Chen~Jason Zhang, Xiaoyong Wei, and Qing Li. 2025.
\newblock \href {https://aclanthology.org/2025.acl-long.770/} {Removal of hallucination on hallucination: Debate-augmented {RAG}}.
\newblock In \emph{Proceedings of the 63rd Annual Meeting of the Association for Computational Linguistics (Volume 1: Long Papers), {ACL} 2025, Vienna, Austria, July 27 - August 1, 2025}, pages 15839--15853. Association for Computational Linguistics.

\bibitem[{Huang et~al.(2025)Huang, Feng, Ma, Fan, Feng, Gu, Ye, Zhao, Zhong, Wang, Wu, Hu, Kong, Xiao, Liu, and Qin}]{DBLP:conf/acl/HuangFMFFGYZZWW25}
Lei Huang, Xiaocheng Feng, Weitao Ma, Yuchun Fan, Xiachong Feng, Yuxuan Gu, Yangfan Ye, Liang Zhao, Weihong Zhong, Baoxin Wang, Dayong Wu, Guoping Hu, Lingpeng Kong, Tong Xiao, Ting Liu, and Bing Qin. 2025.
\newblock \href {https://aclanthology.org/2025.acl-long.1199/} {Alleviating hallucinations from knowledge misalignment in large language models via selective abstention learning}.
\newblock In \emph{Proceedings of the 63rd Annual Meeting of the Association for Computational Linguistics (Volume 1: Long Papers), {ACL} 2025, Vienna, Austria, July 27 - August 1, 2025}, pages 24564--24579. Association for Computational Linguistics.

\bibitem[{Jiang et~al.(2023)Jiang, Sablayrolles, Mensch, Bamford, Chaplot, de~las Casas, Bressand, Lengyel, Lample, Saulnier, Lavaud, Lachaux, Stock, Scao, Lavril, Wang, Lacroix, and Sayed}]{jiang2023mistral7b}
Albert~Q. Jiang, Alexandre Sablayrolles, Arthur Mensch, Chris Bamford, Devendra~Singh Chaplot, Diego de~las Casas, Florian Bressand, Gianna Lengyel, Guillaume Lample, Lucile Saulnier, Lélio~Renard Lavaud, Marie-Anne Lachaux, Pierre Stock, Teven~Le Scao, Thibaut Lavril, Thomas Wang, Timothée Lacroix, and William~El Sayed. 2023.
\newblock \href {https://arxiv.org/abs/2310.06825} {Mistral 7b}.
\newblock \emph{Preprint}, arXiv:2310.06825.

\bibitem[{Joshi et~al.(2017)Joshi, Choi, Weld, and Zettlemoyer}]{joshi-etal-2017-triviaqa}
Mandar Joshi, Eunsol Choi, Daniel Weld, and Luke Zettlemoyer. 2017.
\newblock \href {https://doi.org/10.18653/v1/P17-1147} {{T}rivia{QA}: A large scale distantly supervised challenge dataset for reading comprehension}.
\newblock In \emph{Proceedings of the 55th Annual Meeting of the Association for Computational Linguistics (Volume 1: Long Papers)}, pages 1601--1611, Vancouver, Canada. Association for Computational Linguistics.

\bibitem[{Kadavath et~al.(2022)Kadavath, Conerly, Askell, Henighan, Drain, Perez, Schiefer, Dodds, Dassarma, Tran-Johnson, Johnston, El-Showk, Jones, Elhage, Hume, Chen, Bai, Bowman, Fort, Ganguli, Hernandez, Jacobson, Kernion, Kravec, Lovitt, Ndousse, Olsson, Ringer, Amodei, Brown, Clark, Joseph, Mann, McCandlish, Olah, and Kaplan}]{Kadavath2022LanguageM}
Saurav Kadavath, Tom Conerly, Amanda Askell, T.~J. Henighan, Dawn Drain, Ethan Perez, Nicholas Schiefer, Zachary Dodds, Nova Dassarma, Eli Tran-Johnson, Scott Johnston, Sheer El-Showk, Andy Jones, Nelson Elhage, Tristan Hume, Anna Chen, Yuntao Bai, Sam Bowman, Stanislav Fort, and 17 others. 2022.
\newblock \href {https://api.semanticscholar.org/CorpusID:250451161} {Language models (mostly) know what they know}.
\newblock \emph{ArXiv}, abs/2207.05221.

\bibitem[{Kwiatkowski et~al.(2019)Kwiatkowski, Palomaki, Redfield, Collins, Parikh, Alberti, Epstein, Polosukhin, Devlin, Lee, Toutanova, Jones, Kelcey, Chang, Dai, Uszkoreit, Le, and Petrov}]{kwiatkowski-etal-2019-natural}
Tom Kwiatkowski, Jennimaria Palomaki, Olivia Redfield, Michael Collins, Ankur Parikh, Chris Alberti, Danielle Epstein, Illia Polosukhin, Jacob Devlin, Kenton Lee, Kristina Toutanova, Llion Jones, Matthew Kelcey, Ming-Wei Chang, Andrew~M. Dai, Jakob Uszkoreit, Quoc Le, and Slav Petrov. 2019.
\newblock \href {https://doi.org/10.1162/tacl_a_00276} {Natural questions: A benchmark for question answering research}.
\newblock \emph{Transactions of the Association for Computational Linguistics}, 7:452--466.

\bibitem[{Li et~al.(2023)Li, Patel, Vi{\'{e}}gas, Pfister, and Wattenberg}]{DBLP:conf/nips/0002PVPW23}
Kenneth Li, Oam Patel, Fernanda~B. Vi{\'{e}}gas, Hanspeter Pfister, and Martin Wattenberg. 2023.
\newblock \href {http://papers.nips.cc/paper\_files/paper/2023/hash/81b8390039b7302c909cb769f8b6cd93-Abstract-Conference.html} {Inference-time intervention: Eliciting truthful answers from a language model}.
\newblock In \emph{Advances in Neural Information Processing Systems 36: Annual Conference on Neural Information Processing Systems 2023, NeurIPS 2023, New Orleans, LA, USA, December 10 - 16, 2023}.

\bibitem[{Luo et~al.(2024)Luo, Shen, Li, Peng, Xuan, Wang, and Yang}]{luo2024halludiallargescalebenchmarkautomatic}
Wen Luo, Tianshu Shen, Wei Li, Guangyue Peng, Richeng Xuan, Houfeng Wang, and Xi~Yang. 2024.
\newblock \href {https://arxiv.org/abs/2406.07070} {Halludial: A large-scale benchmark for automatic dialogue-level hallucination evaluation}.
\newblock \emph{Preprint}, arXiv:2406.07070.

\bibitem[{Luo et~al.(2025)Luo, Song, Li, Peng, Wei, and Wang}]{luo-etal-2025-odysseus}
Wen Luo, Feifan Song, Wei Li, Guangyue Peng, Shaohang Wei, and Houfeng Wang. 2025.
\newblock \href {https://doi.org/10.18653/v1/2025.acl-long.1320} {Odysseus navigates the sirens' song: Dynamic focus decoding for factual and diverse open-ended text generation}.
\newblock In \emph{Proceedings of the 63rd Annual Meeting of the Association for Computational Linguistics (Volume 1: Long Papers)}, pages 27200--27218, Vienna, Austria. Association for Computational Linguistics.

\bibitem[{Mallen et~al.(2023)Mallen, Asai, Zhong, Das, Khashabi, and Hajishirzi}]{mallen-etal-2023-trust}
Alex Mallen, Akari Asai, Victor Zhong, Rajarshi Das, Daniel Khashabi, and Hannaneh Hajishirzi. 2023.
\newblock \href {https://doi.org/10.18653/v1/2023.acl-long.546} {When not to trust language models: Investigating effectiveness of parametric and non-parametric memories}.
\newblock In \emph{Proceedings of the 61st Annual Meeting of the Association for Computational Linguistics (Volume 1: Long Papers)}, pages 9802--9822, Toronto, Canada. Association for Computational Linguistics.

\bibitem[{Michel et~al.(2019)Michel, Levy, and Neubig}]{michel2019sixteen}
Paul Michel, Omer Levy, and Graham Neubig. 2019.
\newblock Are sixteen heads really better than one?
\newblock \emph{Advances in neural information processing systems}, 32.

\bibitem[{Min et~al.(2023)Min, Krishna, Lyu, Lewis, Yih, Koh, Iyyer, Zettlemoyer, and Hajishirzi}]{DBLP:conf/emnlp/MinKLLYKIZH23}
Sewon Min, Kalpesh Krishna, Xinxi Lyu, Mike Lewis, Wen{-}tau Yih, Pang~Wei Koh, Mohit Iyyer, Luke Zettlemoyer, and Hannaneh Hajishirzi. 2023.
\newblock \href {https://doi.org/10.18653/V1/2023.EMNLP-MAIN.741} {Factscore: Fine-grained atomic evaluation of factual precision in long form text generation}.
\newblock In \emph{Proceedings of the 2023 Conference on Empirical Methods in Natural Language Processing, {EMNLP} 2023, Singapore, December 6-10, 2023}, pages 12076--12100. Association for Computational Linguistics.

\bibitem[{Niu et~al.(2025)Niu, Haddadi, and Pang}]{niu2025robust}
Mengjia Niu, Hamed Haddadi, and Guansong Pang. 2025.
\newblock Robust hallucination detection in llms via adaptive token selection.
\newblock \emph{arXiv preprint arXiv:2504.07863}.

\bibitem[{Orgad et~al.(2025)Orgad, Toker, Gekhman, Reichart, Szpektor, Kotek, and Belinkov}]{DBLP:conf/iclr/OrgadTGRSKB25}
Hadas Orgad, Michael Toker, Zorik Gekhman, Roi Reichart, Idan Szpektor, Hadas Kotek, and Yonatan Belinkov. 2025.
\newblock \href {https://openreview.net/forum?id=KRnsX5Em3W} {Llms know more than they show: On the intrinsic representation of {LLM} hallucinations}.
\newblock In \emph{The Thirteenth International Conference on Learning Representations, {ICLR} 2025, Singapore, April 24-28, 2025}. OpenReview.net.

\bibitem[{Pagnoni et~al.(2021)Pagnoni, Balachandran, and Tsvetkov}]{pagnoni2021understanding}
Artidoro Pagnoni, Vidhisha Balachandran, and Yulia Tsvetkov. 2021.
\newblock Understanding factuality in abstractive summarization with frank: A benchmark for factuality metrics.
\newblock In \emph{Proceedings of the 2021 Conference of the North American Chapter of the Association for Computational Linguistics: Human Language Technologies}, pages 4812--4829.

\bibitem[{Qian et~al.(2025)Qian, Liu, Wen, Bai, Liu, and Shao}]{qian2025demystifying}
Chen Qian, Dongrui Liu, Haochen Wen, Zhen Bai, Yong Liu, and Jing Shao. 2025.
\newblock Demystifying reasoning dynamics with mutual information: Thinking tokens are information peaks in llm reasoning.
\newblock \emph{arXiv preprint arXiv:2506.02867}.

\bibitem[{Shi et~al.(2024)Shi, Zhang, Sun, Gao, Ren, Chen, and Ren}]{DBLP:conf/acl/Shi00GRCR24}
Zhengliang Shi, Shuo Zhang, Weiwei Sun, Shen Gao, Pengjie Ren, Zhumin Chen, and Zhaochun Ren. 2024.
\newblock \href {https://doi.org/10.18653/V1/2024.ACL-LONG.397} {Generate-then-ground in retrieval-augmented generation for multi-hop question answering}.
\newblock In \emph{Proceedings of the 62nd Annual Meeting of the Association for Computational Linguistics (Volume 1: Long Papers), {ACL} 2024, Bangkok, Thailand, August 11-16, 2024}, pages 7339--7353. Association for Computational Linguistics.

\bibitem[{Simonyan et~al.(2014)Simonyan, Vedaldi, and Zisserman}]{DBLP:journals/corr/SimonyanVZ13}
Karen Simonyan, Andrea Vedaldi, and Andrew Zisserman. 2014.
\newblock \href {http://arxiv.org/abs/1312.6034} {Deep inside convolutional networks: Visualising image classification models and saliency maps}.
\newblock In \emph{2nd International Conference on Learning Representations, {ICLR} 2014, Banff, AB, Canada, April 14-16, 2014, Workshop Track Proceedings}.

\bibitem[{Tian et~al.(2024)Tian, Gan, Song, Zhang, and Zhang}]{DBLP:conf/acl/TianGSZ024}
Yuanhe Tian, Ruyi Gan, Yan Song, Jiaxing Zhang, and Yongdong Zhang. 2024.
\newblock \href {https://doi.org/10.18653/V1/2024.ACL-LONG.386} {Chimed-gpt: {A} chinese medical large language model with full training regime and better alignment to human preferences}.
\newblock In \emph{Proceedings of the 62nd Annual Meeting of the Association for Computational Linguistics (Volume 1: Long Papers), {ACL} 2024, Bangkok, Thailand, August 11-16, 2024}, pages 7156--7173. Association for Computational Linguistics.

\bibitem[{Todd et~al.(2024)Todd, Li, Sharma, Mueller, Wallace, and Bau}]{DBLP:conf/iclr/ToddLSMWB24}
Eric Todd, Millicent~L. Li, Arnab~Sen Sharma, Aaron Mueller, Byron~C. Wallace, and David Bau. 2024.
\newblock \href {https://openreview.net/forum?id=AwyxtyMwaG} {Function vectors in large language models}.
\newblock In \emph{The Twelfth International Conference on Learning Representations, {ICLR} 2024, Vienna, Austria, May 7-11, 2024}. OpenReview.net.

\bibitem[{Wang et~al.(2023)Wang, Li, Dai, Chen, Zhou, Meng, Zhou, and Sun}]{wang2023label}
Lean Wang, Lei Li, Damai Dai, Deli Chen, Hao Zhou, Fandong Meng, Jie Zhou, and Xu~Sun. 2023.
\newblock Label words are anchors: An information flow perspective for understanding in-context learning.
\newblock In \emph{Proceedings of the 2023 Conference on Empirical Methods in Natural Language Processing}, pages 9840--9855.

\bibitem[{Wu et~al.(2025)Wu, Wang, Xiao, Peng, and Fu}]{DBLP:conf/iclr/WuWX0F25}
Wenhao Wu, Yizhong Wang, Guangxuan Xiao, Hao Peng, and Yao Fu. 2025.
\newblock \href {https://openreview.net/forum?id=EytBpUGB1Z} {Retrieval head mechanistically explains long-context factuality}.
\newblock In \emph{The Thirteenth International Conference on Learning Representations, {ICLR} 2025, Singapore, April 24-28, 2025}. OpenReview.net.

\bibitem[{Xue et~al.(2025{\natexlab{a}})Xue, Mi, Zhu, Wang, Wang, Wang, Yu, Hu, and Wong}]{xue-etal-2025-ualign}
Boyang Xue, Fei Mi, Qi~Zhu, Hongru Wang, Rui Wang, Sheng Wang, Erxin Yu, Xuming Hu, and Kam-Fai Wong. 2025{\natexlab{a}}.
\newblock \href {https://doi.org/10.18653/v1/2025.acl-long.299} {{UA}lign: Leveraging uncertainty estimations for factuality alignment on large language models}.
\newblock In \emph{Proceedings of the 63rd Annual Meeting of the Association for Computational Linguistics (Volume 1: Long Papers)}, pages 6002--6024, Vienna, Austria. Association for Computational Linguistics.

\bibitem[{Xue et~al.(2025{\natexlab{b}})Xue, Greenewald, Mroueh, and Mirzasoleiman}]{xue2025verify}
Yihao Xue, Kristjan Greenewald, Youssef Mroueh, and Baharan Mirzasoleiman. 2025{\natexlab{b}}.
\newblock Verify when uncertain: Beyond self-consistency in black box hallucination detection.
\newblock \emph{arXiv preprint arXiv:2502.15845}.

\bibitem[{Yang et~al.(2025)Yang, Li, Yang, Zhang, Hui, Zheng, Yu, Gao, Huang, Lv, Zheng, Liu, Zhou, Huang, Hu, Ge, Wei, Lin, Tang, Yang, Tu, Zhang, Yang, Yang, Zhou, Zhou, Lin, Dang, Bao, Yang, Yu, Deng, Li, Xue, Li, Zhang, Wang, Zhu, Men, Gao, Liu, Luo, Li, Tang, Yin, Ren, Wang, Zhang, Ren, Fan, Su, Zhang, Zhang, Wan, Liu, Wang, Cui, Zhang, Zhou, and Qiu}]{yang2025qwen3technicalreport}
An~Yang, Anfeng Li, Baosong Yang, Beichen Zhang, Binyuan Hui, Bo~Zheng, Bowen Yu, Chang Gao, Chengen Huang, Chenxu Lv, Chujie Zheng, Dayiheng Liu, Fan Zhou, Fei Huang, Feng Hu, Hao Ge, Haoran Wei, Huan Lin, Jialong Tang, and 41 others. 2025.
\newblock \href {https://arxiv.org/abs/2505.09388} {Qwen3 technical report}.
\newblock \emph{Preprint}, arXiv:2505.09388.

\bibitem[{Yang et~al.(2018)Yang, Qi, Zhang, Bengio, Cohen, Salakhutdinov, and Manning}]{yang-etal-2018-hotpotqa}
Zhilin Yang, Peng Qi, Saizheng Zhang, Yoshua Bengio, William Cohen, Ruslan Salakhutdinov, and Christopher~D. Manning. 2018.
\newblock \href {https://doi.org/10.18653/v1/D18-1259} {{H}otpot{QA}: A dataset for diverse, explainable multi-hop question answering}.
\newblock In \emph{Proceedings of the 2018 Conference on Empirical Methods in Natural Language Processing}, pages 2369--2380, Brussels, Belgium. Association for Computational Linguistics.

\bibitem[{Zhang et~al.(2025)Zhang, Li, Cui, Cai, Liu, Fu, Huang, Zhao, Zhang, Chen et~al.}]{zhang2025siren}
Yue Zhang, Yafu Li, Leyang Cui, Deng Cai, Lemao Liu, Tingchen Fu, Xinting Huang, Enbo Zhao, Yu~Zhang, Yulong Chen, and 1 others. 2025.
\newblock Siren’s song in the ai ocean: A survey on hallucination in large language models.
\newblock \emph{Computational Linguistics}, pages 1--46.

\end{thebibliography}

\clearpage
\appendix

\section{LLM Usage}

In this work, we employ LLMs solely for language refinement to enhance clarity and explanatory quality. All content has been carefully verified for factual accuracy, and the authors take full responsibility for the entire manuscript. The core ideas, experimental design, and methodological framework are conceived and developed independently by the authors, without the use of LLMs.

\section{Implementation Details}
\label{sec:appendix all Implementation Details}

\subsection{Identifying Exact Question and Answer Tokens}
\label{sec:appendix Identifying Exact Question and Answer Tokens}

To locate the exact question and answer tokens within a QA pair, we prompt GPT-4o (version \texttt{gpt-4o\_2024-11-20}) to identify the precise positions of the core frame elements. The instruction templates are presented in Tables \ref{tab:appendix_prompt_exact_question} and \ref{tab:appendix_prompt_exact_answer}. A token is considered an \emph{exact question} or \emph{exact answer} if and only if it constitutes a valid substring of the corresponding question or answer. To mitigate potential biases, each example is prompted at most five times, and only successfully extracted instances are retained for downstream analysis. 
Prior work \citep{DBLP:conf/iclr/OrgadTGRSKB25} has shown that LLMs can accurately identify exact answer tokens, typically achieving over 95\% accuracy. 
In addition, we manually verified GPT-4o’s identification quality in our setting. Specifically, it achieves 99.92\%, 95.83\%, and 96.62\% accuracy on exact subject tokens, exact property tokens, and exact answer tokens, respectively. Furthermore, we also explore alternative configurations without the use of exact tokens to ensure the robustness of our findings (see Section \ref{sec:appendix Probing Implementation Details}).

\subsection{Probing Implementation Details}
\label{sec:appendix Probing Implementation Details}

We investigate multiple probing configurations. 
For token selection, we consider three types of tokens: 
(1) the final token of the answer, which is the most commonly adopted choice in prior work due to its global receptive field under attention \citep{DBLP:conf/iclr/0026L0GWTFY24}; 
(2) the token immediately preceding the exact answer span; and 
(3) the final token within the exact answer span. 
For activation extraction, we obtain representations from either 
(1) the output of each attention sublayer or 
(2) the output of the final multi-layer perceptron (MLP) in each transformer layer. 
Across all configurations, our experimental results exhibit consistent trends, indicating that the observed findings are robust to these design choices.
For the probing classifier, we follow standard practice \citep{DBLP:conf/iclr/0026L0GWTFY24,DBLP:conf/iclr/OrgadTGRSKB25} and employ a logistic regression model implemented in \texttt{scikit-learn}.

\subsection{Models}
\label{sec:appendix Models}

Our analysis covers a diverse collection of 12 LLMs that vary in both scale and architectural design. Specifically, we consider three categories: (1) \emph{base models}, including Llama-3.2-1B \citep{grattafiori2024llama3herdmodels}, Llama-3.2-3B, Llama-3-8B, Llama-3-70B, Mistral-7B-v0.1 \citep{jiang2023mistral7b}, and Mistral-7B-v0.3; (2) \emph{instruction-tuned models}, including Llama-3.2-3B-Instruct, Llama-3-8B-Instruct, Mistral-7B-Instruct-v0.1, and Mistral-7B-Instruct-v0.3; and (3) \emph{reasoning-oriented models}, namely Qwen3-8B \citep{yang2025qwen3technicalreport} and Qwen3-32B.

\subsection{Datasets}
\label{sec:appendix Datasets}

We consider four widely used question-answering datasets: PopQA \citep{mallen-etal-2023-trust}, TriviaQA \citep{joshi-etal-2017-triviaqa}, HotpotQA \citep{yang-etal-2018-hotpotqa}, and Natural Questions \citep{kwiatkowski-etal-2019-natural}.  

\textbf{PopQA} is an open-domain question-answering dataset that emphasizes entity-centric factual knowledge with a long-tail distribution. It is designed to probe LLMs' ability to memorize less frequent facts, highlighting limitations in parametric knowledge.

\textbf{TriviaQA} is a reading comprehension dataset constructed by pairing trivia questions authored independently of evidence documents. The questions are often complex, requiring multi-sentence reasoning, and exhibit substantial lexical and syntactic variability.

\textbf{HotpotQA} is a challenging multi-hop question-answering dataset that requires reasoning. It includes diverse question types—span extraction, yes/no, and novel comparison questions—along with sentence-level supporting fact annotations, promoting the development of explainable QA systems.  

\textbf{Natural Questions} is an open-domain dataset consisting of real, anonymized questions from Google search queries. Each question is annotated with both a long answer (paragraph or section) and a short answer (span or yes/no), or marked as null when no answer is available.
Due to computational constraints, we randomly sample 2,000 training samples and 2,000 test samples for each dataset.

\subsection{Implementation Details of Baselines}
\label{sec:appendix Baseline Implementation Details}

In our experiments regarding applications, we compare our proposed methods against several internal-based baselines for hallucination detection. These baselines leverage the LLM's internal signals, such as output probabilities, logits, and hidden representations, without relying on external resources. Below, we detail the implementation of each baseline.

\paragraph{P(True)}
P(True) \citep{Kadavath2022LanguageM} exploits the LLM's self-awareness of its knowledge boundaries by prompting the model to assess the correctness of its own generated answer. Specifically, for each question-answer pair \((q_i, \hat{y}^f_i)\), we prompt the LLM with a template that asks it to evaluate whether its answer is factually correct. Following \citet{Kadavath2022LanguageM}, the prompt template is shown in Table~\ref{tab:appendix_prompt_ptrue}.

\begin{table}[!htb]
\centering
\begin{tabular}{p{0.9\columnwidth}}
\toprule
Question: \{Here is the question\}\\
Possible answer: \{Here is the answer\}\\
Is the possible answer:\\
(A) True\\
(B) False\\
The possible answer is:\\
\bottomrule
\end{tabular}
\caption{Prompt template used for the P(True) baseline.}
\label{tab:appendix_prompt_ptrue}
\end{table}

\paragraph{Logits-based Baselines}

The logits-based baselines utilize the raw logits produced by the LLM during the generation of the exact answer tokens. Let \(\hat{y}^f_{i,E_A} = [t_1, t_2, \dots, t_m]\) represent the sequence of exact answer tokens for a given question-answer pair, where \(m\) is the number of exact answer tokens. For each token \(t_j\) (where \(j \in \{1, \dots, m\}\)), the LLM produces a logit vector \(L_j \in \mathbb{R}^V\), where \(V\) is the vocabulary size, and the logit for the generated token \(t_j\) is denoted \(L_j[t_j]\). The logits-based metrics are defined as follows:
\begin{itemize}[left=0cm]
    \item \textbf{Logits-mean}: The average of the logits across all exact answer tokens:
    \begin{equation}
        \text{Logits-mean} = \frac{1}{m} \sum_{j=1}^m L_j[t_j]
    \end{equation}
    \item \textbf{Logits-max}: The maximum logit value among the exact answer tokens:
    \begin{equation}
    \text{Logits-max} = \max_{j \in \{1, \dots, m\}} L_j[t_j]
    \end{equation}
    \item \textbf{Logits-min}: The minimum logit value among the exact answer tokens:
    \begin{equation}
    \text{Logits-min} = \min_{j \in \{1, \dots, m\}} L_j[t_j]
    \end{equation}
\end{itemize}

These metrics serve as proxies for the model’s confidence in the generated answer, with lower logit values potentially indicating uncertainty or hallucination. 

\paragraph{Scores-based Baselines}
The scores-based baselines are derived from the softmax probabilities of the exact answer tokens. Using the same notation as above, for each exact answer token \(t_j\), the softmax probability is computed as:
\begin{equation}
p_j[t_j] = \frac{\exp(L_j[t_j])}{\sum_{k=1}^V \exp(L_j[k])}
\end{equation}
where \(L_j[k]\) is the logit for the \(k\)-th token in the vocabulary. The scores-based metrics are defined as follows:
\begin{itemize}[left=0cm]
    \item \textbf{Scores-mean}: The average of the softmax probabilities across all exact answer tokens:
    \begin{equation}
    \text{Scores-mean} = \frac{1}{m} \sum_{j=1}^m p_j[t_j]
    \end{equation}
    \item \textbf{Scores-max}: The maximum softmax probability among the exact answer tokens:
    \begin{equation}
    \text{Scores-max} = \max_{j \in \{1, \dots, m\}} p_j[t_j]
    \end{equation}
    \item \textbf{Scores-min}: The minimum softmax probability among the exact answer tokens:
    \begin{equation}
    \text{Scores-min} = \min_{j \in \{1, \dots, m\}} p_j[t_j]
    \end{equation}
\end{itemize}

These probabilities provide a normalized measure of the model’s confidence, bounded between 0 and 1, with lower values potentially indicating a higher likelihood of hallucination. 

\paragraph{Probing Baseline}
\label{sec:appendix baseline Implementation Details}
The probing baseline follows the standard approach described in \citet{DBLP:conf/iclr/0026L0GWTFY24,DBLP:conf/iclr/OrgadTGRSKB25}. A linear classifier is trained on the hidden representations of the last exact answer token from the best-performing layer. 
The training and evaluation data for the probing classifier are constructed following the procedure described in Appendix \ref{sec:appendix Datasets}.
The classifier is implemented using \texttt{scikit-learn} with default hyperparameters, consistent with the probing setup described in Appendix \ref{sec:appendix Probing Implementation Details}. The probing baseline serves as a direct comparison to our proposed applications, as it relies on the same type of internal signals but does not account for the heterogeneity of truthfulness encoding pathways.

\subsection{Implementation Details of MoP and PR}
\label{sec:appendix Our Implementation Details}

\paragraph{Model Backbone and Hidden Representations}

All experiments use the same base LLM as in the main paper.  
Hidden representations $\mathbf{h}^{l^*}(x)$ are extracted from the best-performing layer $l^*$ determined on a held-out validation split.  

\paragraph{Mixture-of-Probes (MoP)}

Similar to Appendix \ref{sec:appendix baseline Implementation Details}, the two expert probes $p_Q$ and $p_A$ are implemented using \texttt{scikit-learn} with default hyperparameters, consistent with the probing setup described in Appendix \ref{sec:appendix Probing Implementation Details}.
The gating network is directly from the self-awareness probe described in Section \ref{sec:Intrinsic Awareness of Encoding Pathway Selection}.
The training and evaluation data for the probing classifier are the same as Appendix \ref{sec:appendix baseline Implementation Details}.
The proposed MoP framework requires no additional retraining: 
we directly combine the two expert probes with the pathway-discrimination classifier described in Section \ref{sec:Intrinsic Awareness of Encoding Pathway Selection} and perform inference without further parameter updates. 

\paragraph{Pathway Reweighting (PR)}

The training and evaluation data used for the probing classifier are identical to those described in Appendix \ref{sec:appendix baseline Implementation Details}.
For each Transformer layer $l \leq l^*$, we introduce two learnable scalars $\alpha_Q^{l}$ and $\alpha_A^{l}$ for every attention head.  
These parameters, together with the probe parameters, are optimized using the Adam optimizer with a learning rate of $1 \times 10^{-2}$, $\beta_1 = 0.9$, and $\beta_2 = 0.999$. Training is conducted with a batch size of 512 for 10 epochs, while all original LLM parameters remain frozen.

\begin{table*}[!htb]
\centering
\begin{tabular}{p{0.95\textwidth}}
\toprule
You are given a factual open-domain question-answer pair.\\
Your task is to identify:\\
\\
1. Core Entity (c) - the known specific entity in the question that the answer is about (a person, place, organization, or other proper noun).\\
2. Relation (r) - the minimal phrase in the question that expresses what is being asked about the core entity, using only words from the question.\\
\\
Guidelines:\\
\\
The core entity must be a concrete, known entity mentioned in the question, not a general category.\\
If multiple entities appear, choose the one most central to the question—the entity the answer primarily concerns.\\
The relation should be the smallest meaningful span that directly connects the core entity to the answer.\\
Use only words from the question; do not paraphrase or add new words.\\
Exclude extra context, modifiers, or descriptive phrases that are not essential to defining the relationship.\\
For complex questions with long modifiers or embedded clauses, focus on the words that directly express the property, action, or attribute of the core entity relevant to the answer.\\
If you cannot confidently identify the core entity or the relation, output NO ANSWER.\\
\\
Output format:\\
Core Entity: exact text\\
Relation: exact text\\
\\
Example 1\\
Question: Who was the director of Finale?\\
Answer: Ken Kwapis\\
Core Entity: Finale\\
Relation: director\\
\\
Example 2\\
Question: What film, in production between 2007 and 2009, is directed by James Cameron ("Titanic")?\\
Answer: Avatāra\\
Core Entity: James Cameron\\
Relation: film directed by\\
\\
Example 3\\
Question: Which novel, written in 1925 and often cited as a classic of American literature, was authored by F. Scott Fitzgerald?\\
Answer: The Great Gatsby\\
Core Entity: F. Scott Fitzgerald\\
Relation: novel authored by\\
\\
Question: \{Here is the question\}\\
Answer: \{Hese is the answer\}\\
\bottomrule
\end{tabular}
\caption{Prompt template used to locate the exact question tokens.}
\label{tab:appendix_prompt_exact_question}
\end{table*}

\begin{table*}[!htb]
\centering
\begin{tabular}{p{0.95\textwidth}}
\toprule
You are given a factual open-domain question-answer pair.\\
Your task is to extract the short exact answer from the long answer.\\
\\
Guidelines:\\
\\
Retain only the relevant tokens from the answer.\\
Use only words from the answer; do not paraphrase or add new words.\\
Extract the short answer regardless of whether the long answer is correct or incorrect.\\
If the short answer cannot be confidently determined, or if the long answer does not address the question, output NO ANSWER.\\
\\
Example 1\\
Q: Which musical featured the song The Street Where You Live?\\
A: The song "The Street Where You Live" is from the Lerner and Loewe musical "My Fair Lady." It is one of the most famous songs from the show, and it is sung by Professor Henry Higgins as he reflects on the transformation of Eliza Doolittle and the memories they have shared together.\\
Exact answer: My Fair Lady\\
\\
Example 2\\
Q: Which Swedish actress won the Best Supporting Actress Oscar for Murder on the Orient Express?\\
A: I'm glad you asked about a Swedish actress who won an Oscar for "Murder on the Orient Express," but I must clarify that there seems to be a misunderstanding here. No Swedish actress has won an Oscar for Best Supporting Actress for that film. The 1974 "Murder on the Orient Express" was an American production, and the cast was predominantly British and American. If you have any other questions or if there's another\\
Exact answer: NO ANSWER\\
\\
Example 3\\
Q: Who invented the telephone?\\
A: The telephone was invented by Thomas Edison in the 19th century.\\
Exact answer: Thomas Edison\\
\\
Q: \{Here is the question\}\\
A: \{Here is the answer\}\\
Exact answer:\\
\bottomrule
\end{tabular}
\caption{Prompt template used to locate the exact answer tokens.}
\label{tab:appendix_prompt_exact_answer}
\end{table*}

\begin{table*}[!htb]
\centering
\begin{tabular}{p{0.95\textwidth}}
\toprule
You are given one factual question. Interpret it literally and think carefully.\\
Your task is to decide whether you can answer it correctly with high confidence based only on your internal knowledge (no tools or web). If yes, output exactly: YES. If not or uncertain, output exactly: NO. You should output one word only.\\
\\
Question: \{Here is the question\}\\
Your Output:\\
\bottomrule
\end{tabular}
\caption{Prompt template used to obtain the i-don't-know rate.}
\label{tab:appendix_prompt_idk}
\end{table*}

\clearpage
\onecolumn
\section{Attention Knockout}
\label{sec:appendix_attn_knockout}

\begin{minipage}{\textwidth}
    \centering

    \includegraphics[width=\textwidth]{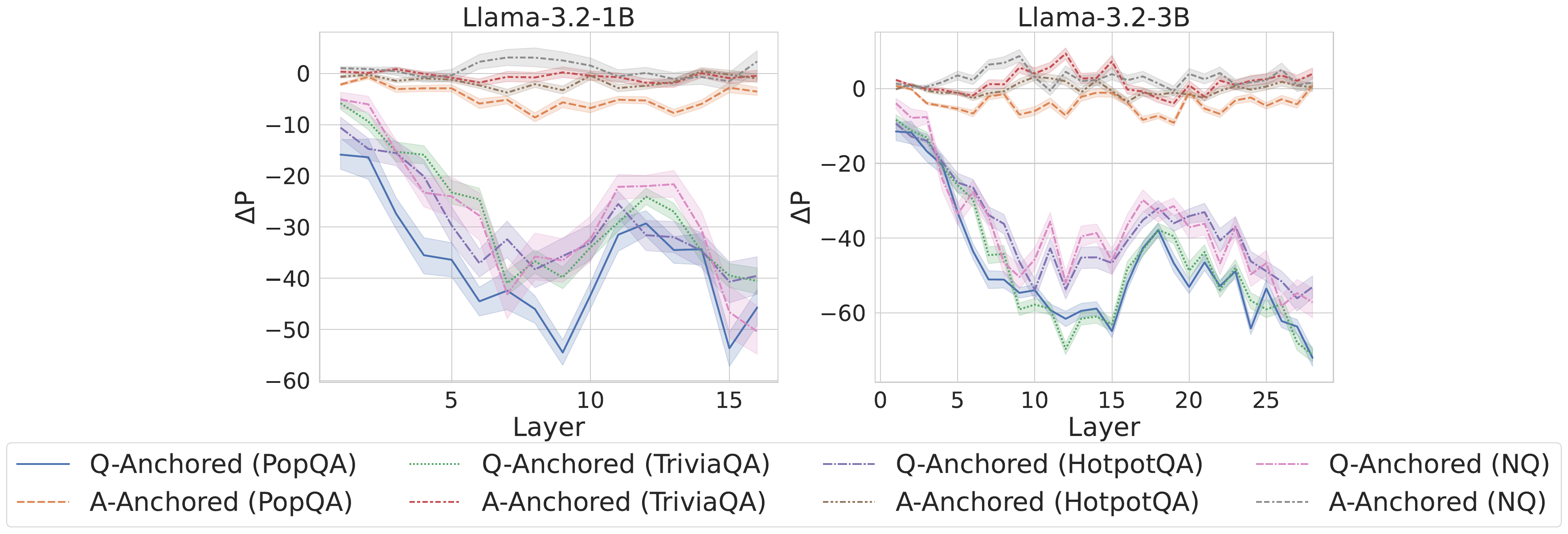}

    \vspace{4ex}

    \includegraphics[width=\textwidth]{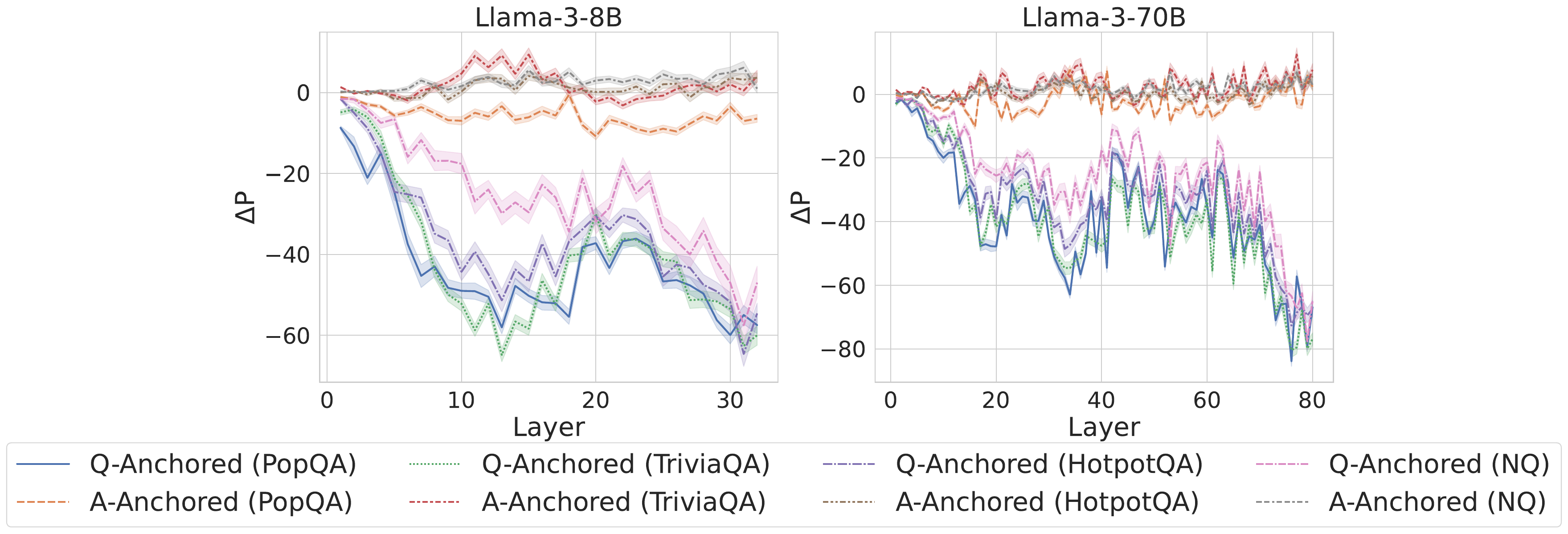}

    \vspace{4ex}

    \includegraphics[width=\textwidth]{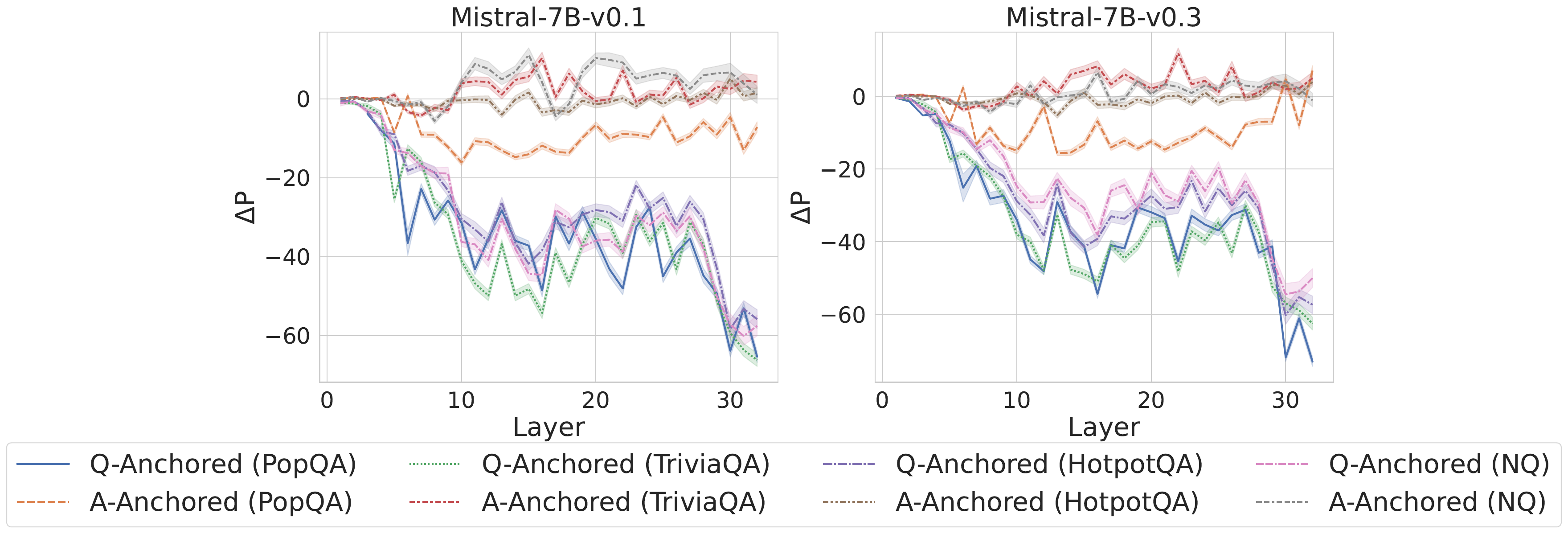}

    \vspace{2ex}

    \captionof{figure}{$\Delta \mathrm{P}$ under attention knockout, probing attention activations of the final token.}
    \label{fig:appendix_attention_knockout_base_attnact_-1}
\end{minipage}

\begin{figure*}[!htb]
\centering
\includegraphics[width=\textwidth]{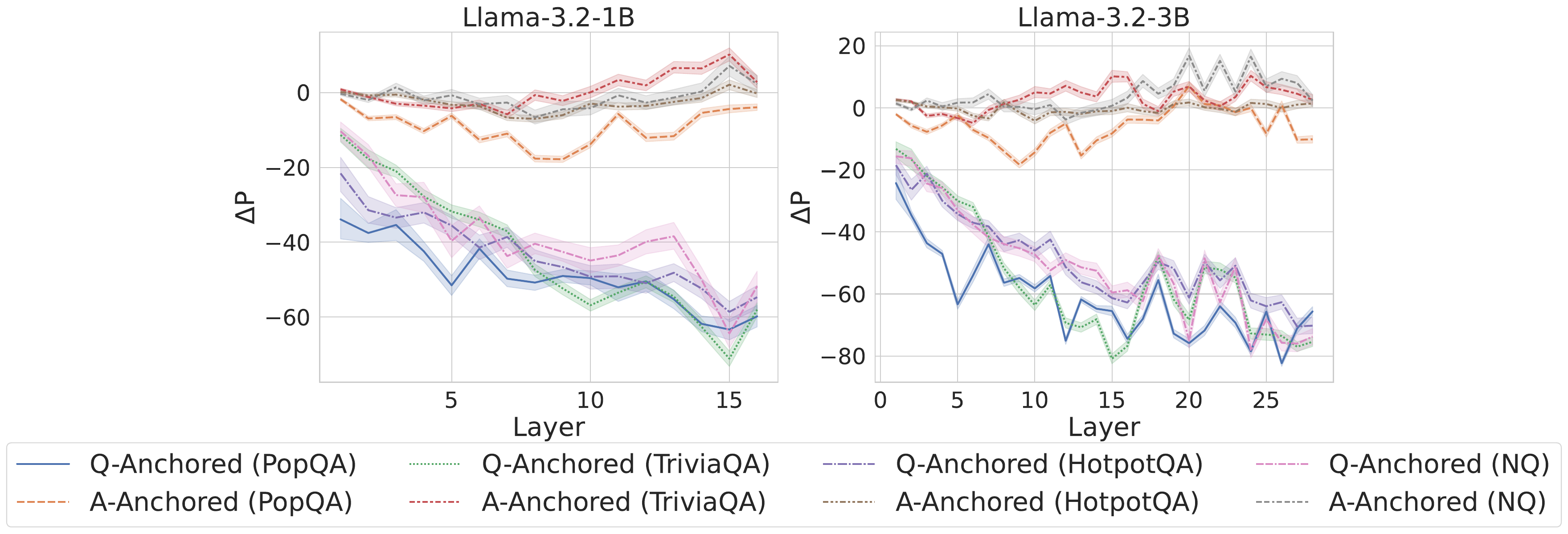}
\\[5ex]
\includegraphics[width=\textwidth]{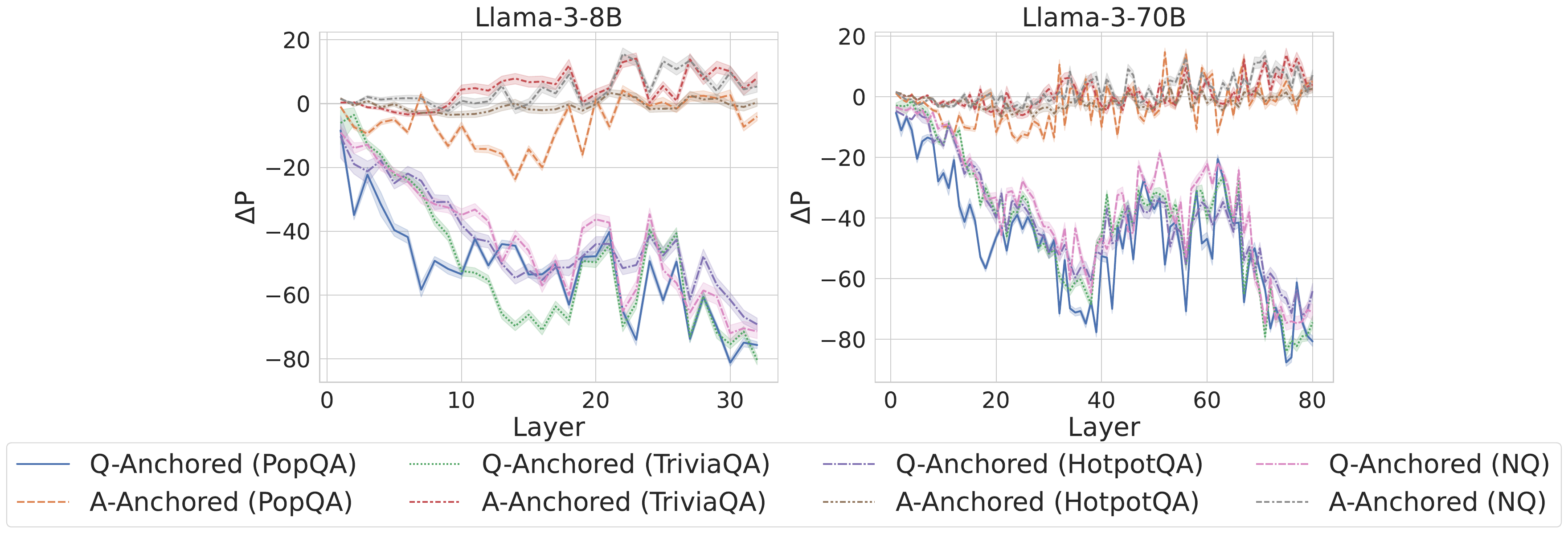}
\\[5ex]
\includegraphics[width=\textwidth]{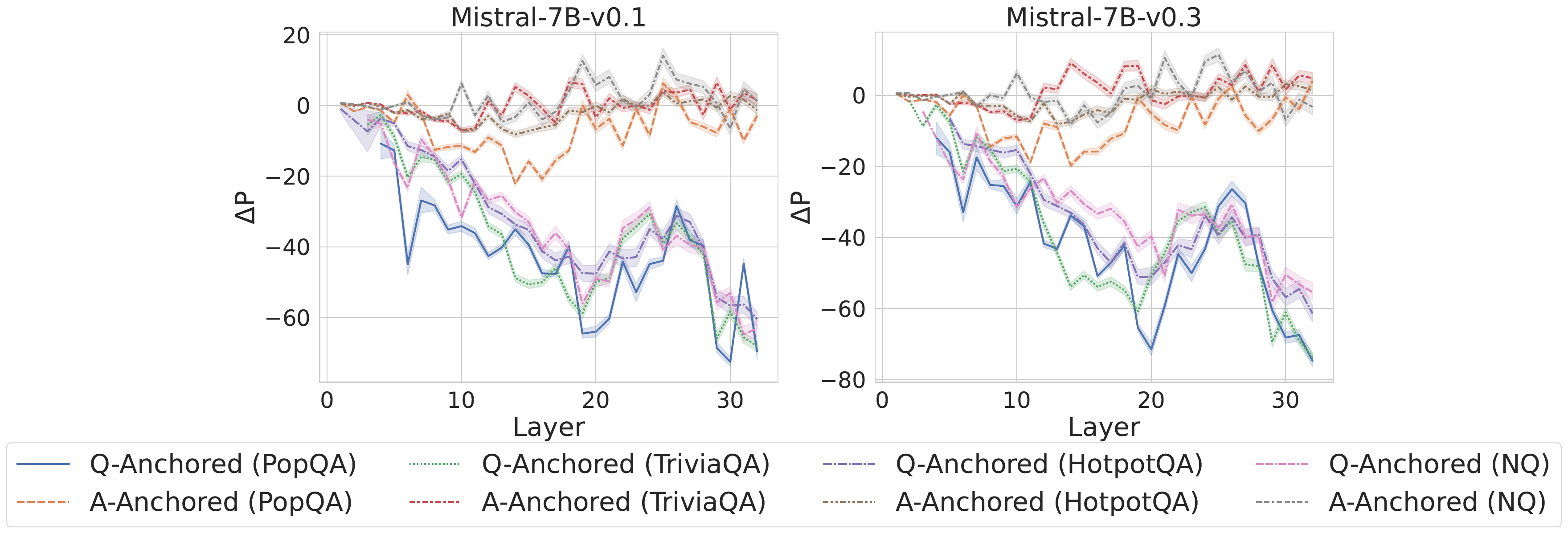}
\caption{$\Delta \mathrm{P}$ under attention knockout, probing attention activations of the token immediately preceding the exact answer tokens.}
\label{fig:appendix_attention_knockout_base_attnact_beforefirst}
\end{figure*}

\begin{figure*}[!htb]
\centering
\includegraphics[width=\textwidth]{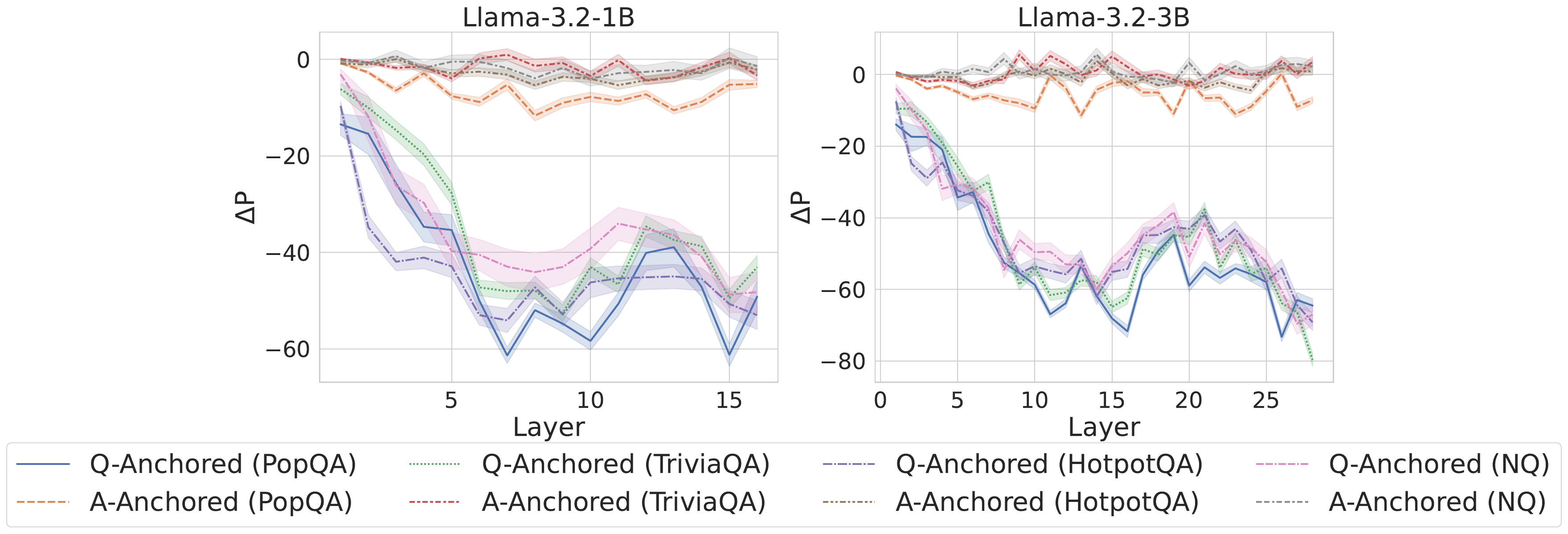}
\\[5ex]
\includegraphics[width=\textwidth]{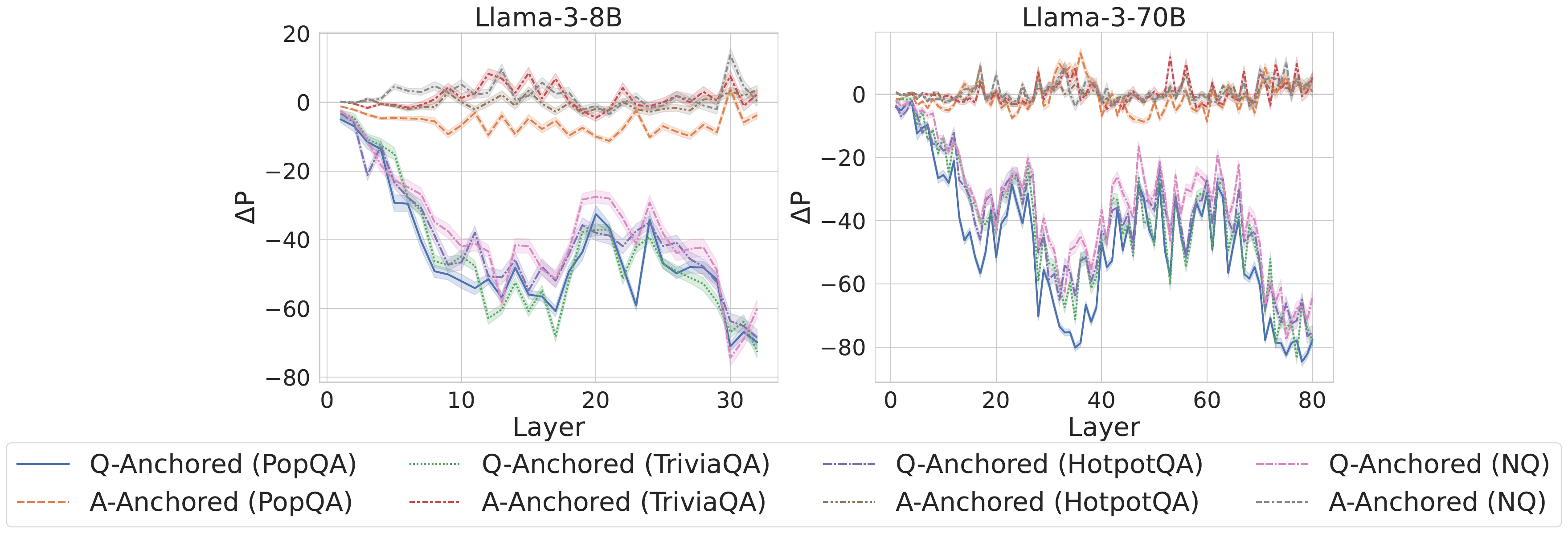}
\\[5ex]
\includegraphics[width=\textwidth]{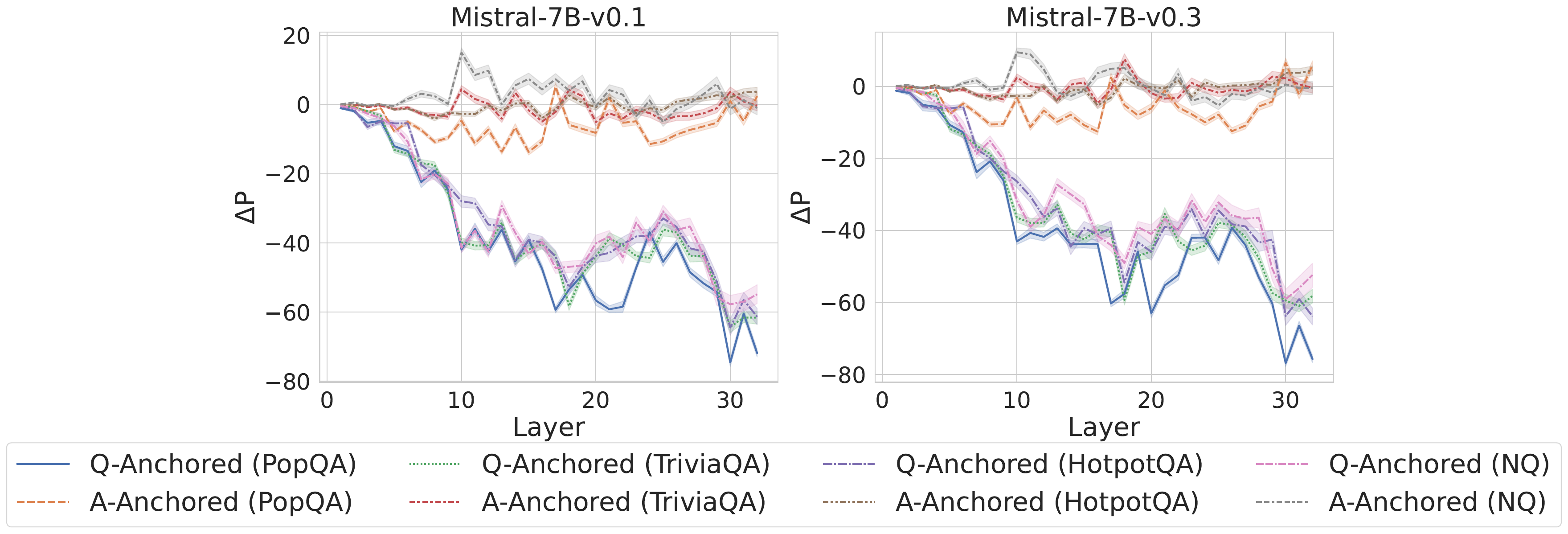}
\caption{$\Delta \mathrm{P}$ under attention knockout, probing attention activations of the last exact answer token.}
\label{fig:appendix_attention_knockout_base_attnact_exactans}
\end{figure*}

\begin{figure*}[!htb]
\centering
\includegraphics[width=\textwidth]{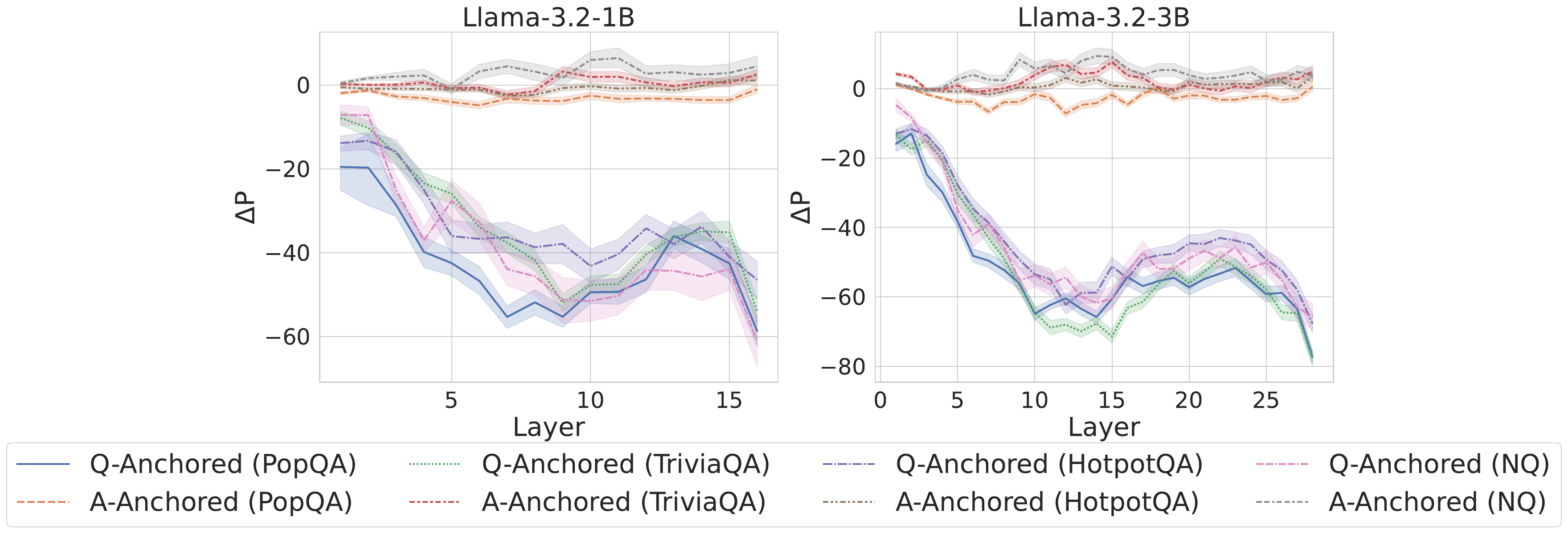}
\\[5ex]
\includegraphics[width=\textwidth]{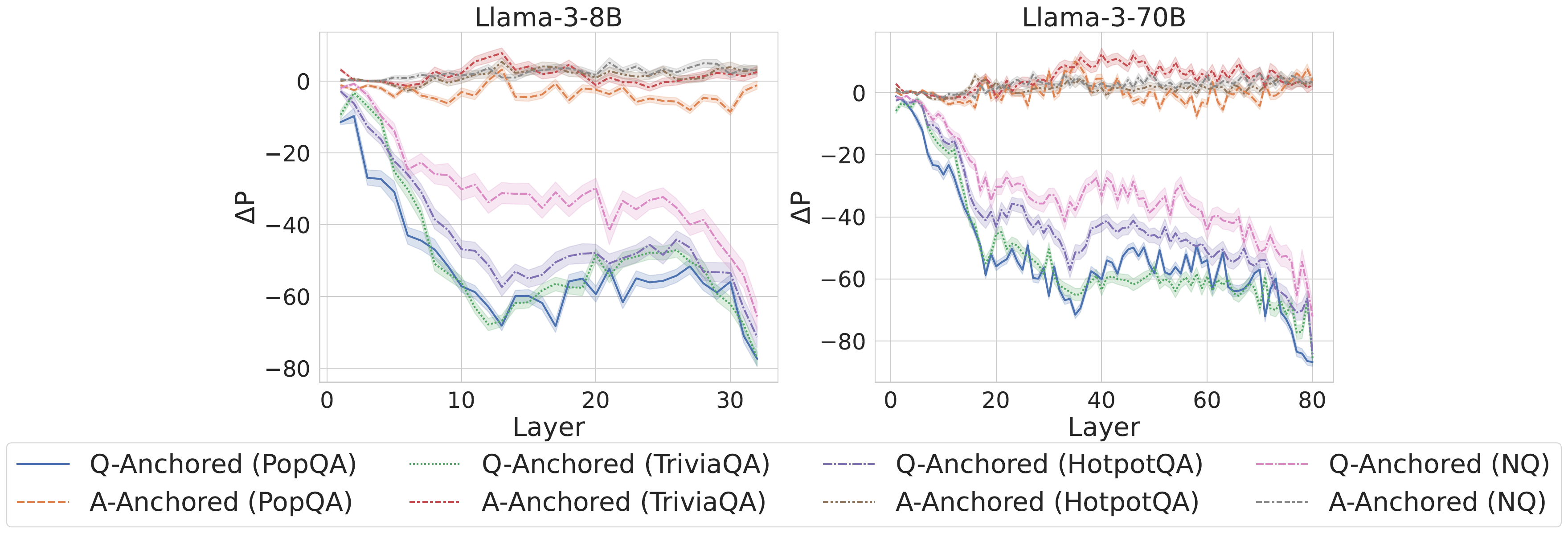}
\\[5ex]
\includegraphics[width=\textwidth]{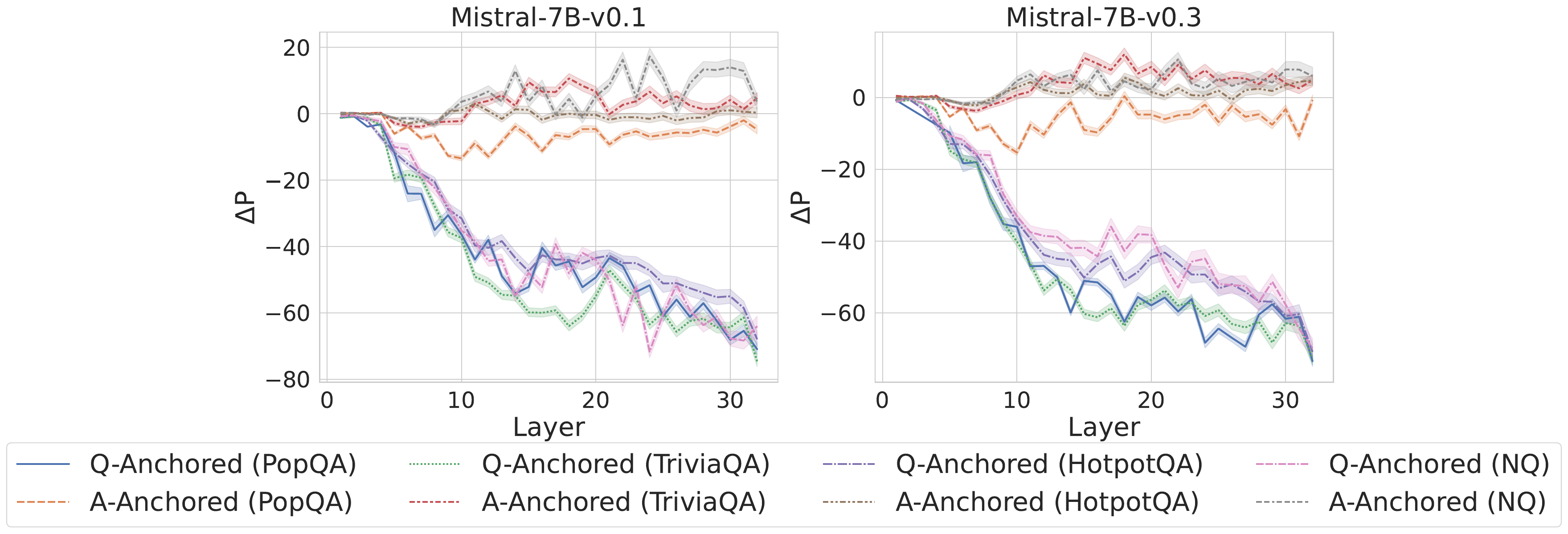}
\caption{$\Delta \mathrm{P}$ under attention knockout, probing mlp activations of the final token.}
\label{fig:appendix_attention_knockout_base_mlpact_-1}
\end{figure*}

\begin{figure*}[!htb]
\centering
\includegraphics[width=\textwidth]{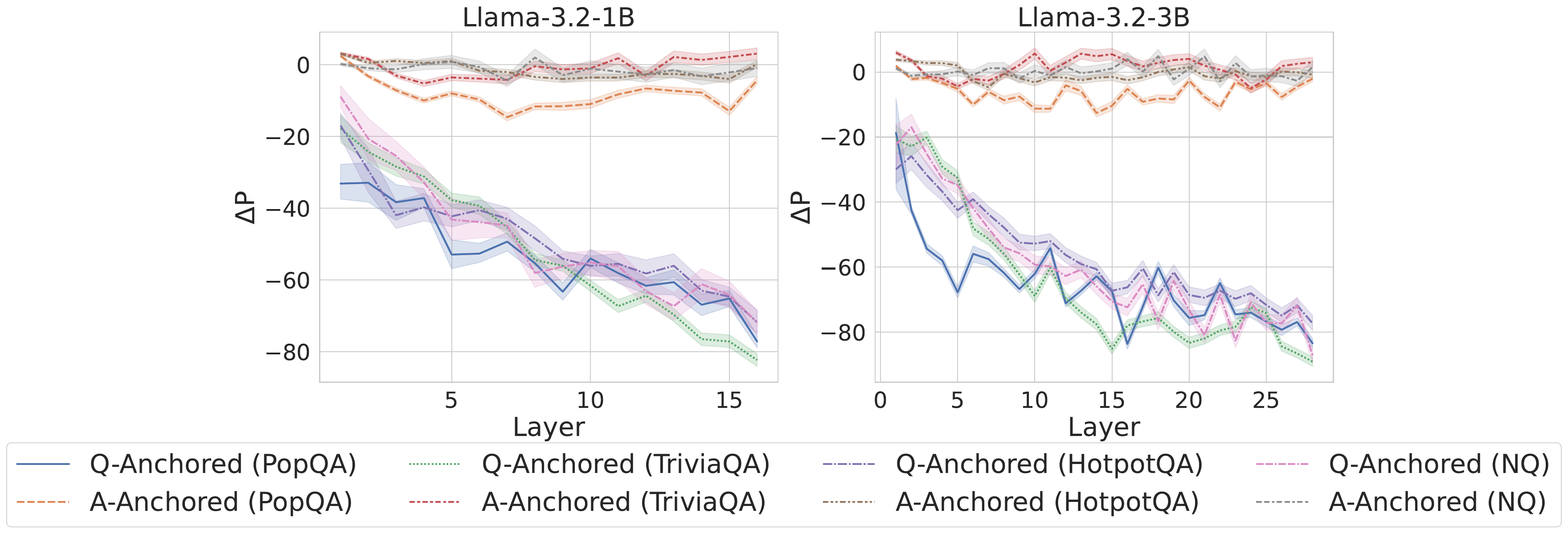}
\\[5ex]
\includegraphics[width=\textwidth]{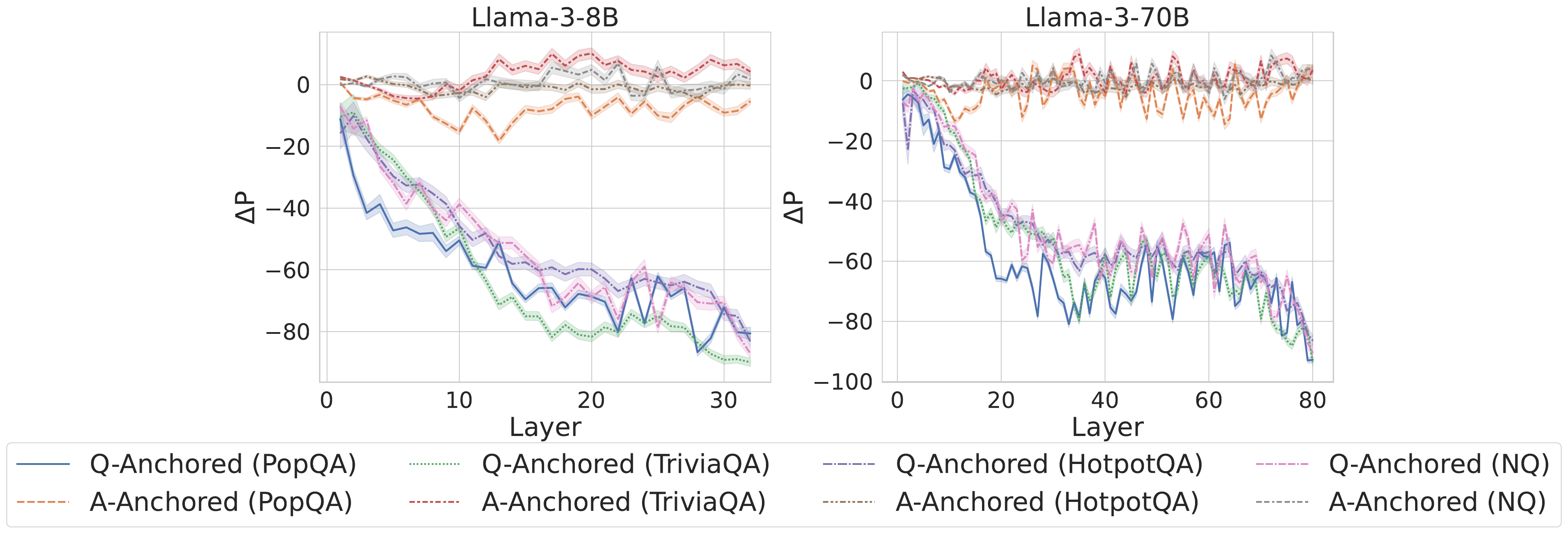}
\\[5ex]
\includegraphics[width=\textwidth]{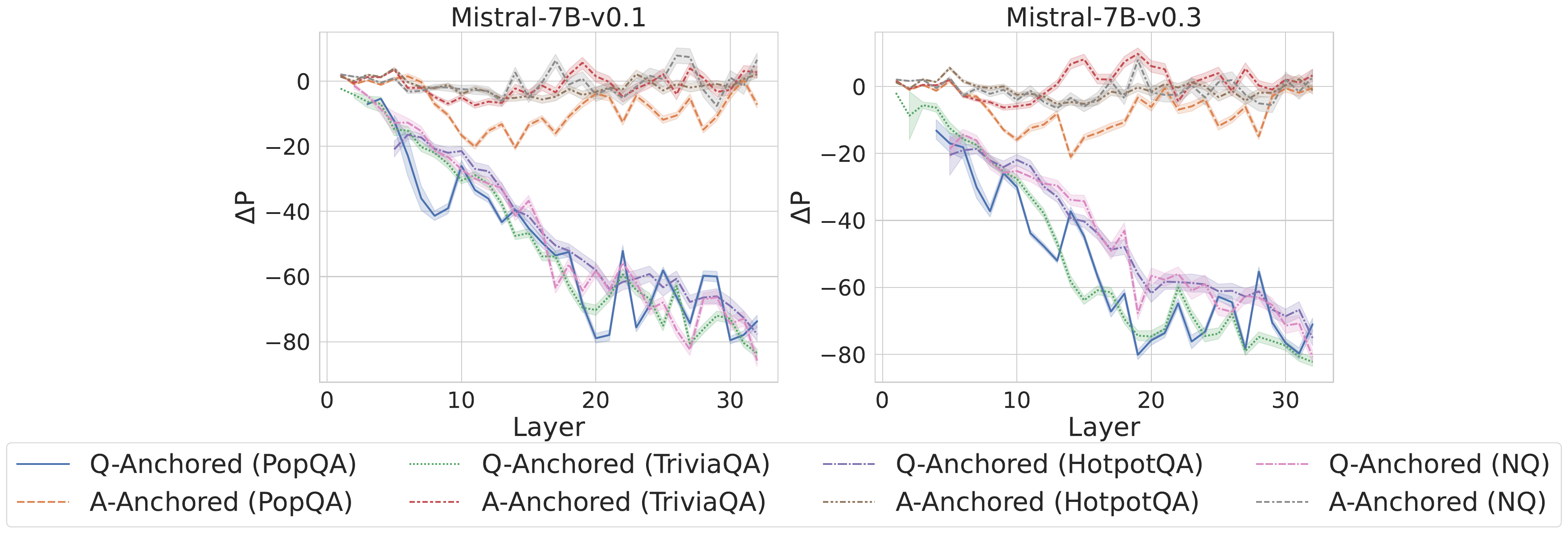}
\caption{$\Delta \mathrm{P}$ under attention knockout, probing mlp activations of the token immediately preceding the exact answer tokens.}
\label{fig:appendix_attention_knockout_base_mlpact_beforefirst}
\end{figure*}

\begin{figure*}[!htb]
\centering
\includegraphics[width=\textwidth]{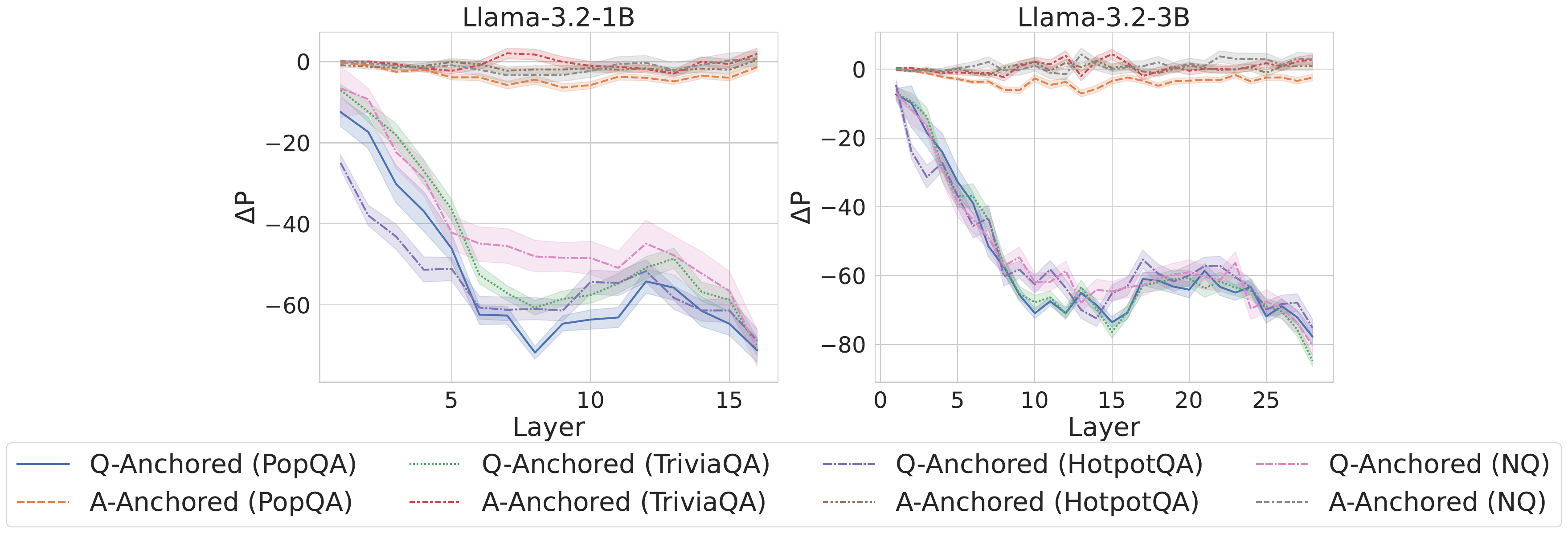}
\\[5ex]
\includegraphics[width=\textwidth]{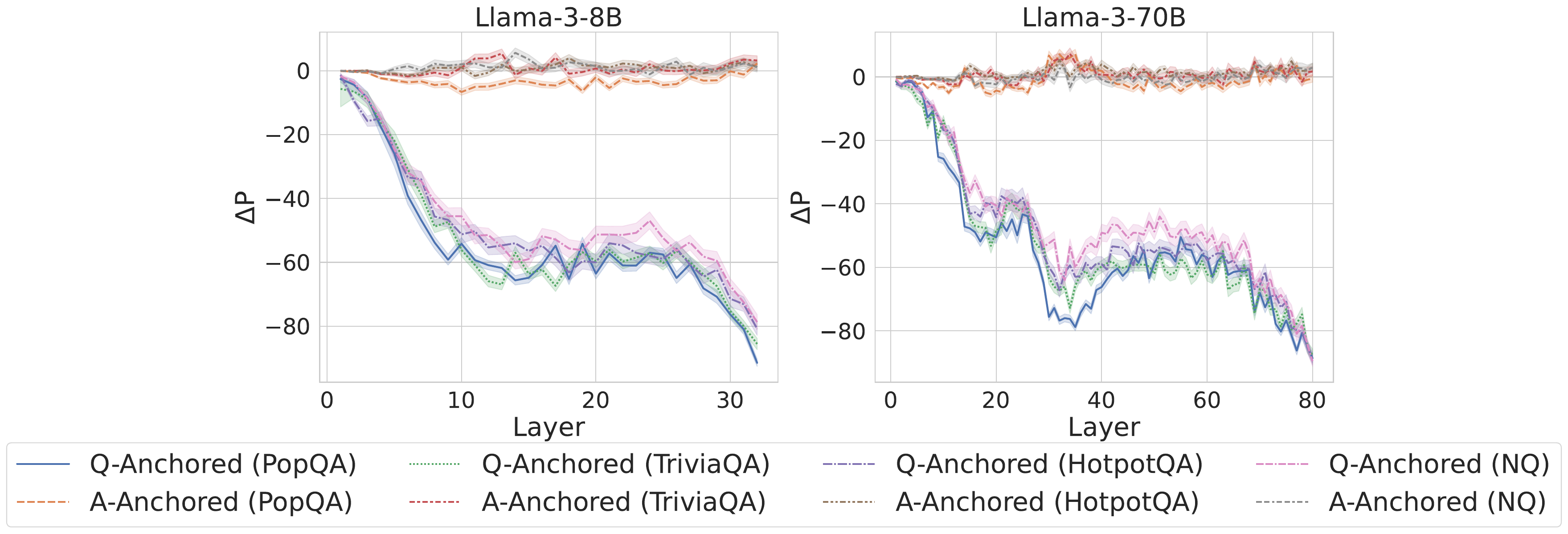}
\\[5ex]
\includegraphics[width=\textwidth]{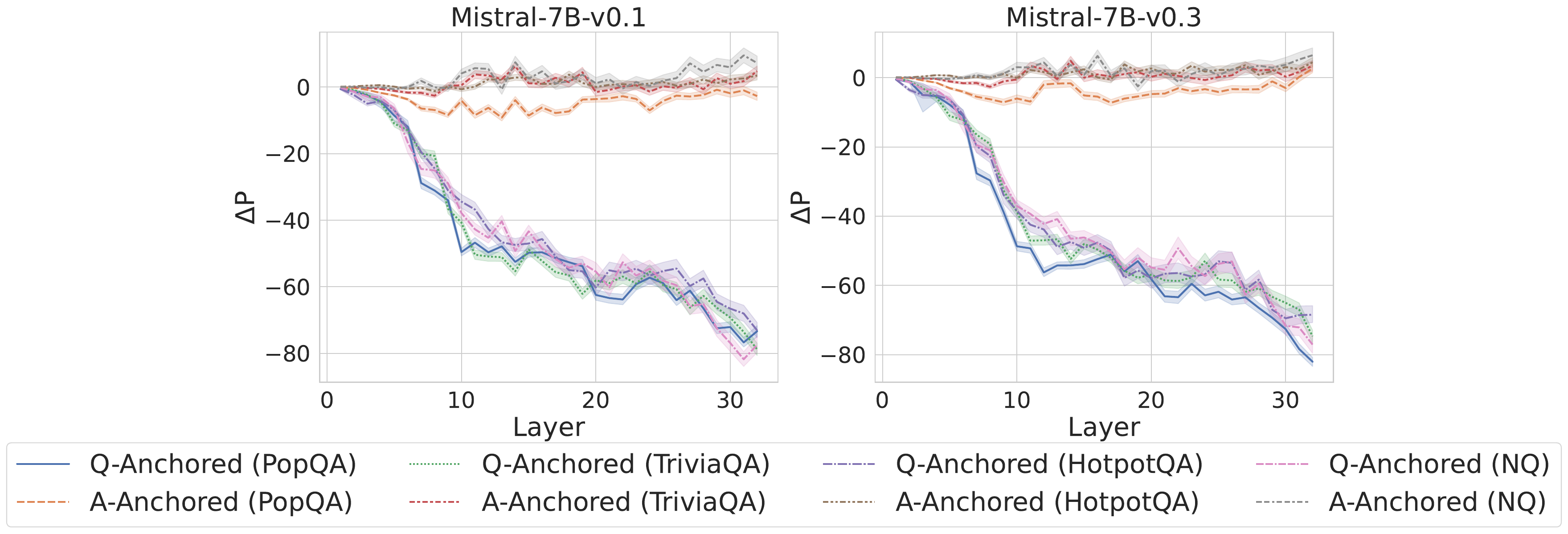}
\caption{$\Delta \mathrm{P}$ under attention knockout, probing mlp activations of the last exact answer token.}
\label{fig:appendix_attention_knockout_base_mlpact_exactans}
\end{figure*}

\begin{figure*}[!htb]
\centering
\includegraphics[width=\textwidth]{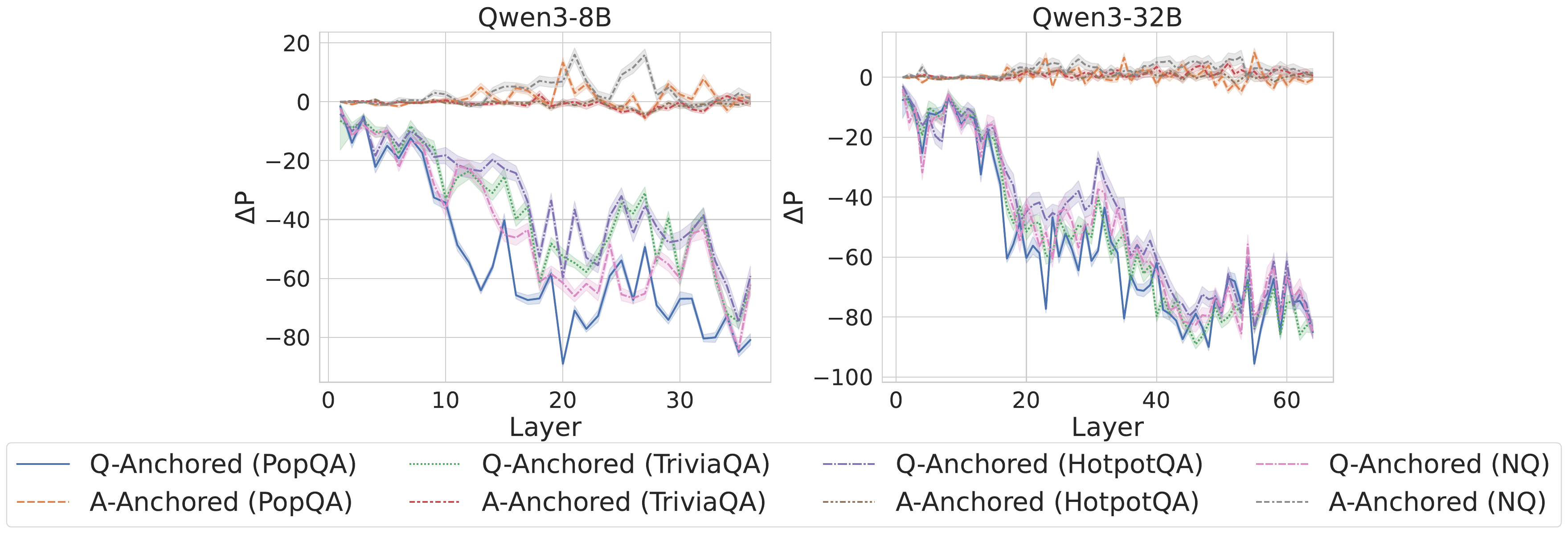}
\\[5ex]
\includegraphics[width=\textwidth]{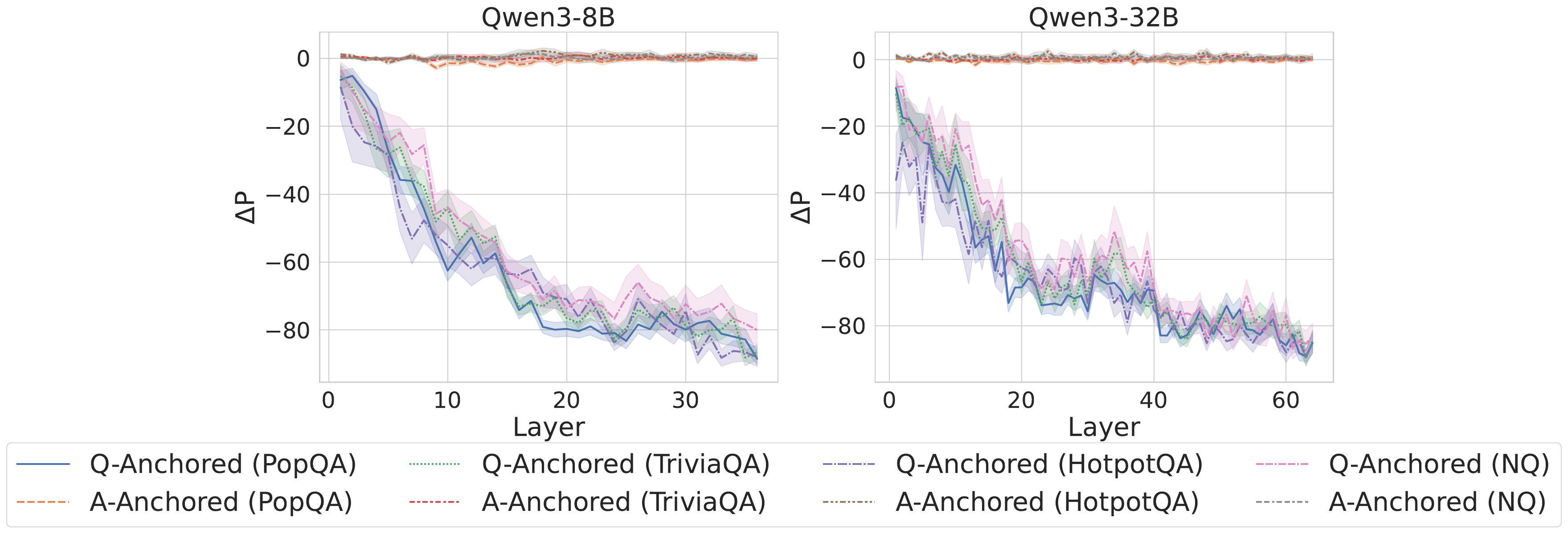}
\\[5ex]
\includegraphics[width=\textwidth]{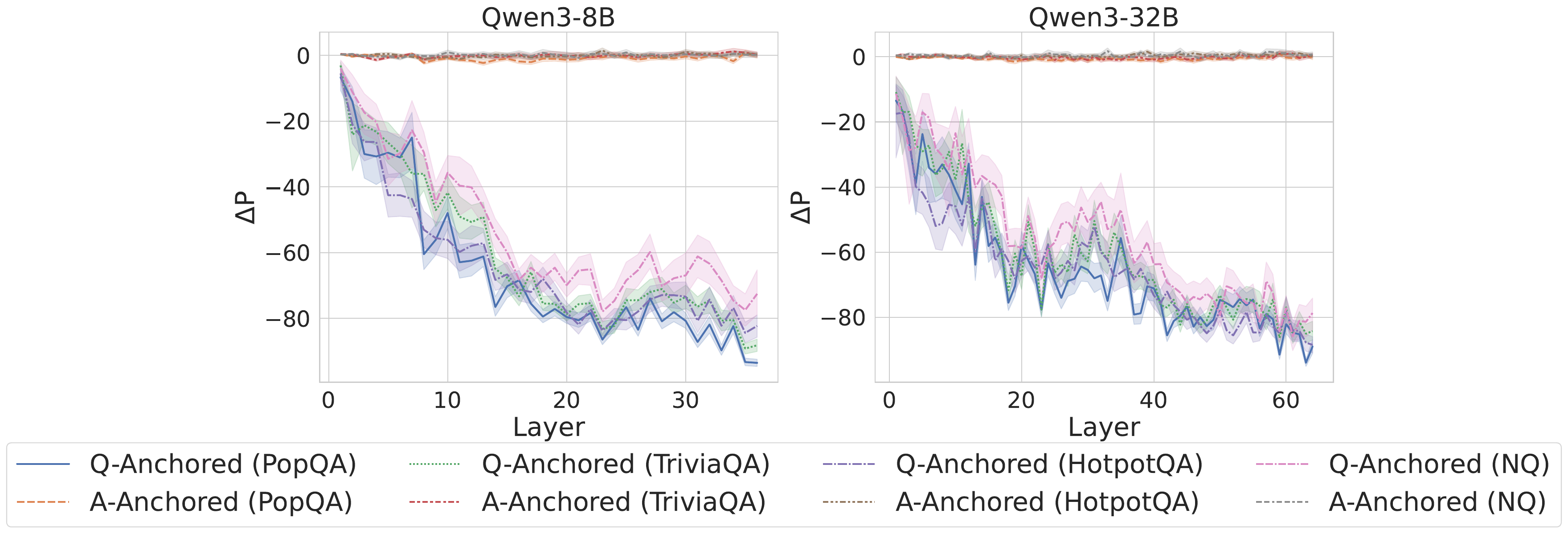}
\caption{$\Delta \mathrm{P}$ under attention knockout for reasoning models. Probing attention activations for the final token (top), the token immediately preceding the exact answer tokens (middle), and the last exact answer token (bottom).}
\label{fig:appendix_attention_knockout_reasoning_attnact}
\end{figure*}

\begin{figure*}[!htb]
\centering
\includegraphics[width=\textwidth]{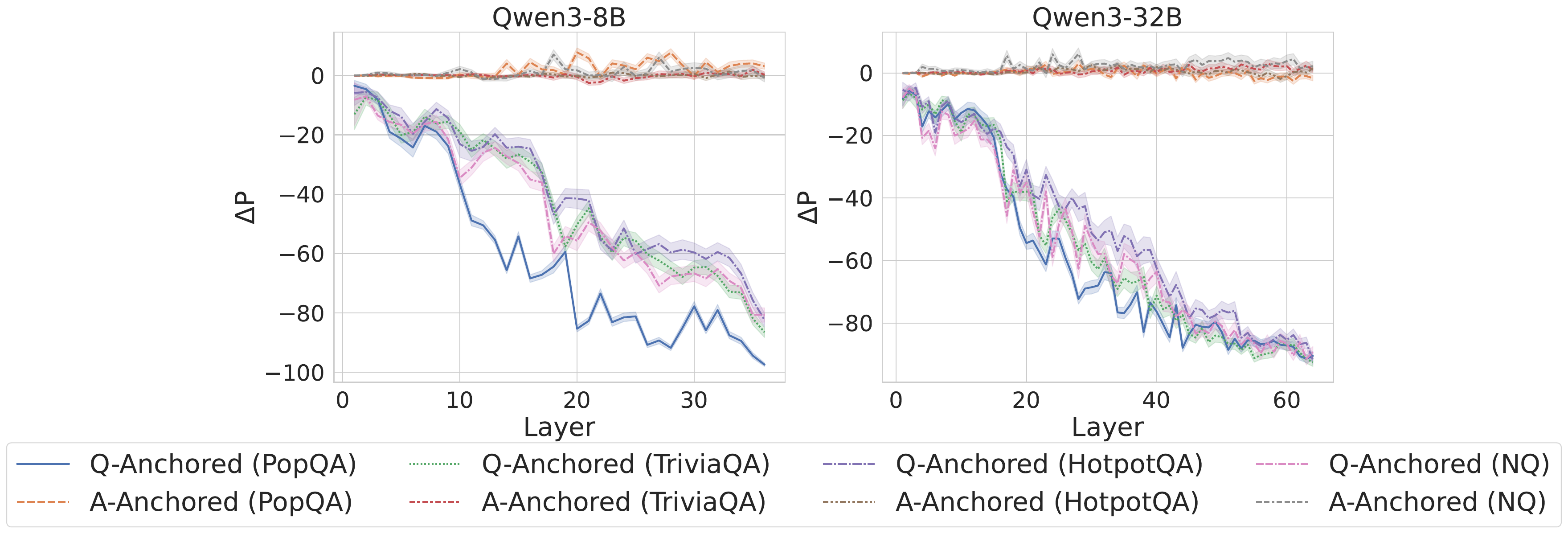}
\\[5ex]
\includegraphics[width=\textwidth]{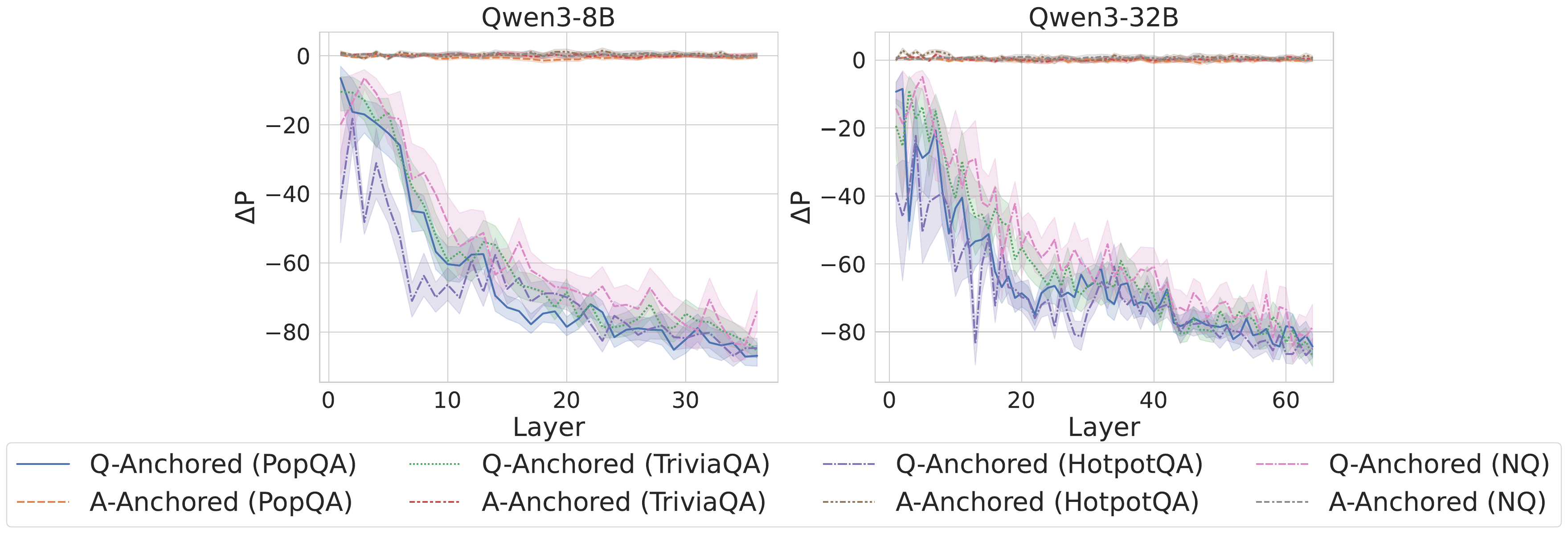}
\\[5ex]
\includegraphics[width=\textwidth]{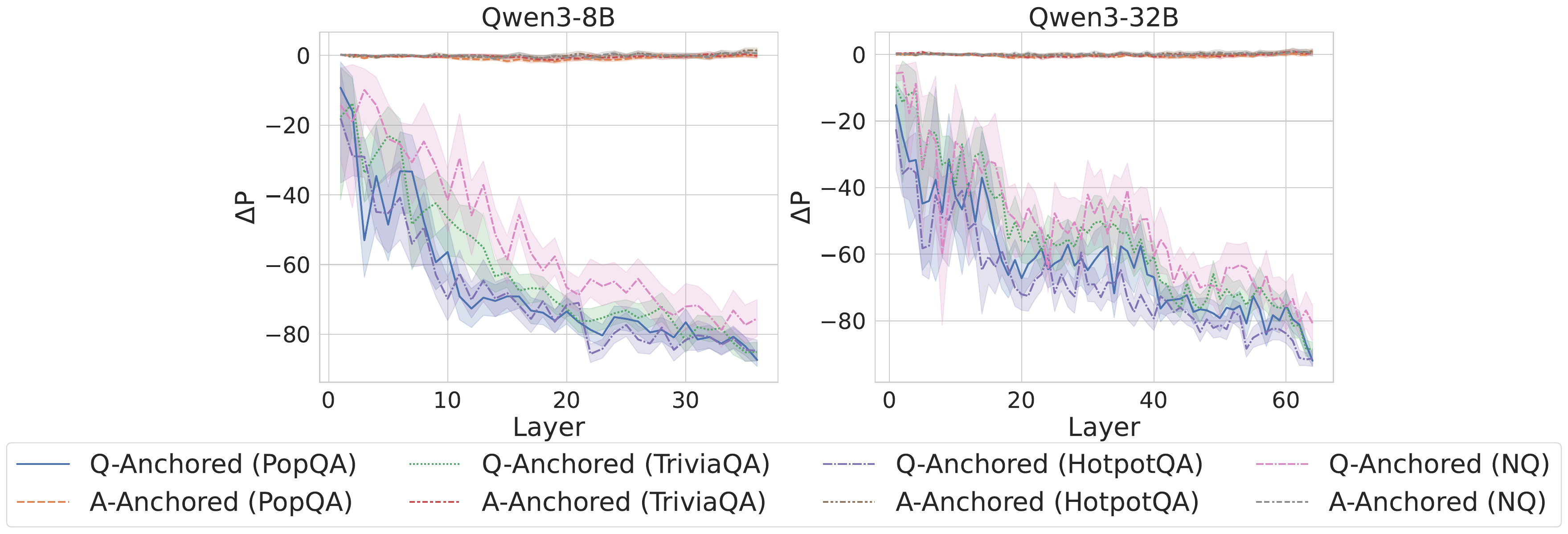}
\caption{$\Delta \mathrm{P}$ under attention knockout for reasoning models. Probing mlp activations for the final token (top), the token immediately preceding the exact answer tokens (middle), and the last exact answer token (bottom).}
\label{fig:appendix_attention_knockout_reasoning_mlpact}
\end{figure*}

\begin{figure*}[!htb]
\centering
\includegraphics[width=\textwidth]{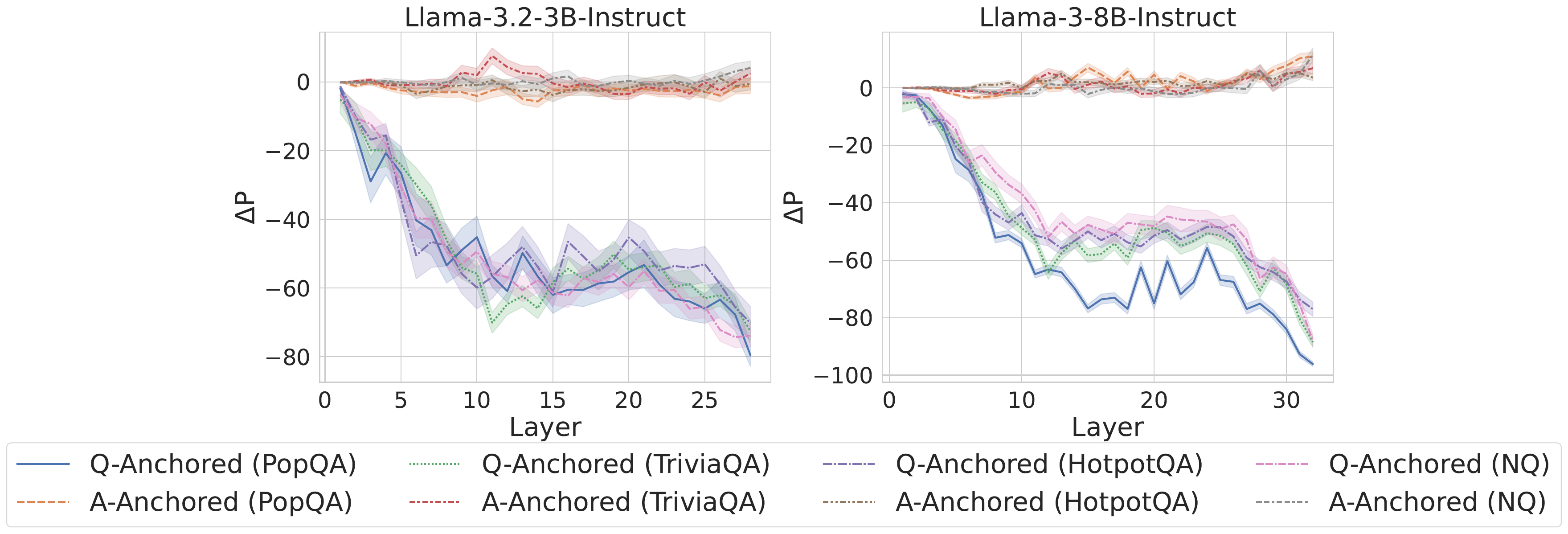}
\\[5ex]
\includegraphics[width=\textwidth]{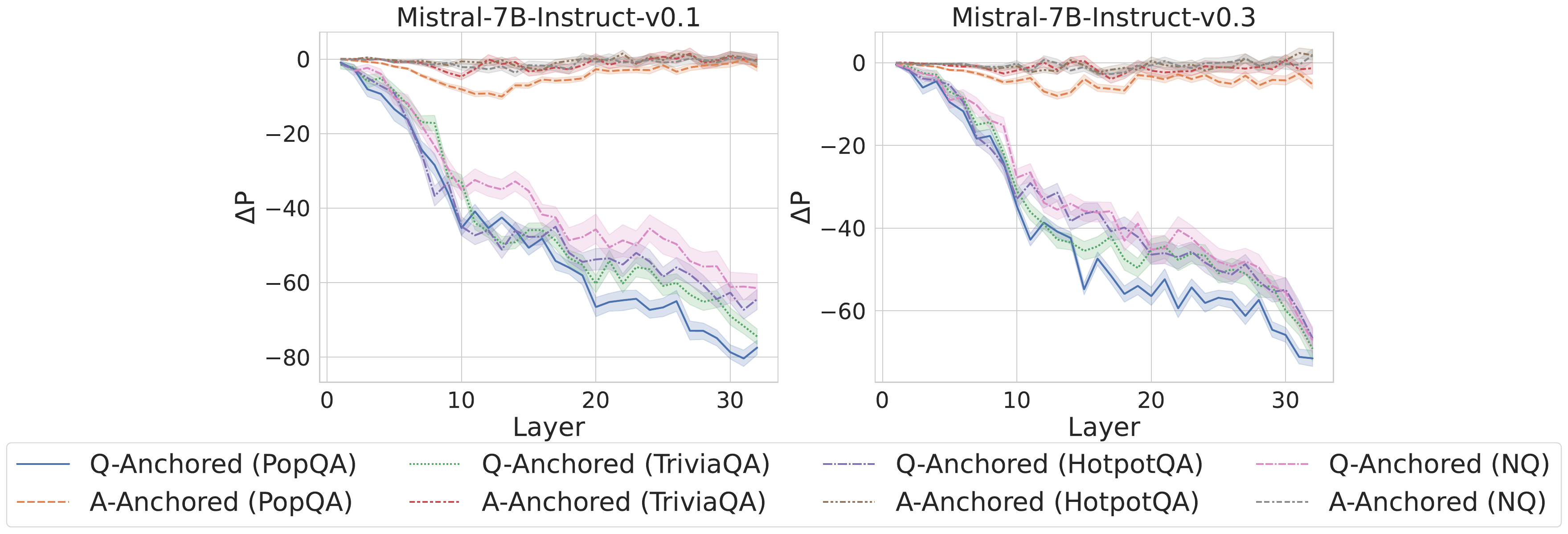}
\caption{$\Delta \mathrm{P}$ under attention knockout for instruct models.}
\label{fig:appendix_attention_knockout_instruct_mlpact_exactans}
\end{figure*}

\begin{figure*}[!htb]
\centering
\includegraphics[width=\textwidth]{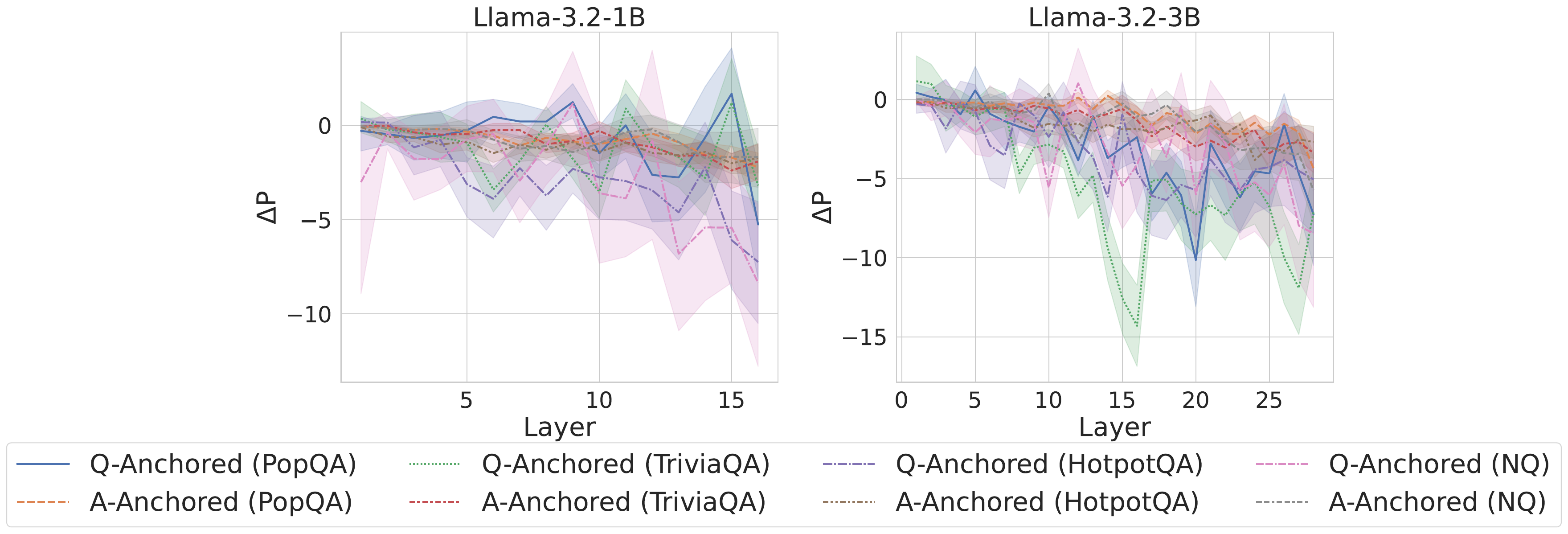}
\\[5ex]
\includegraphics[width=\textwidth]{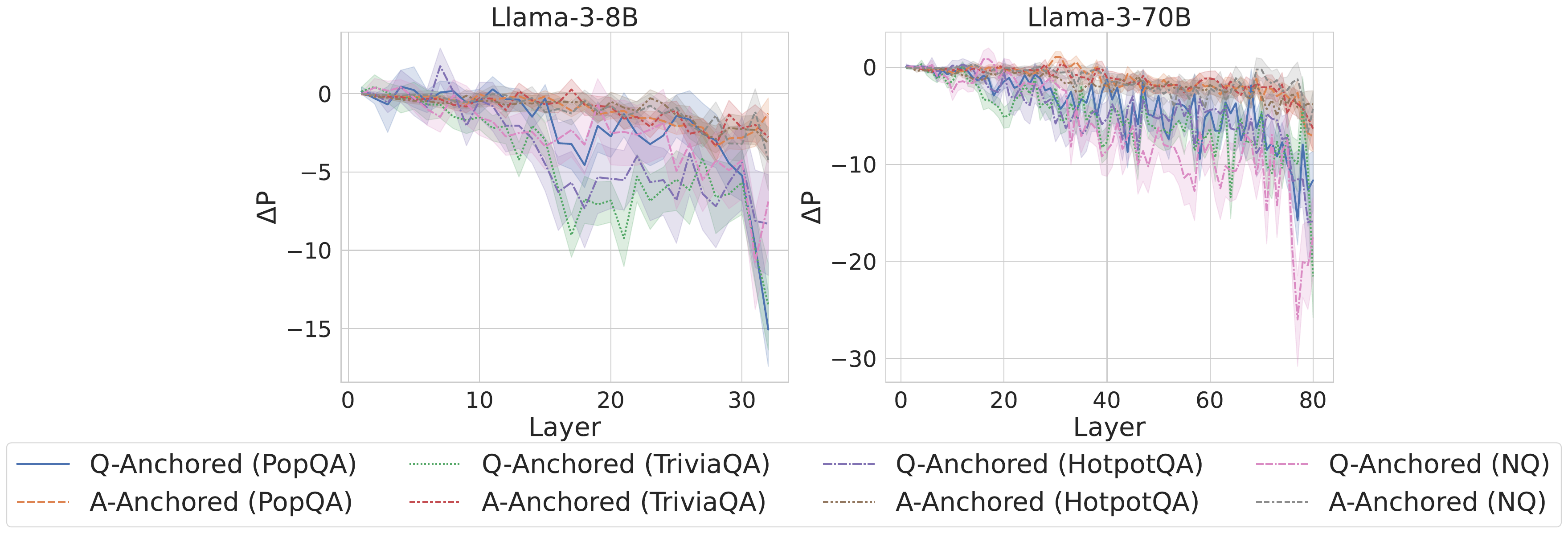}
\\[5ex]
\includegraphics[width=\textwidth]{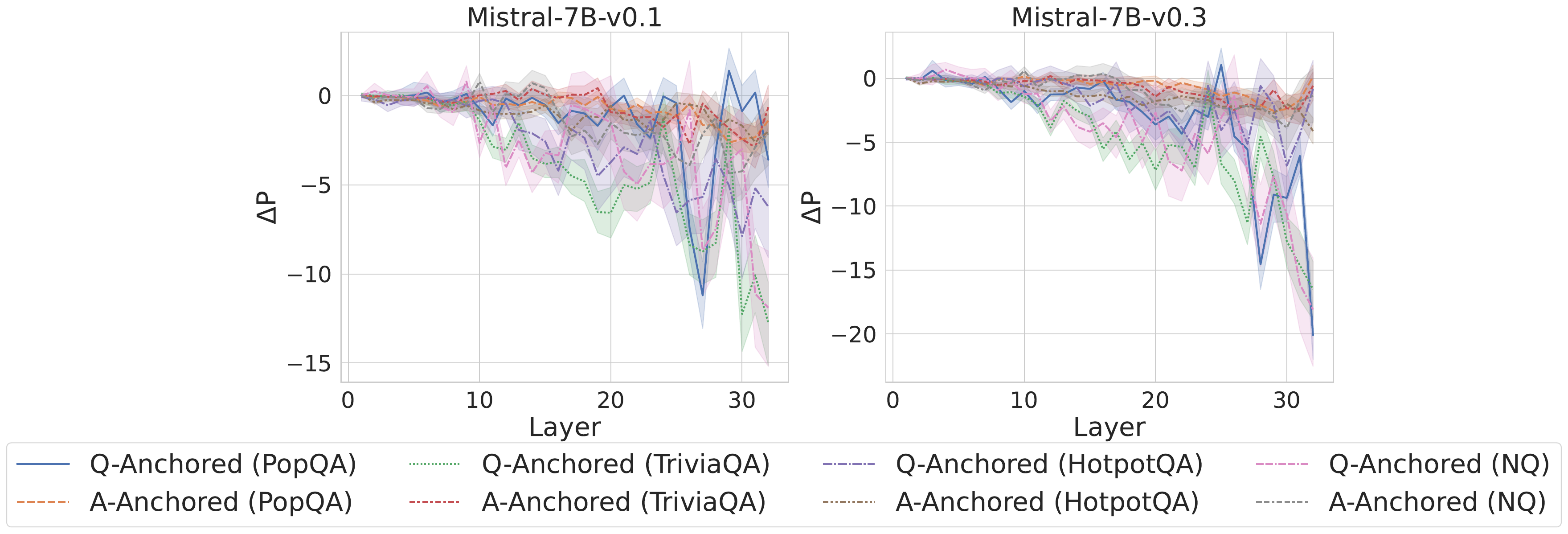}
\caption{$\Delta \mathrm{P}$ under attention knockout with randomly masked question tokens. Unlike selectively blocking the exact question tokens, both Q-Anchored and A-Anchored samples exhibit similar patterns, with substantially smaller probability changes when question tokens are masked at random. This suggests that exact question tokens play a critical role in conveying the semantic information of core frame elements.}
\label{fig:appendix_attention_knockout_base_randomblocking_mlpact_exactans}
\end{figure*}

\clearpage
\section{Token Patching}
\label{sec:appendix_token_patching}

\begin{minipage}{\textwidth}
    \centering

    \includegraphics[width=0.8\textwidth]{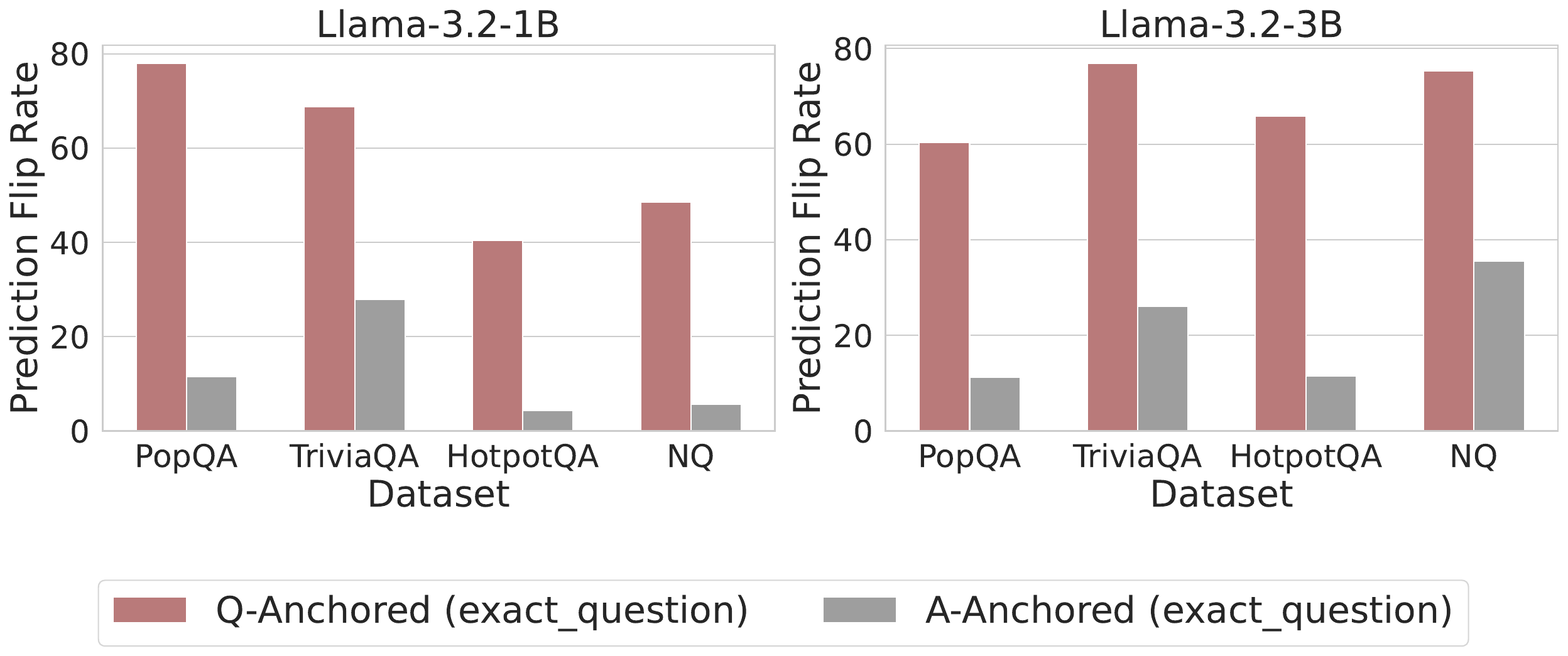}

    \vspace{4ex}

    \includegraphics[width=0.8\textwidth]{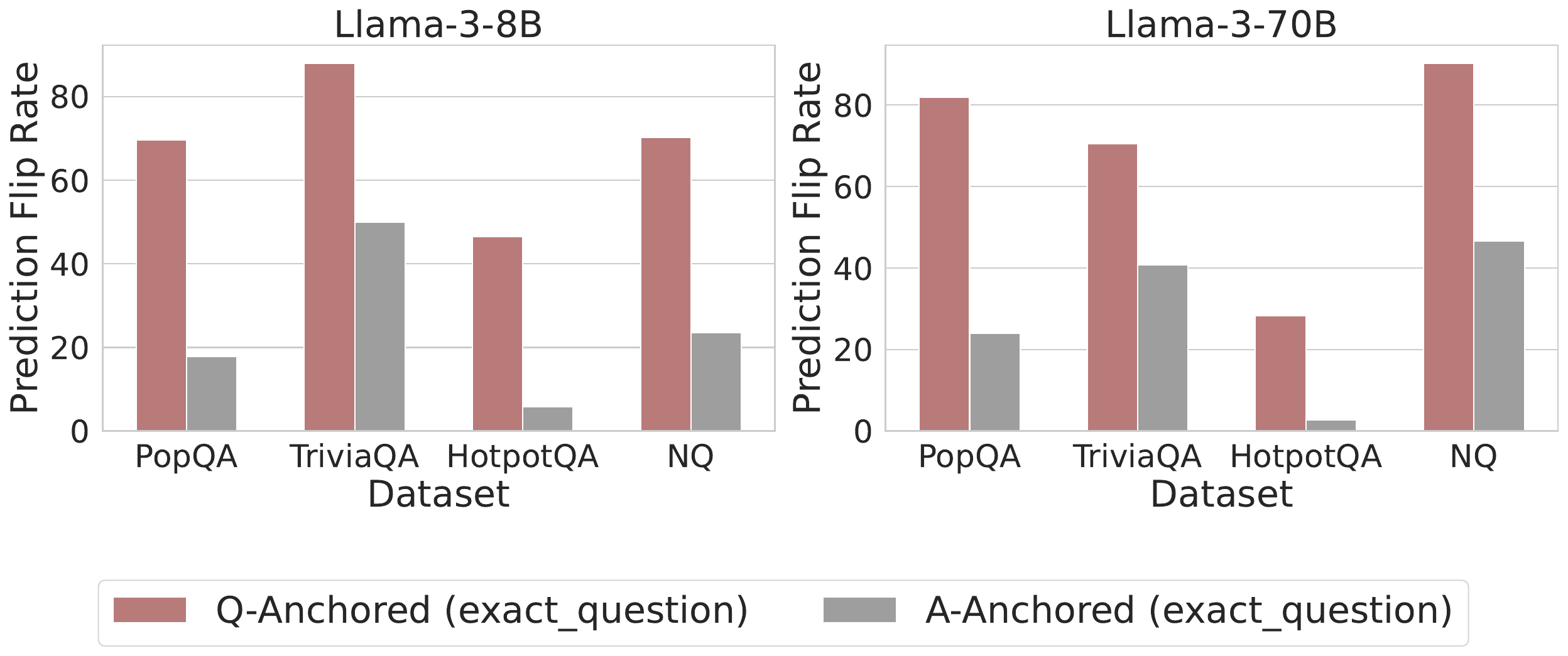}

    \vspace{4ex}

    \includegraphics[width=0.8\textwidth]{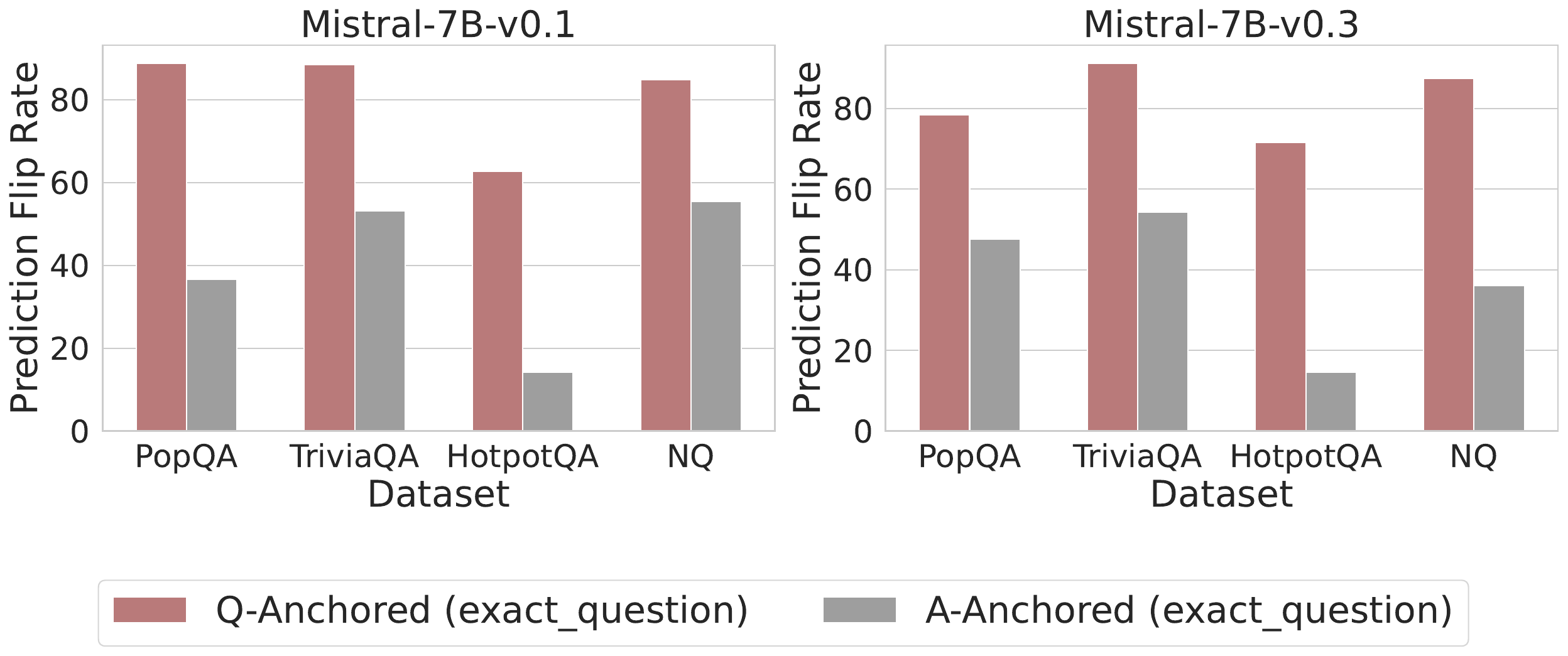}

    \vspace{2ex}

    \captionof{figure}{Prediction flip rate under token patching, probing attention activations of the final token.}
    \label{fig:appendix_token_patching_base_attnact_-1_0}
\end{minipage}

\begin{figure*}[!htb]
\centering
\includegraphics[width=0.8\textwidth]{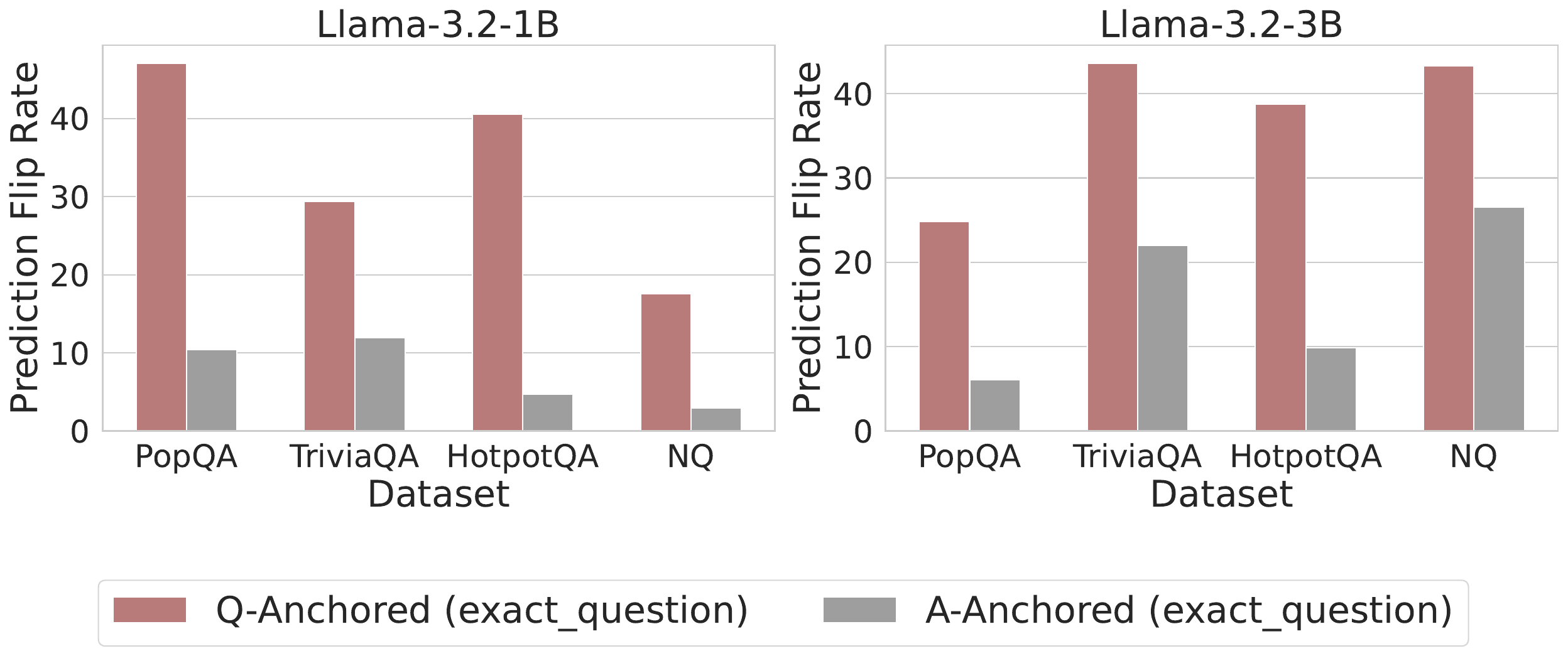}
\\[5ex]
\includegraphics[width=0.8\textwidth]{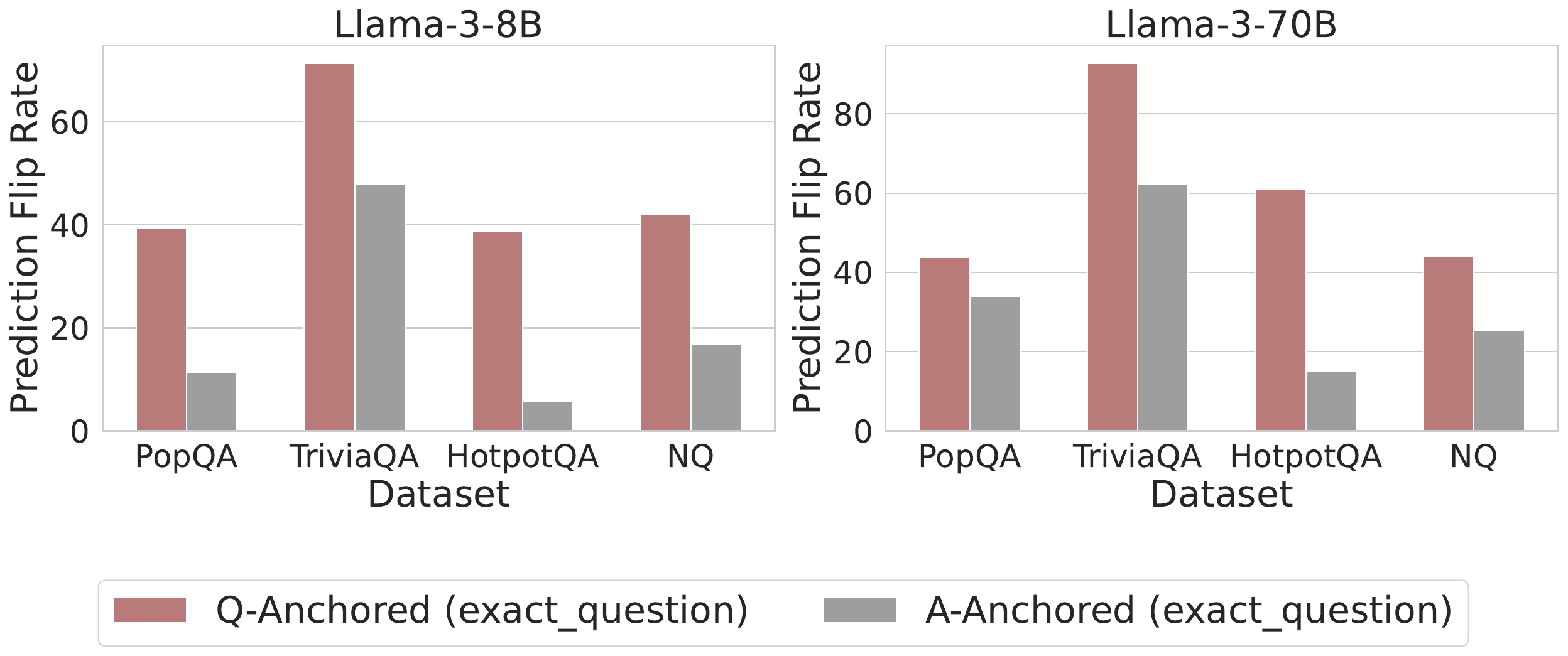}
\\[5ex]
\includegraphics[width=0.8\textwidth]{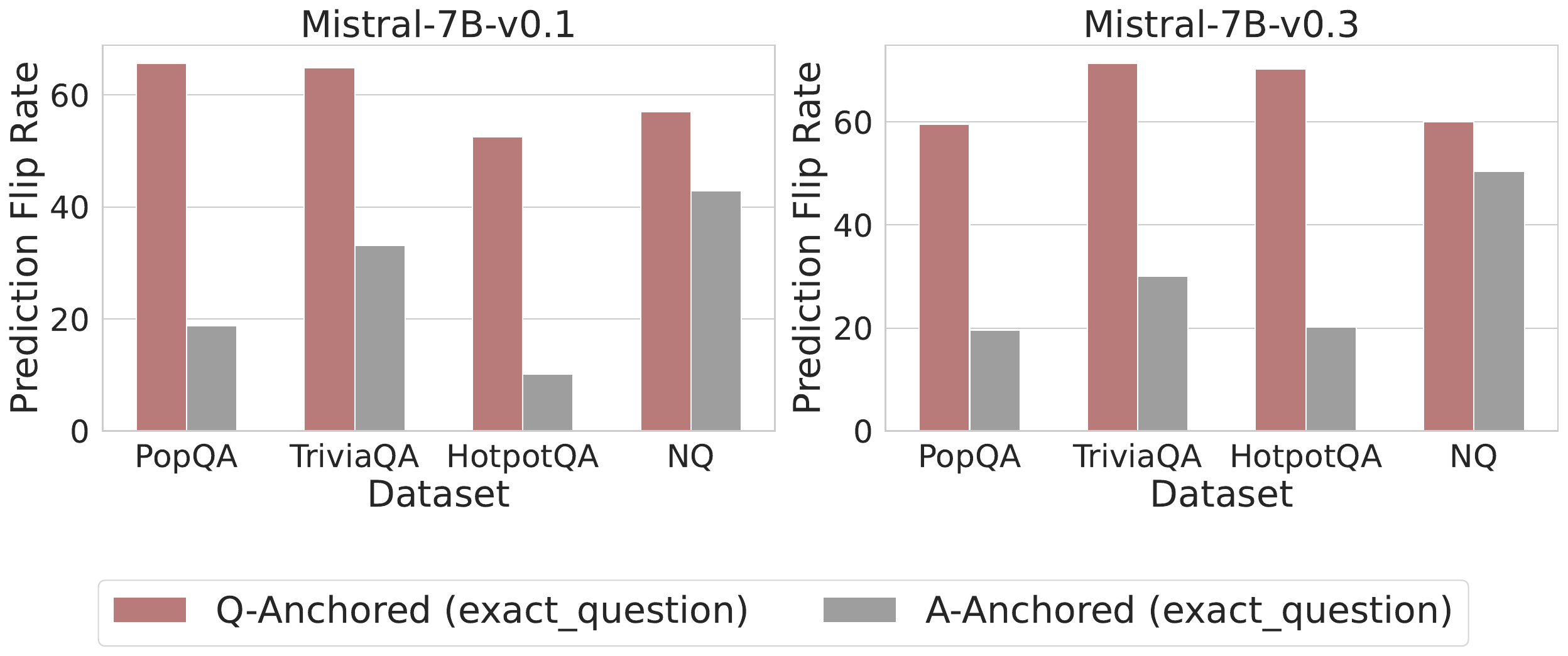}
\caption{Prediction flip rate under token patching, probing attention activations of the token immediately preceding the exact answer tokens.}
\label{fig:appendix_token_patching_base_attnact_beforefirst_0}
\end{figure*}

\begin{figure*}[!htb]
\centering
\includegraphics[width=0.8\textwidth]{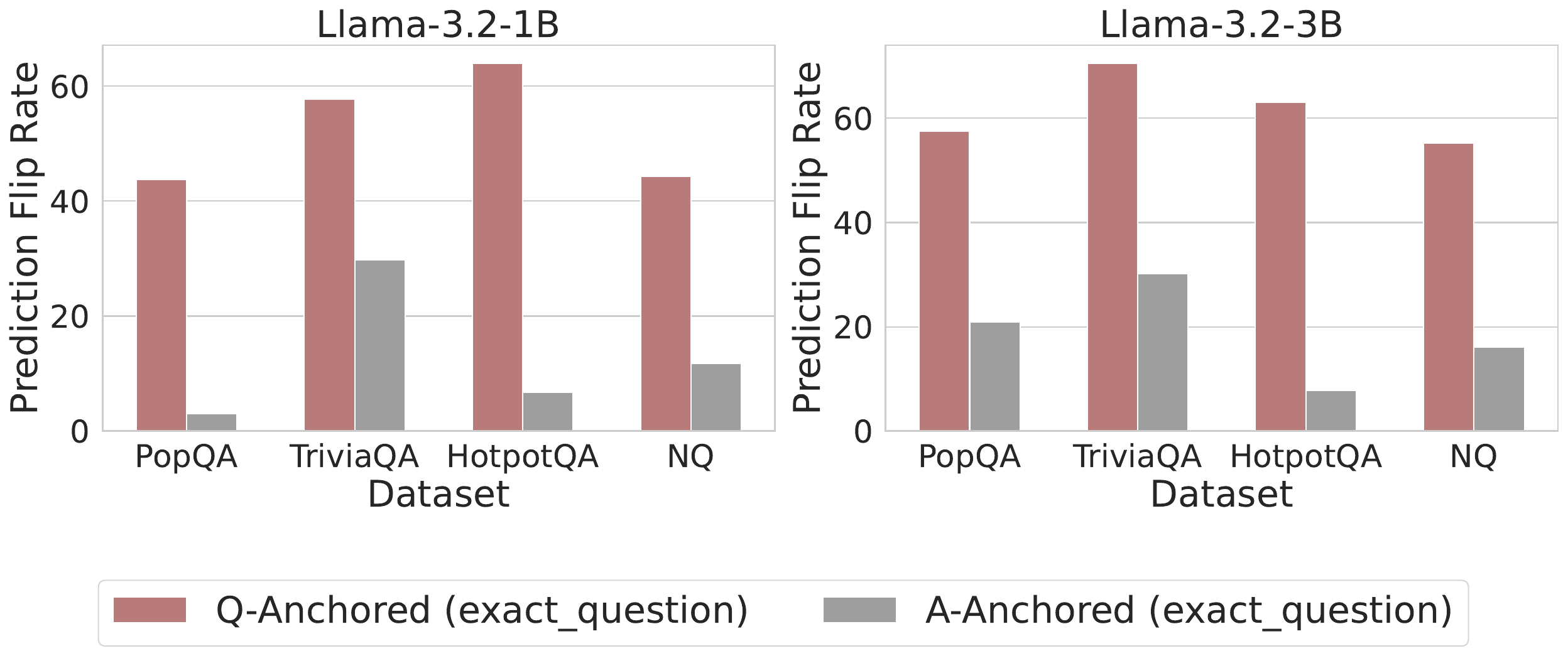}
\\[5ex]
\includegraphics[width=0.8\textwidth]{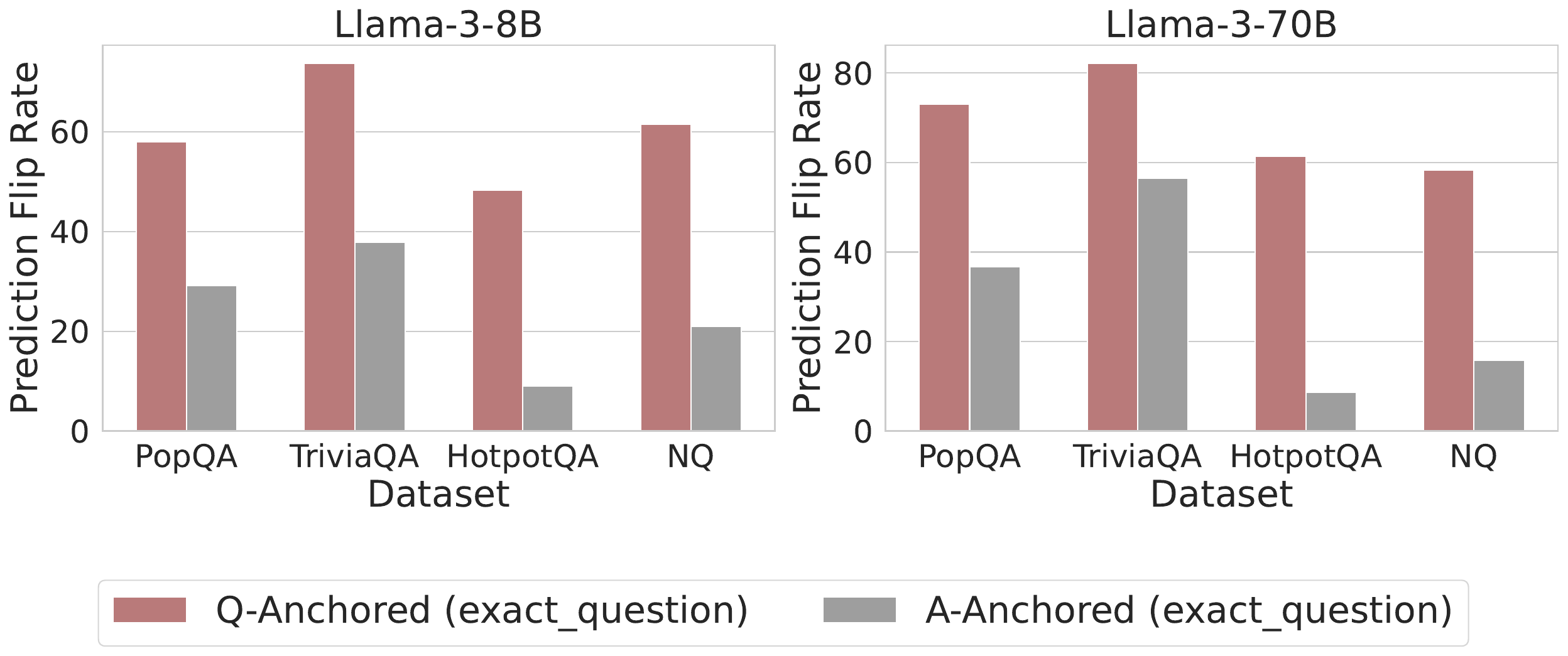}
\\[5ex]
\includegraphics[width=0.8\textwidth]{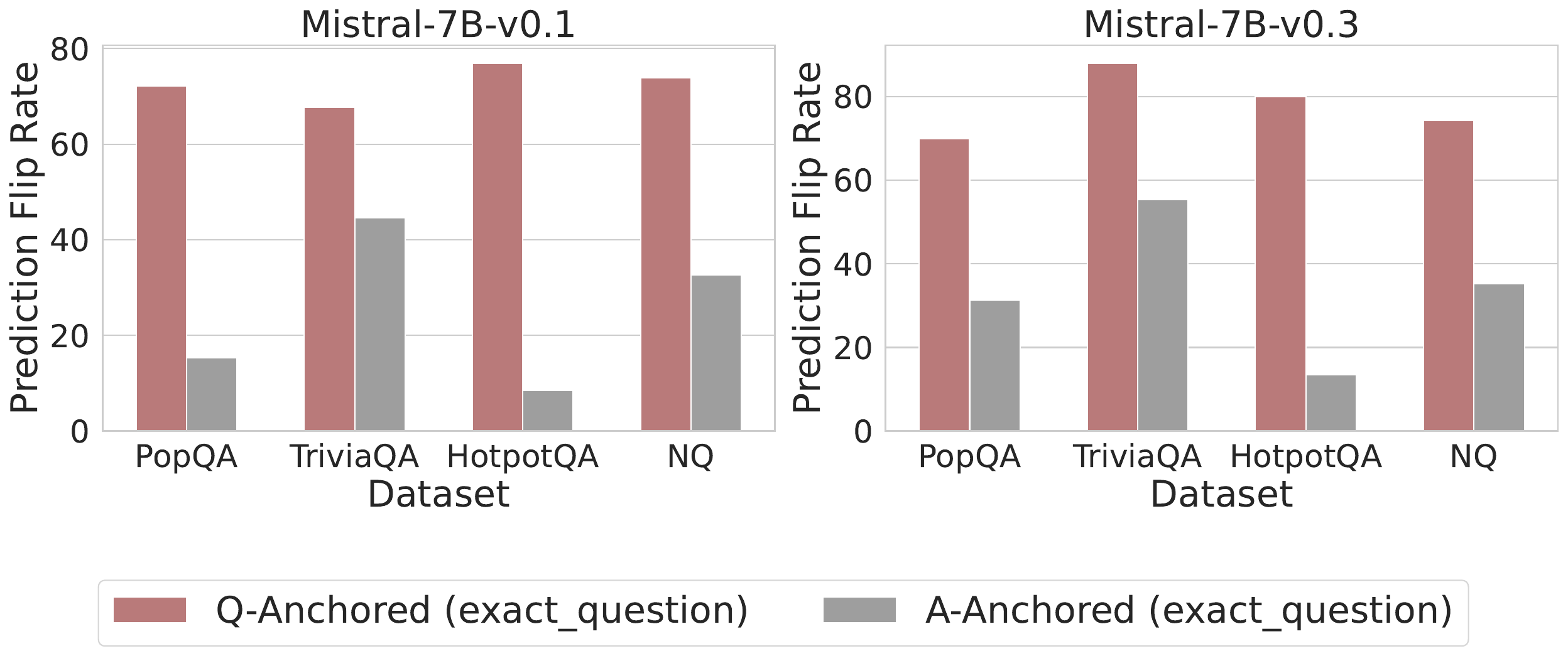}
\caption{Prediction flip rate under token patching, probing attention activations of the last exact answer token.}
\label{fig:appendix_token_patching_base_attnact_exactans_0}
\end{figure*}

\begin{figure*}[!htb]
\centering
\includegraphics[width=0.8\textwidth]{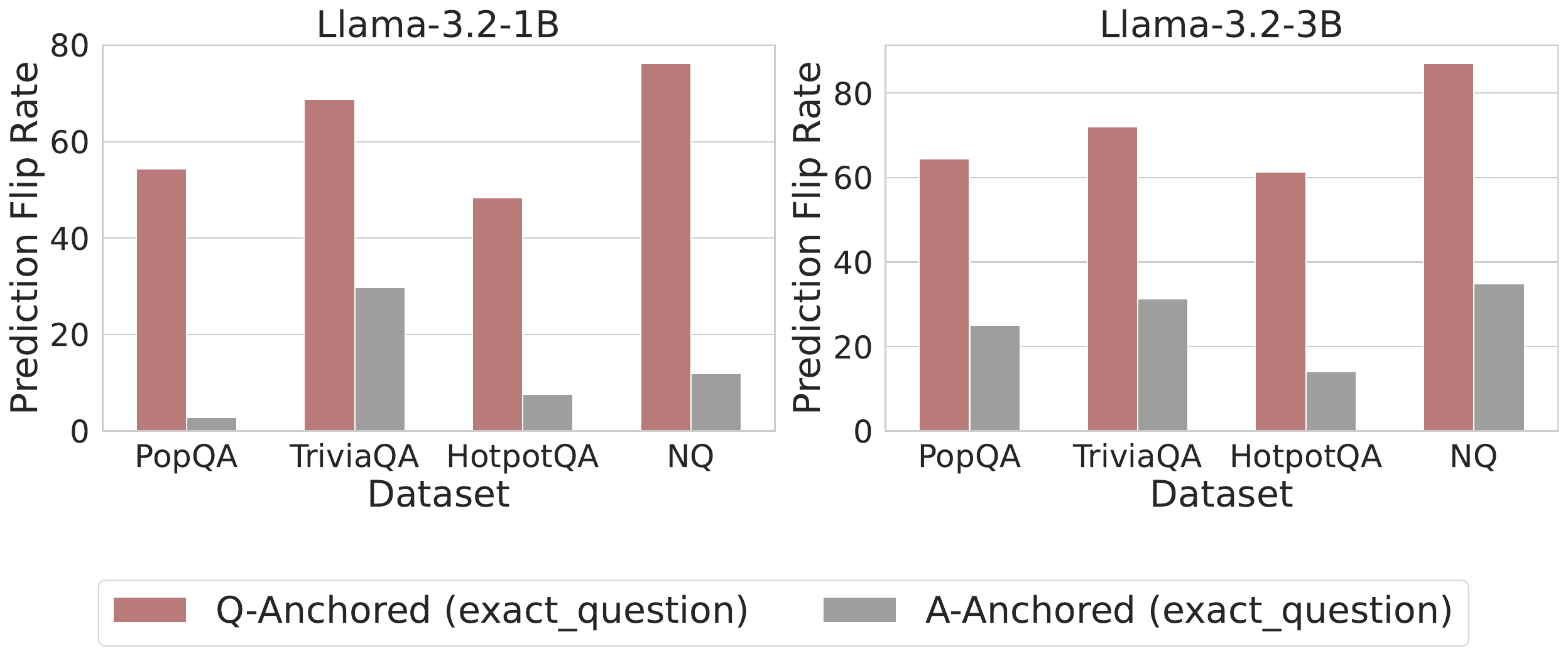}
\\[5ex]
\includegraphics[width=0.8\textwidth]{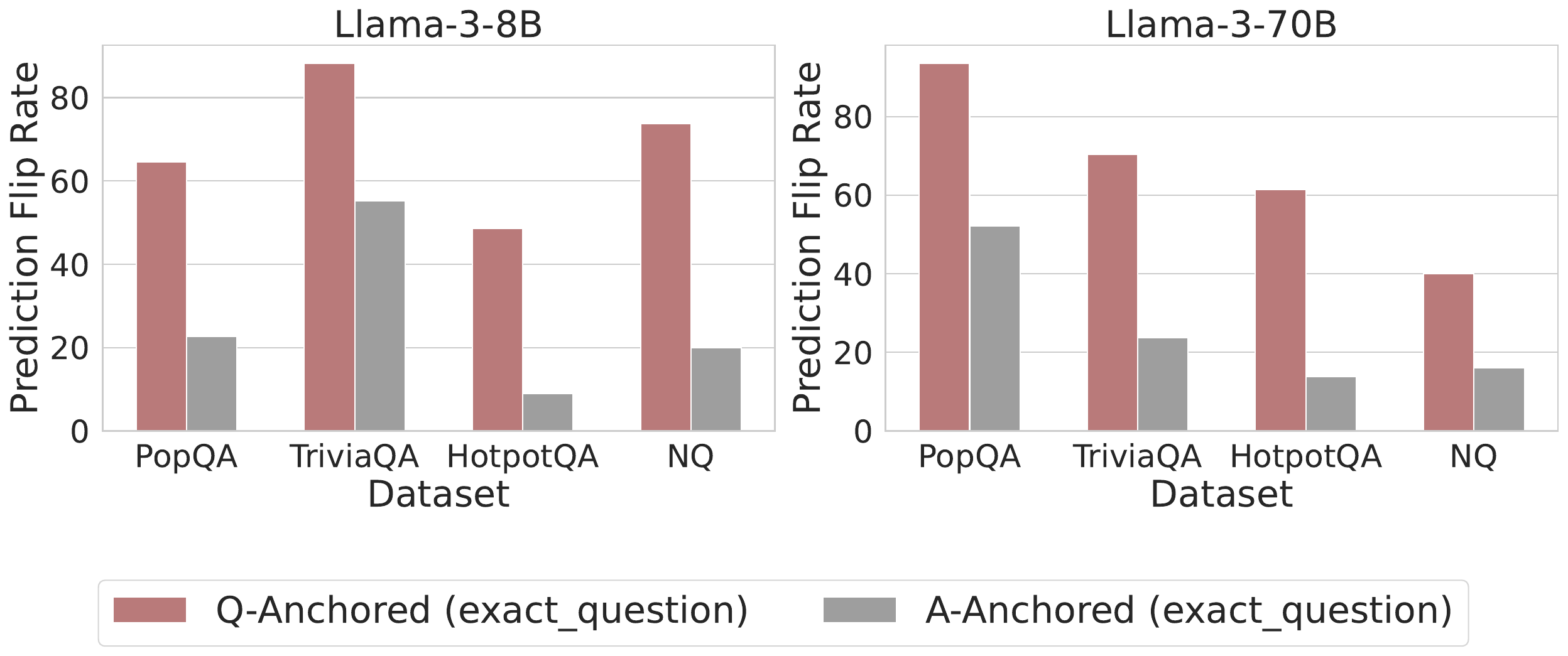}
\\[5ex]
\includegraphics[width=0.8\textwidth]{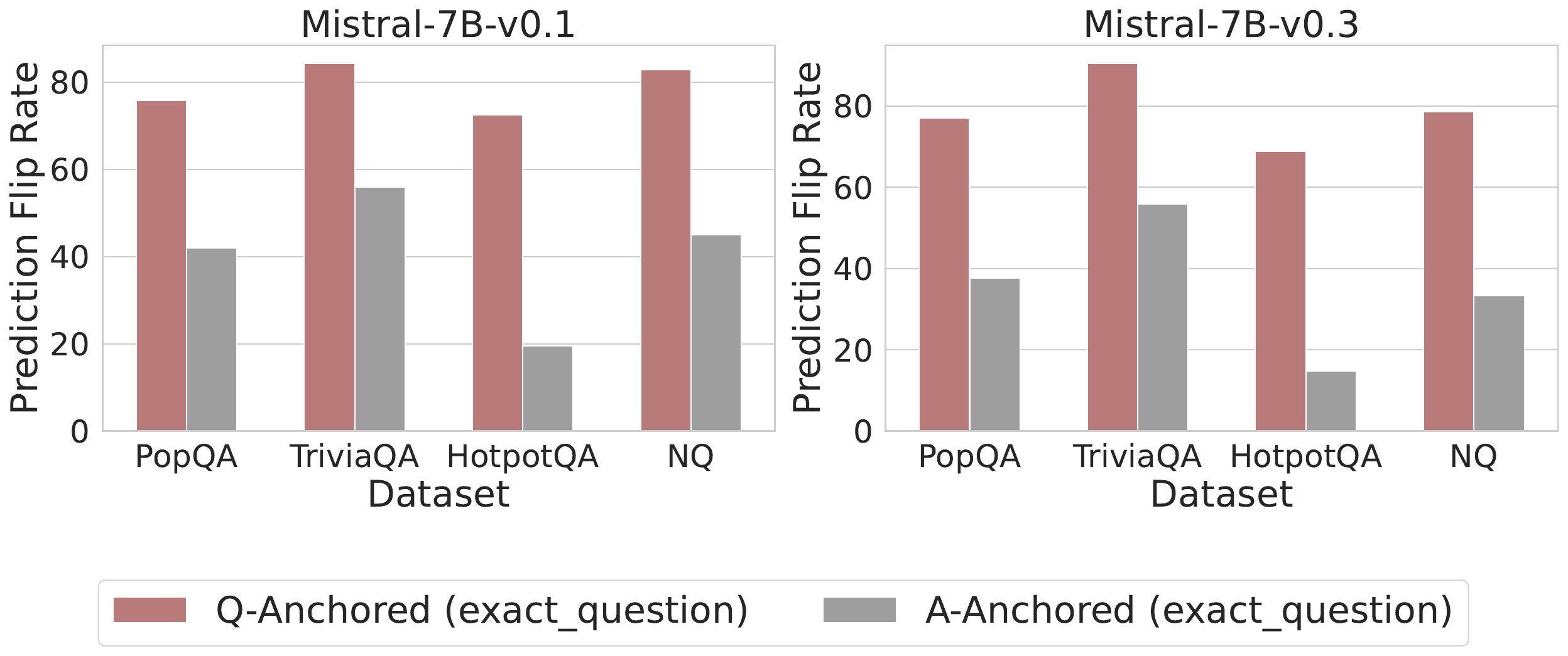}
\caption{Prediction flip rate under token patching, probing mlp activations of the final token.}
\label{fig:appendix_token_patching_base_mlpact_-1_0}
\end{figure*}

\begin{figure*}[!htb]
\centering
\includegraphics[width=0.8\textwidth]{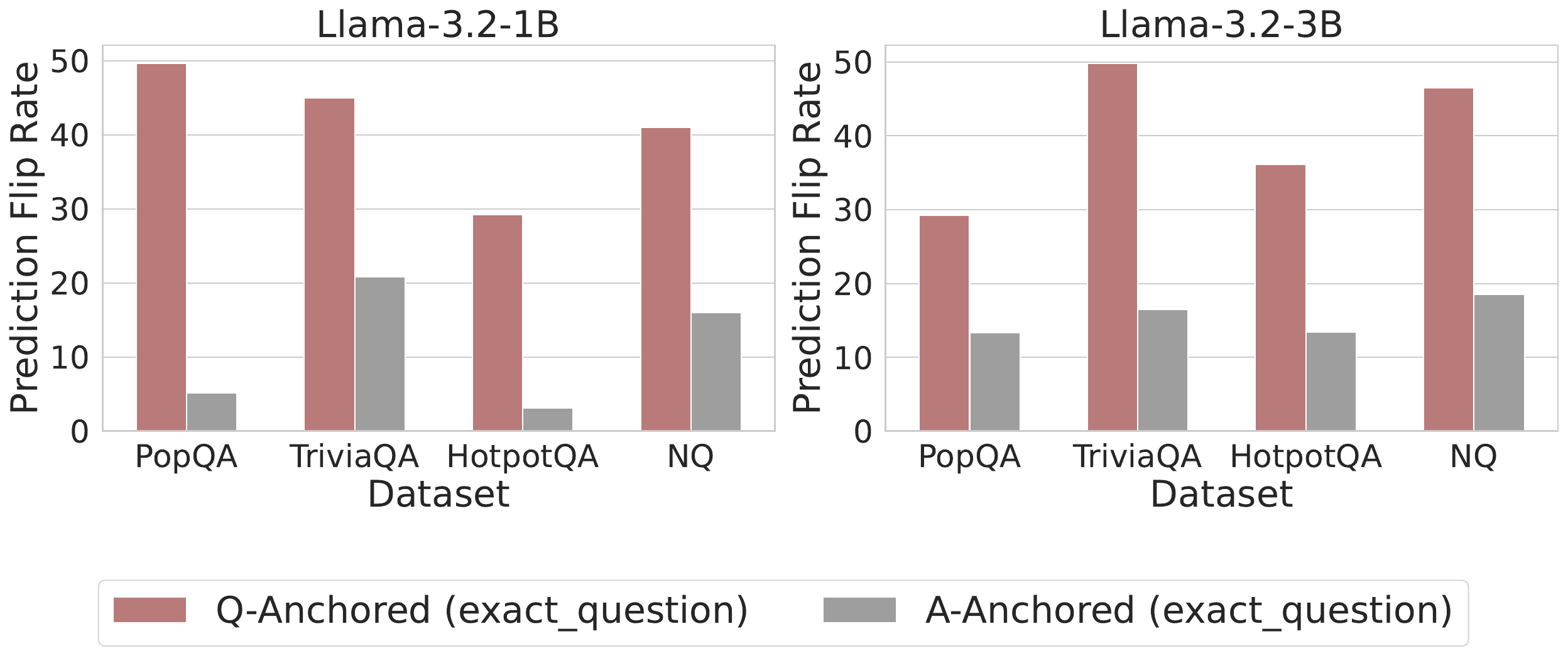}
\\[5ex]
\includegraphics[width=0.8\textwidth]{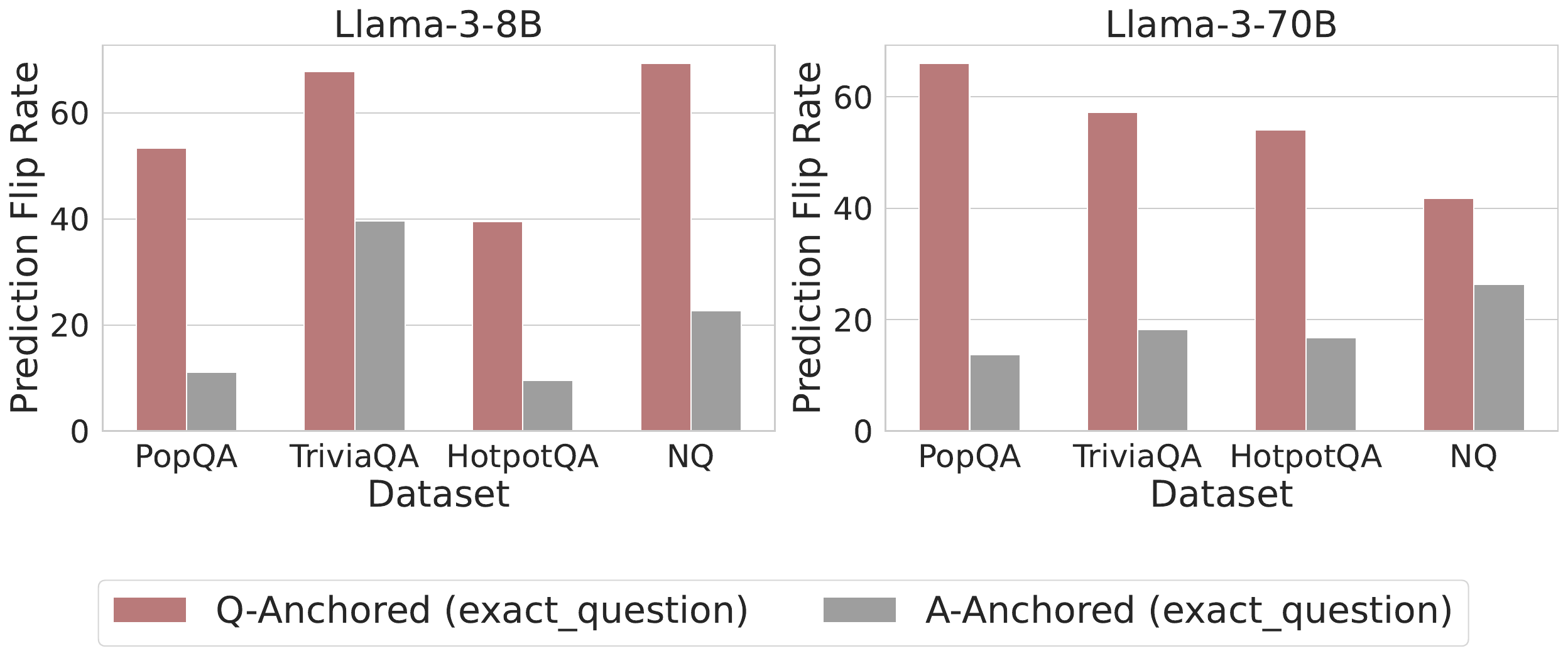}
\\[5ex]
\includegraphics[width=0.8\textwidth]{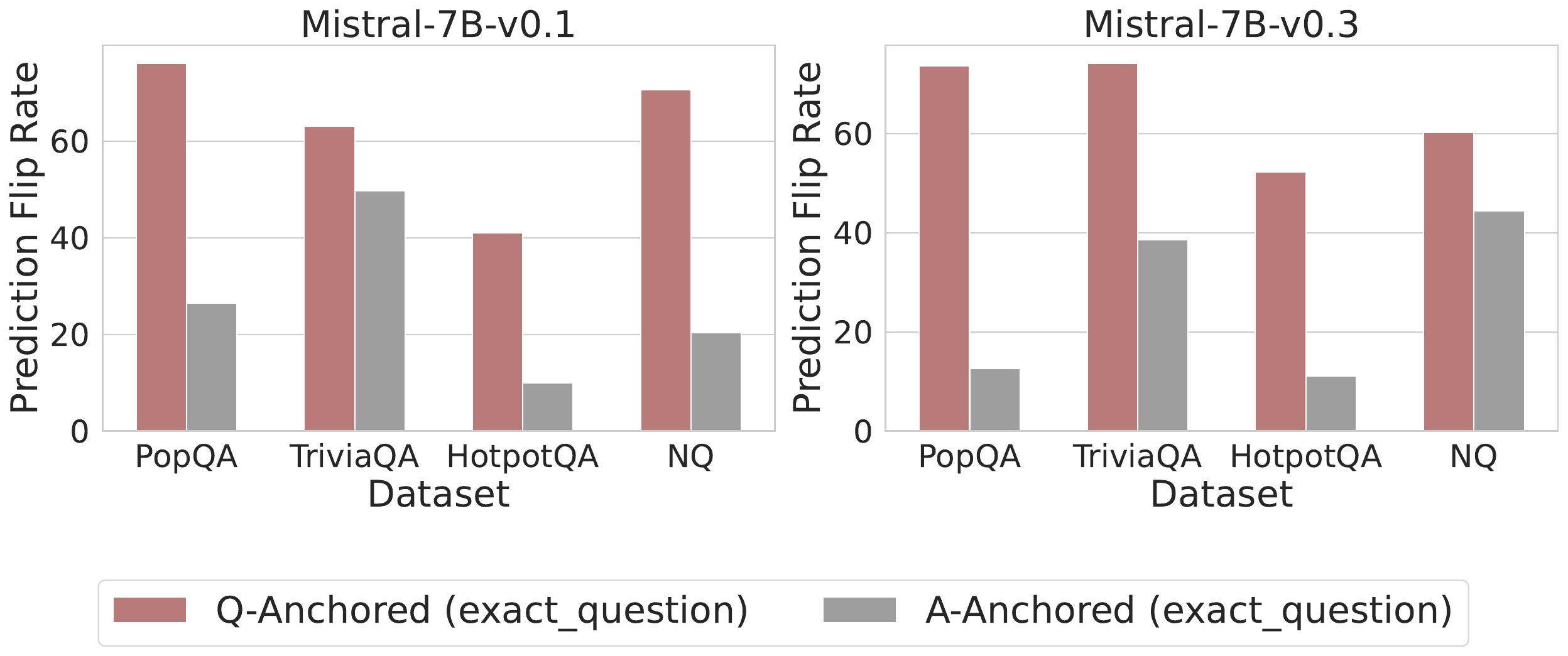}
\caption{Prediction flip rate under token patching, probing mlp activations of the token immediately preceding the exact answer tokens.}
\label{fig:appendix_token_patching_base_mlpact_beforefirst_0}
\end{figure*}

\begin{figure*}[!htb]
\centering
\includegraphics[width=0.8\textwidth]{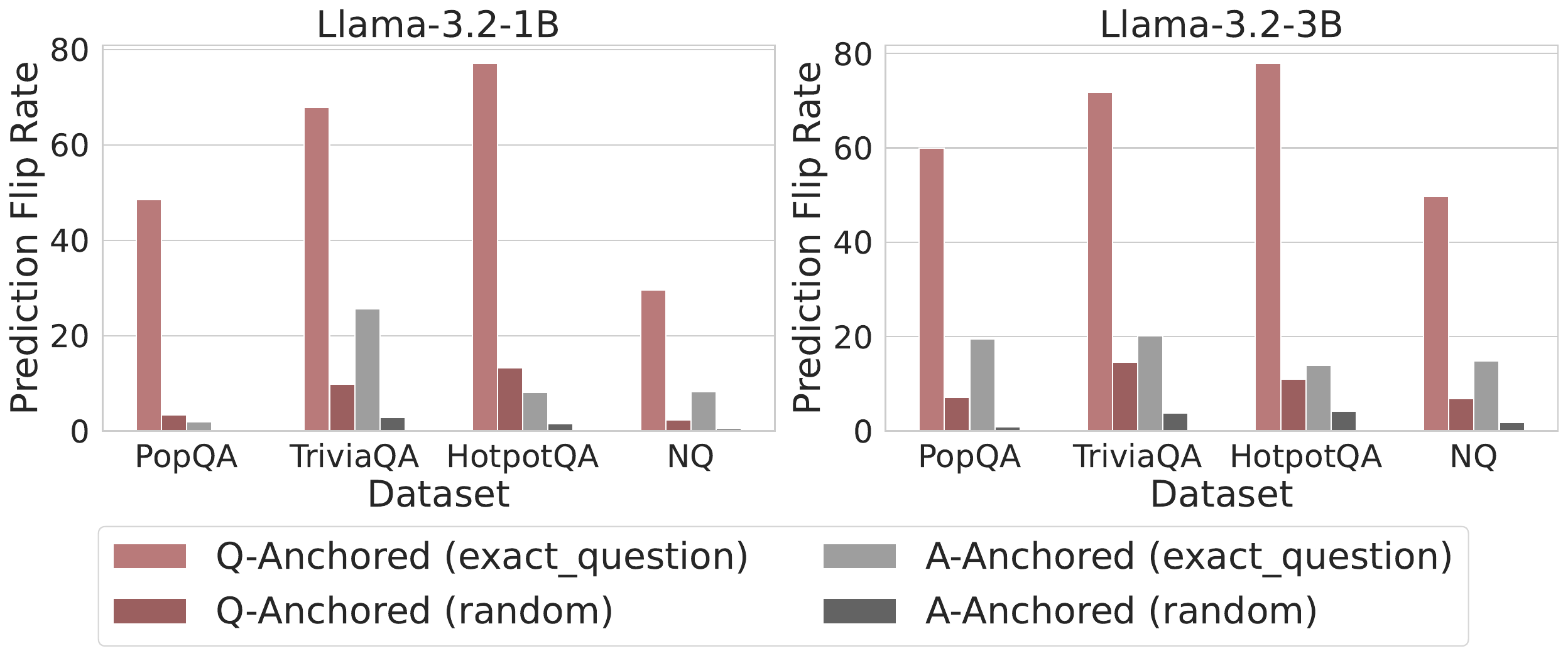}
\\[5ex]
\includegraphics[width=0.8\textwidth]{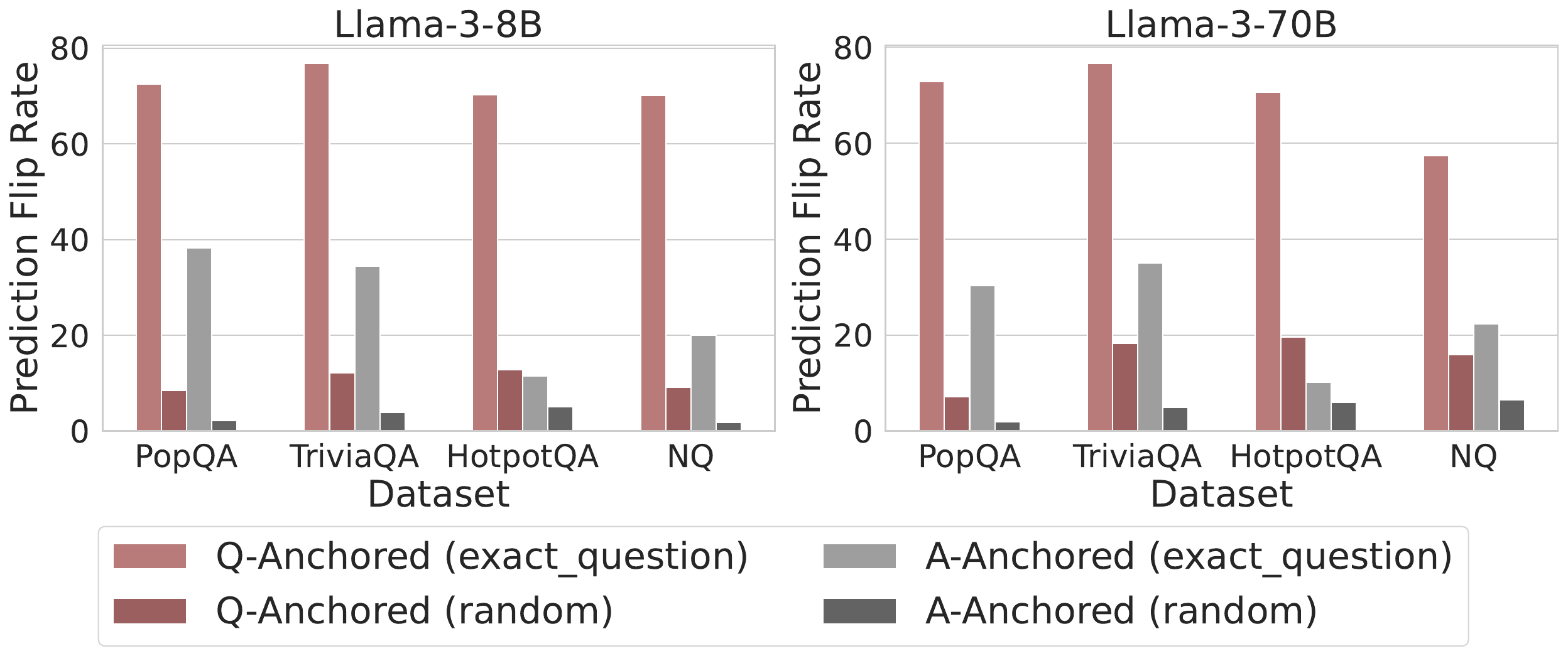}
\\[5ex]
\includegraphics[width=0.8\textwidth]{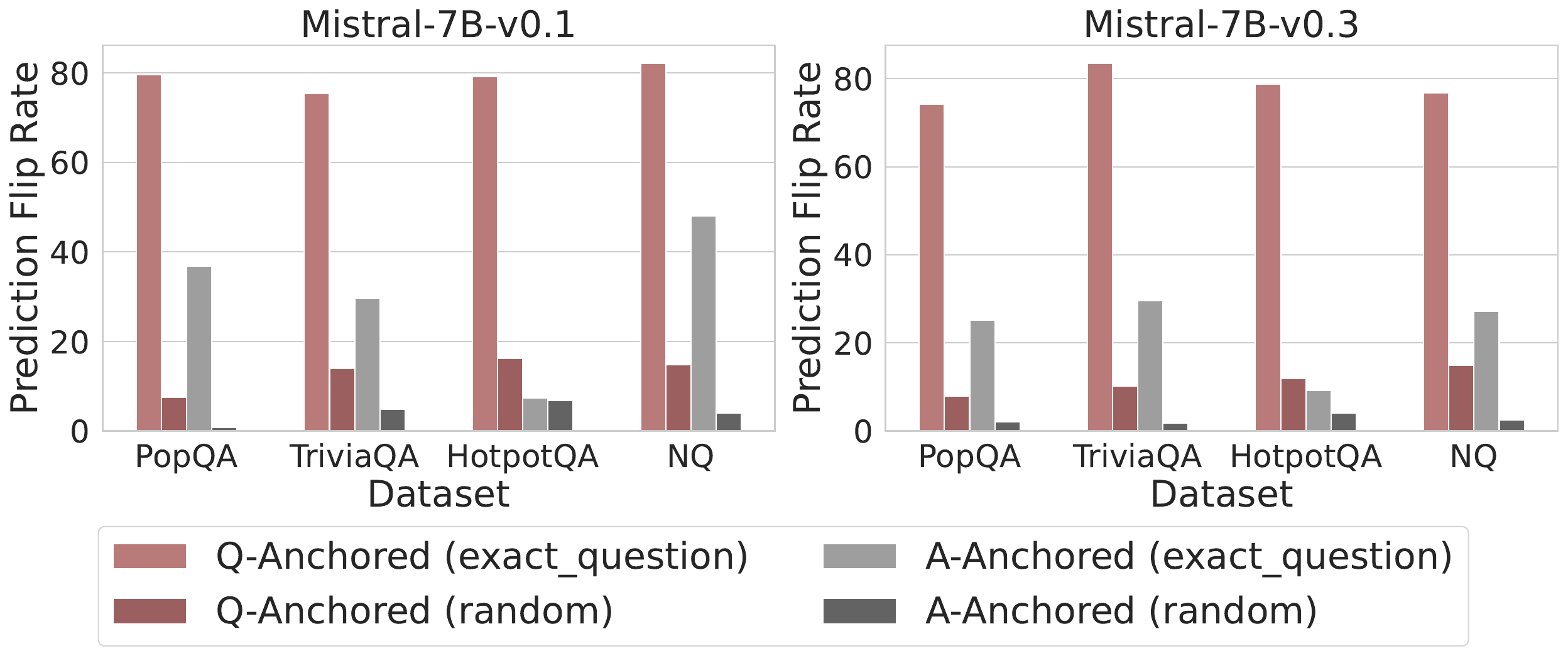}
\caption{Prediction flip rate under token patching, probing mlp activations of the last exact answer token.}
\label{fig:appendix_token_patching_base_mlpact_exactans_0}
\end{figure*}

\clearpage
\section{Answer-Only Input}
\label{sec:appendix_answer_only_input}

\begin{minipage}{\textwidth}
    \centering

    \includegraphics[width=0.8\textwidth]{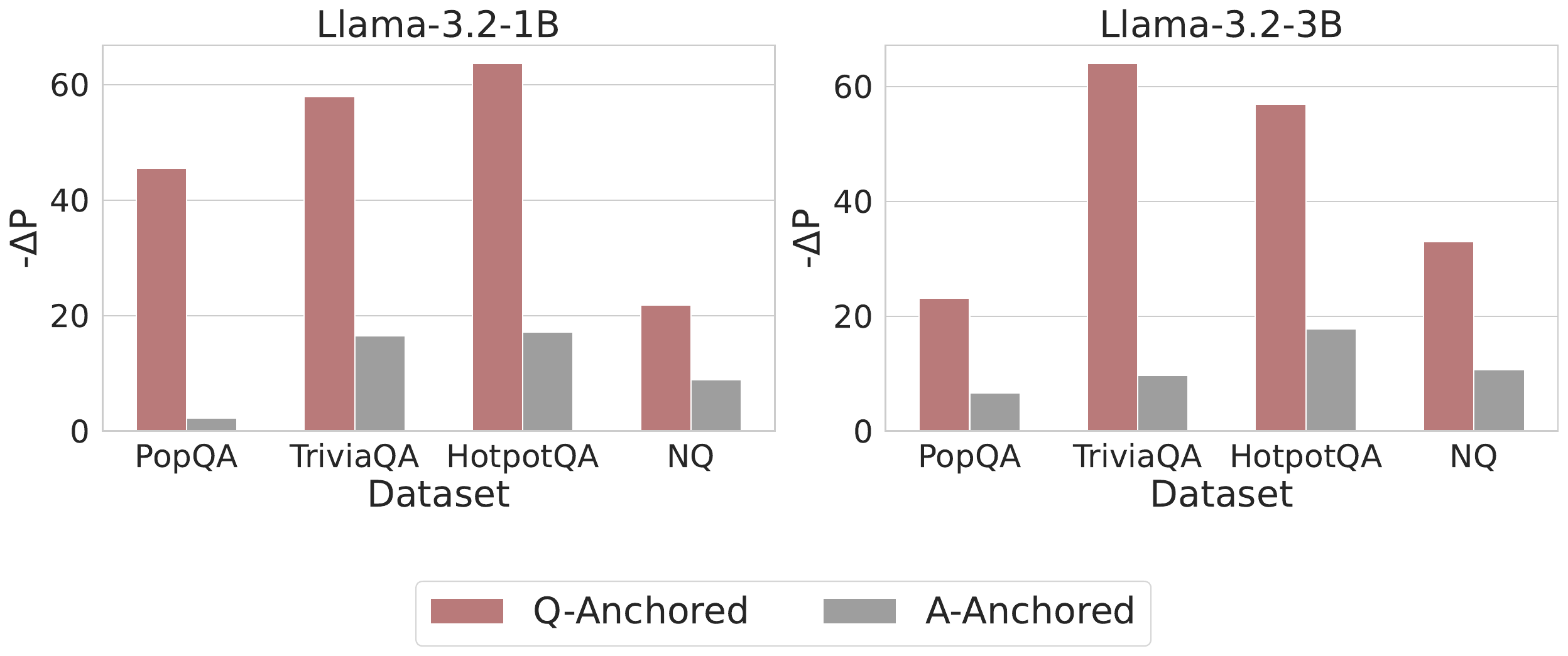}

    \vspace{4ex}

    \includegraphics[width=0.8\textwidth]{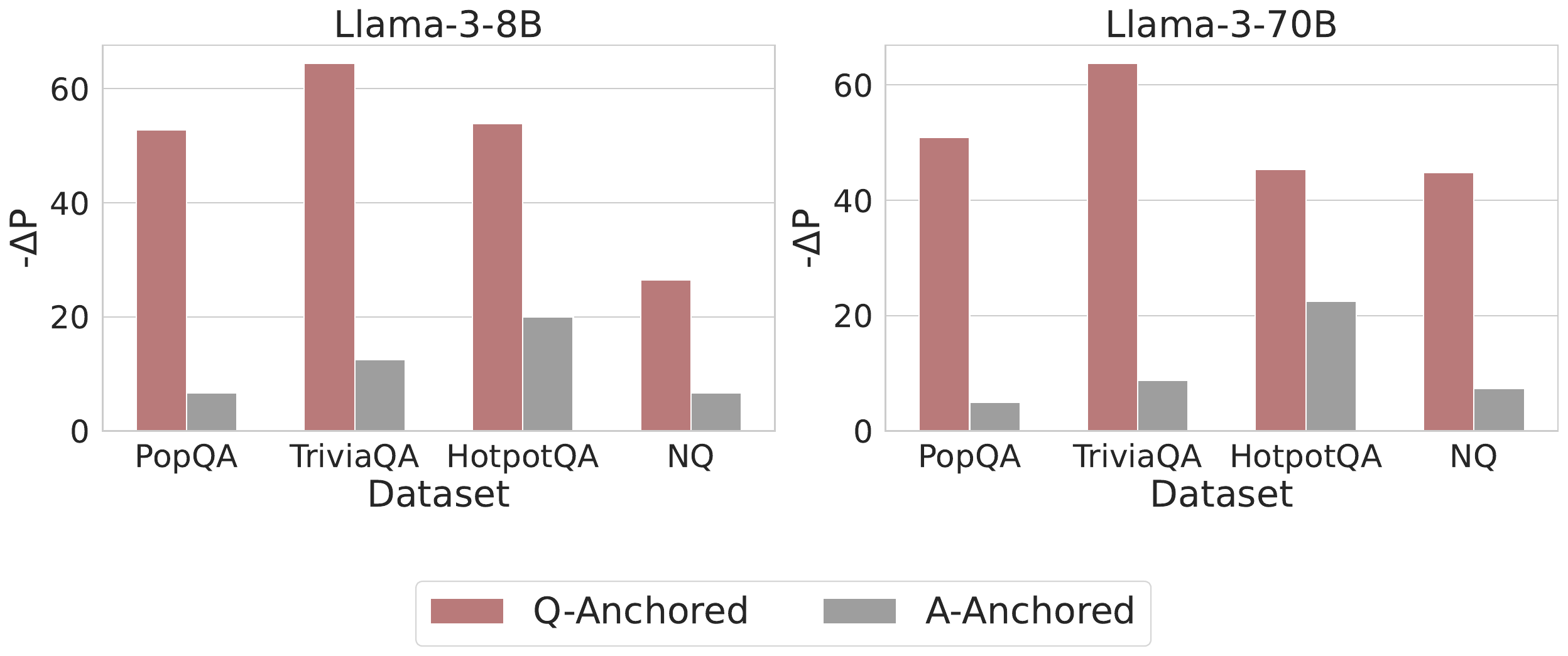}

    \vspace{4ex}

    \includegraphics[width=0.8\textwidth]{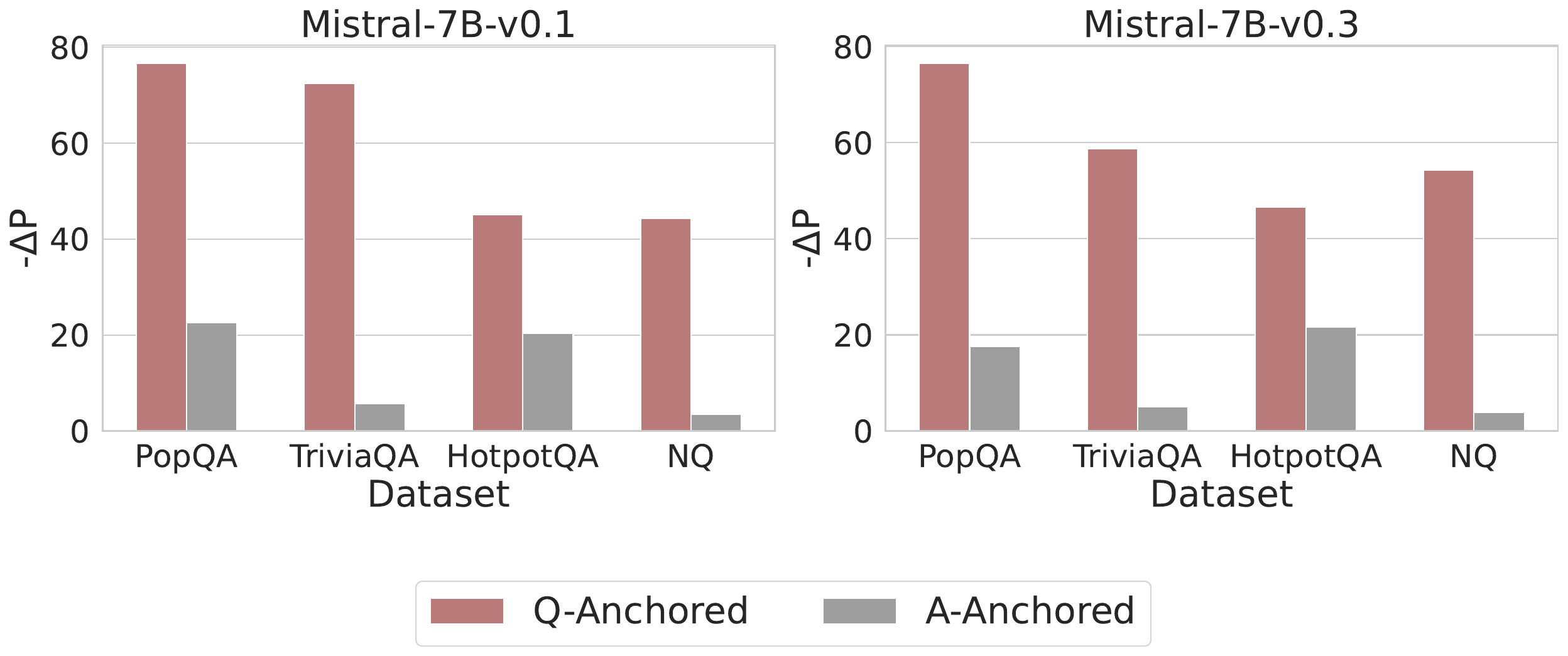}

    \vspace{2ex}

    \captionof{figure}{$-\Delta \mathrm{P}$ with only the LLM-generated answer. Q-Anchored instances exhibit substantial shifts, whereas A-Anchored instances remain stable, confirming that A-Anchored truthfulness encoding relies on information in the LLM-generated answer itself.}
    \label{fig:appendix_answer_only_input_base_mlpact_exactans}
\end{minipage}

\clearpage
\section{Answer Accuracy}
\label{sec:appendix_answer_accuracy}

\begin{minipage}{\textwidth}
    \centering

    \includegraphics[width=\textwidth]{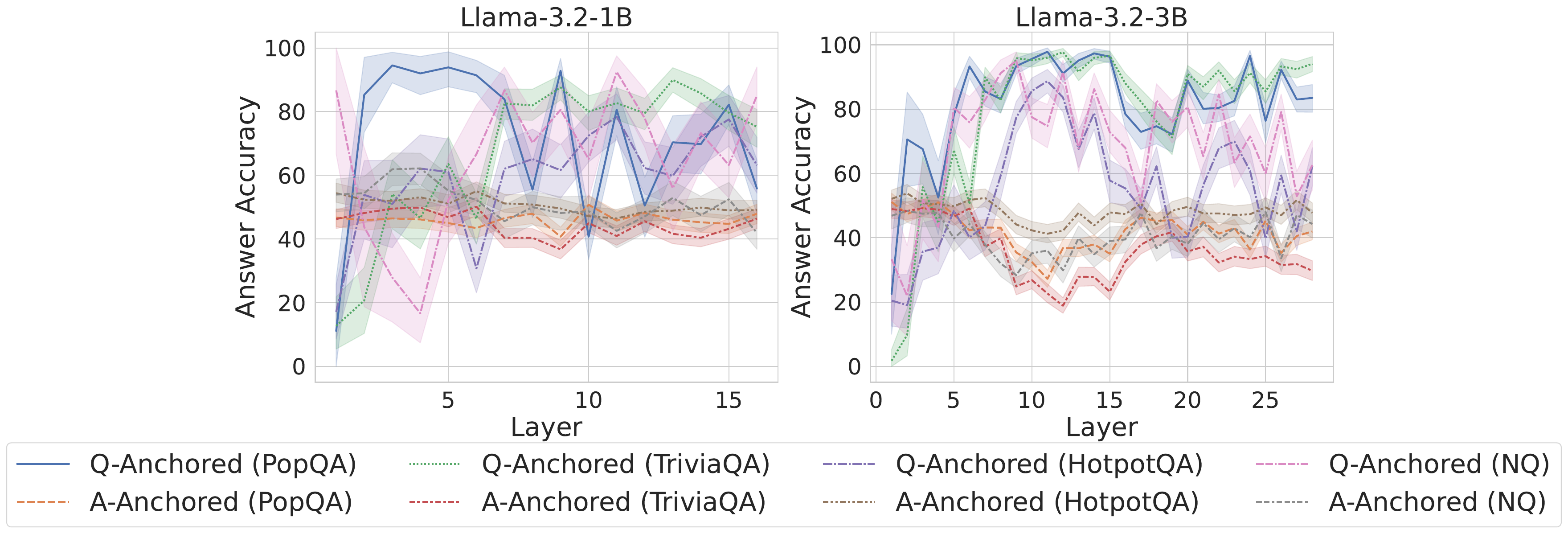}

    \vspace{4ex}

    \includegraphics[width=\textwidth]{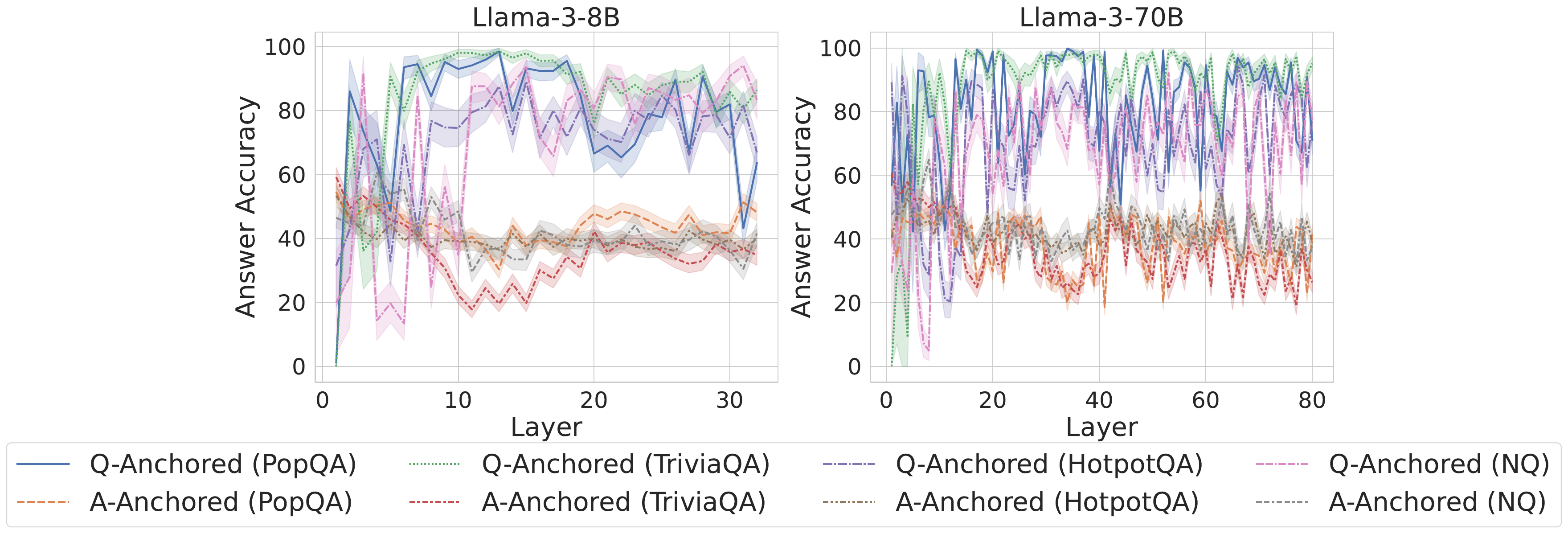}

    \vspace{4ex}

    \includegraphics[width=\textwidth]{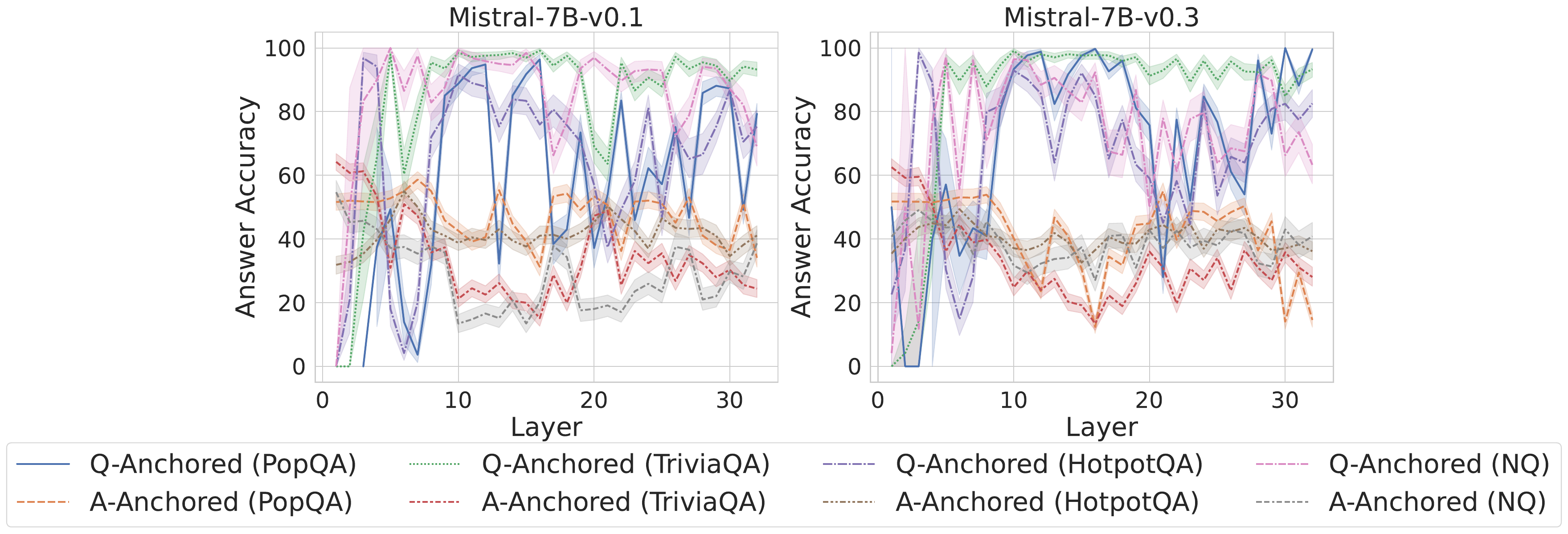}

    \vspace{2ex}

    \captionof{figure}{Comparisons of answer accuracy between pathways, probing attention activations of the final token.}
    \label{fig:appendix_answer_acc_base_attnact_-1}
\end{minipage}

\begin{figure*}[!htb]
\centering
\includegraphics[width=\textwidth]{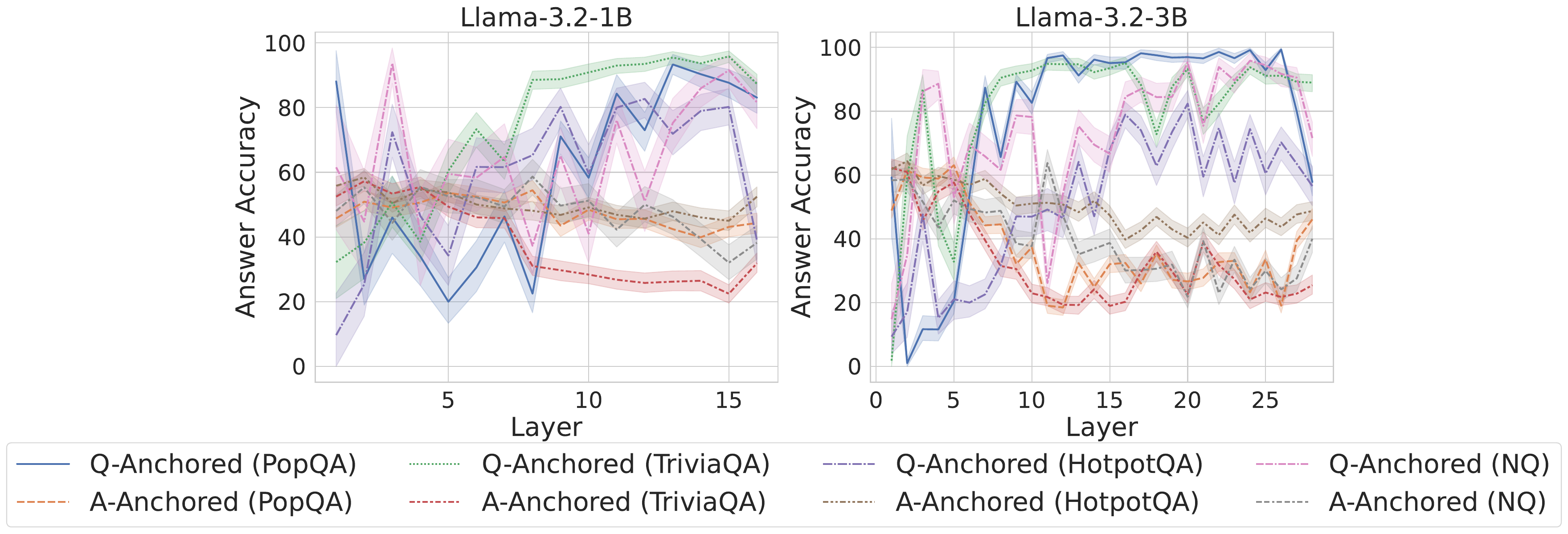}
\\[5ex]
\includegraphics[width=\textwidth]{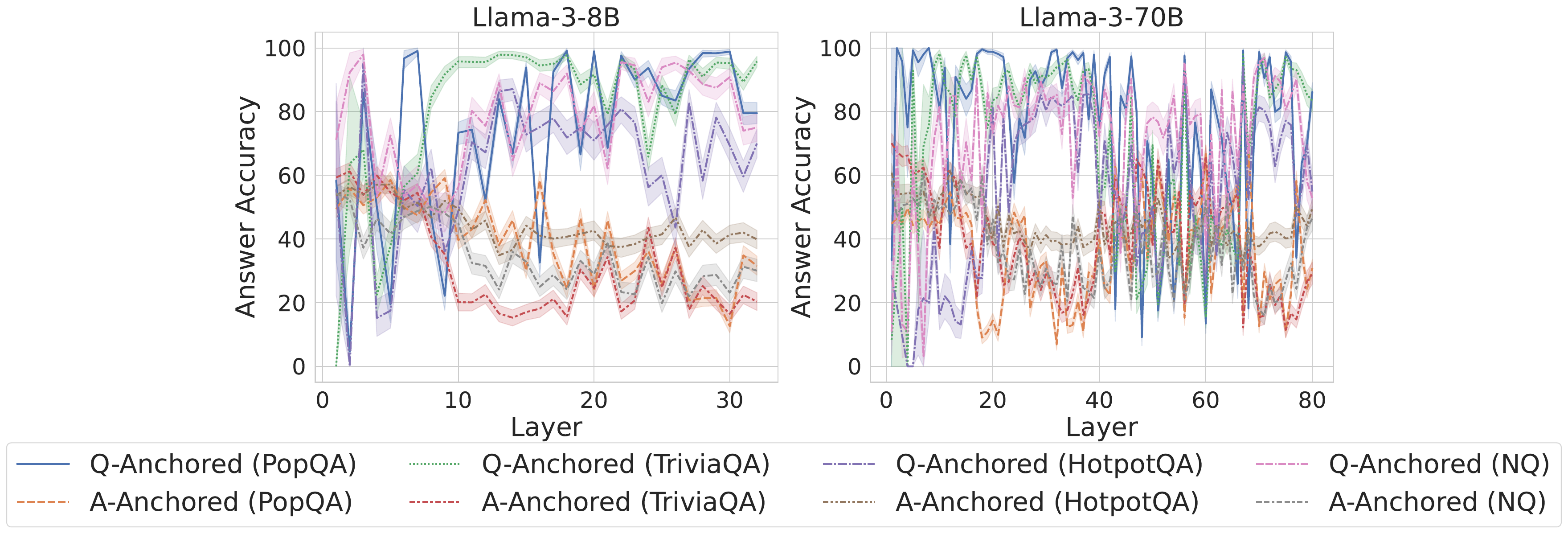}
\\[5ex]
\includegraphics[width=\textwidth]{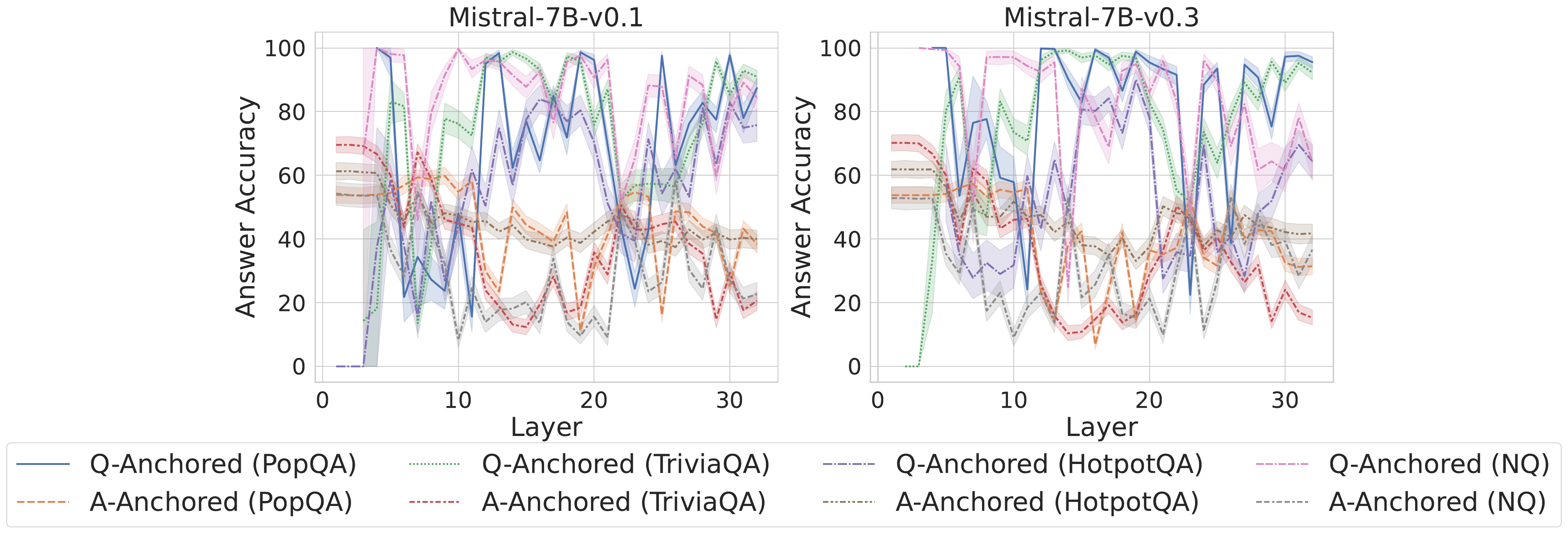}
\caption{Comparisons of answer accuracy between pathways, probing attention activations of the token immediately preceding the exact answer tokens.}
\label{fig:appendix_answer_acc_base_attnact_beforefirst}
\end{figure*}

\begin{figure*}[!htb]
\centering
\includegraphics[width=\textwidth]{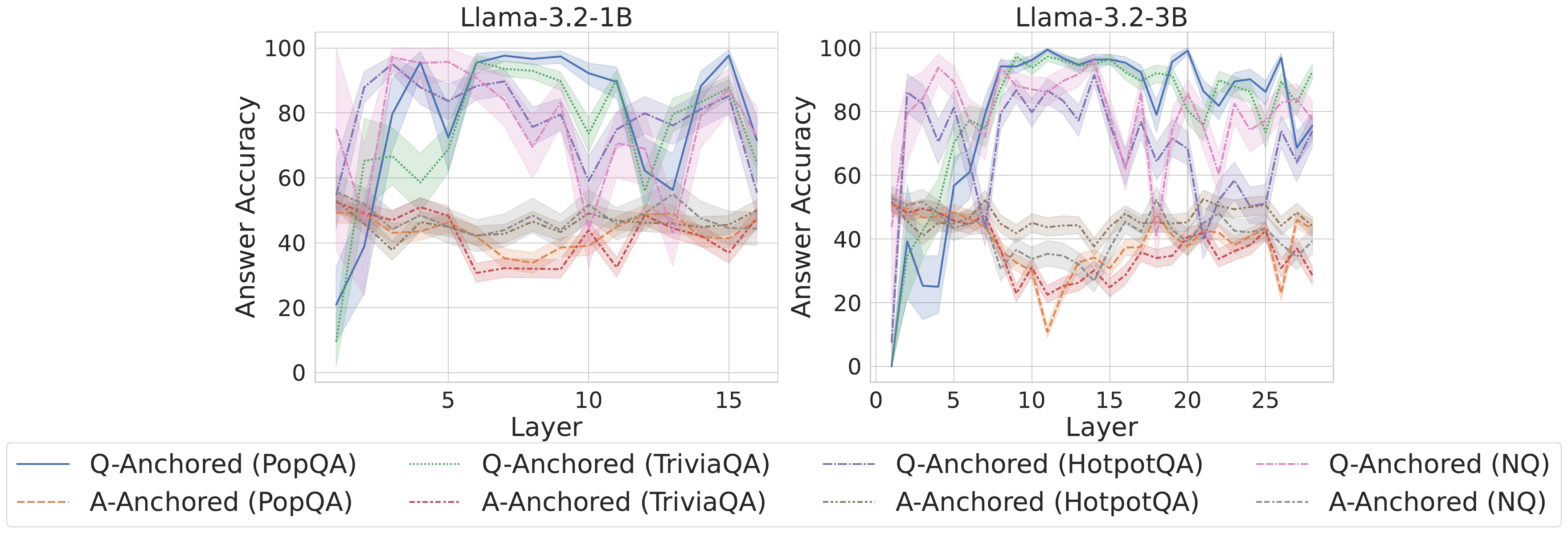}
\\[5ex]
\includegraphics[width=\textwidth]{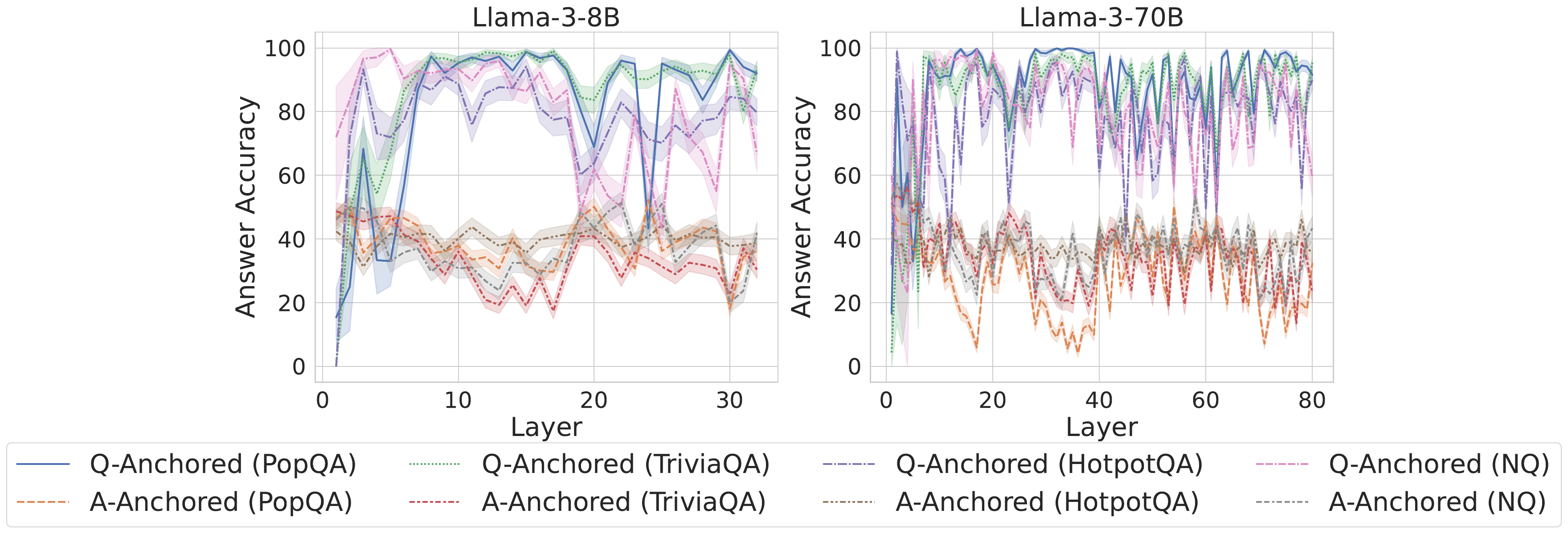}
\\[5ex]
\includegraphics[width=\textwidth]{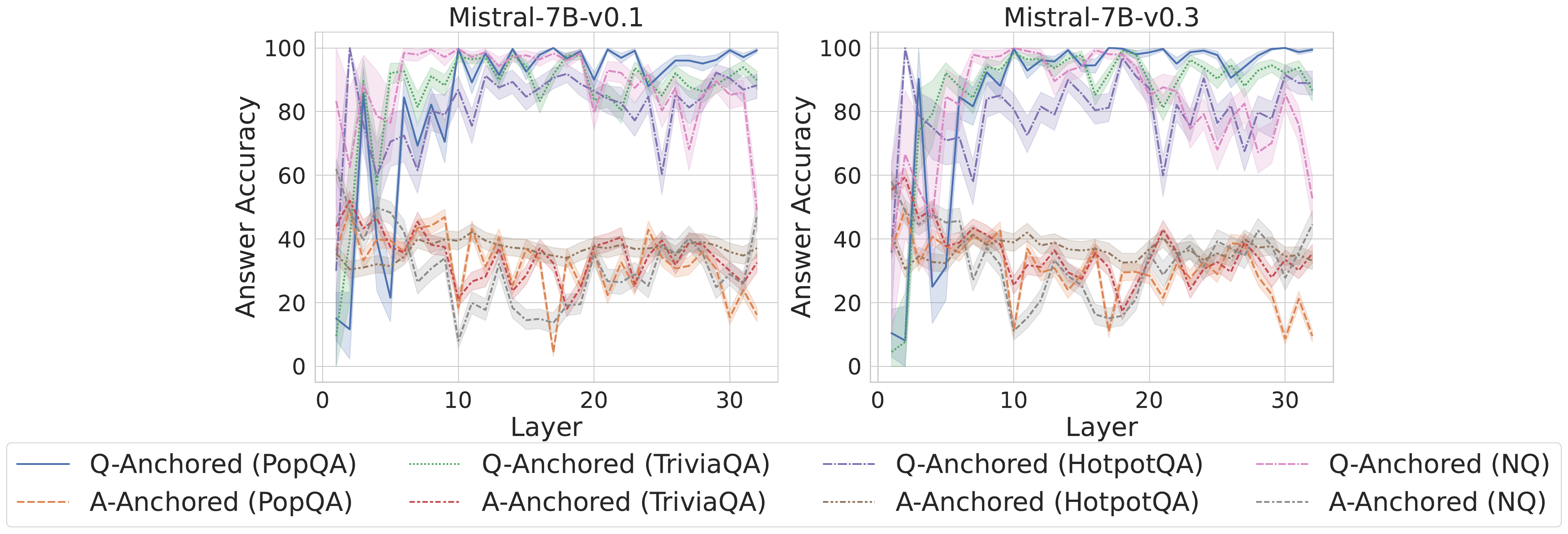}
\caption{Comparisons of answer accuracy between pathways, probing attention activations of the last exact answer token.}
\label{fig:appendix_answer_acc_base_attnact_exactans}
\end{figure*}

\begin{figure*}[!htb]
\centering
\includegraphics[width=\textwidth]{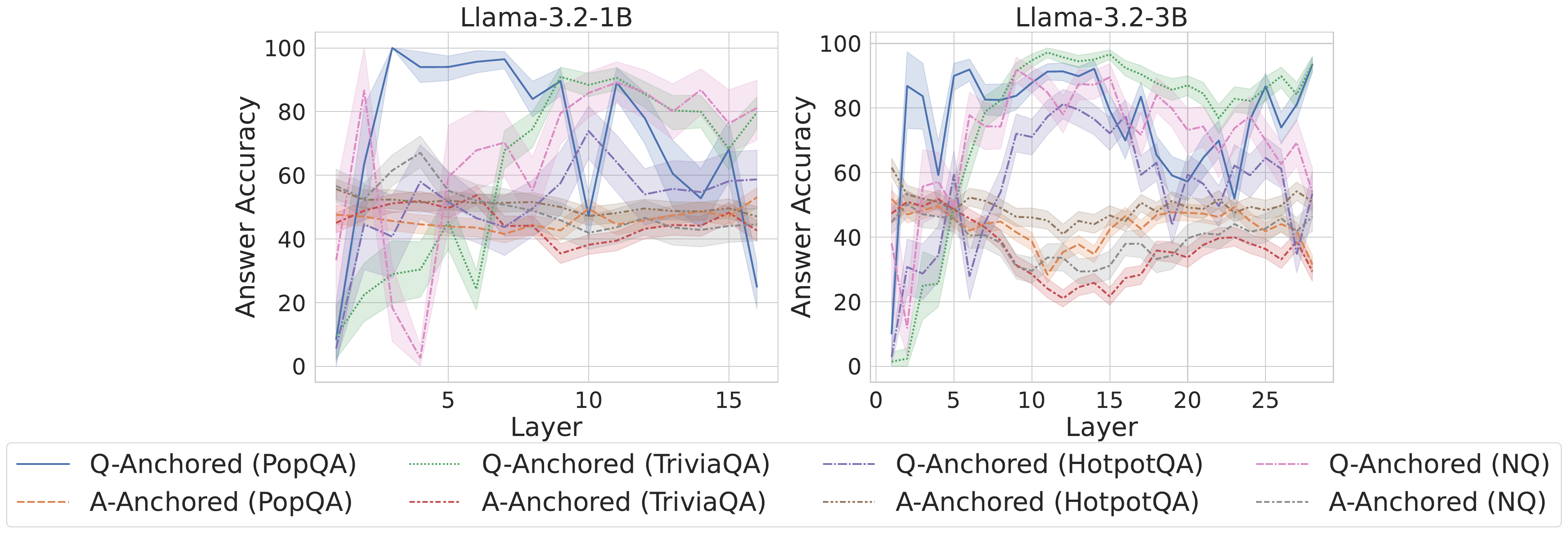}
\\[5ex]
\includegraphics[width=\textwidth]{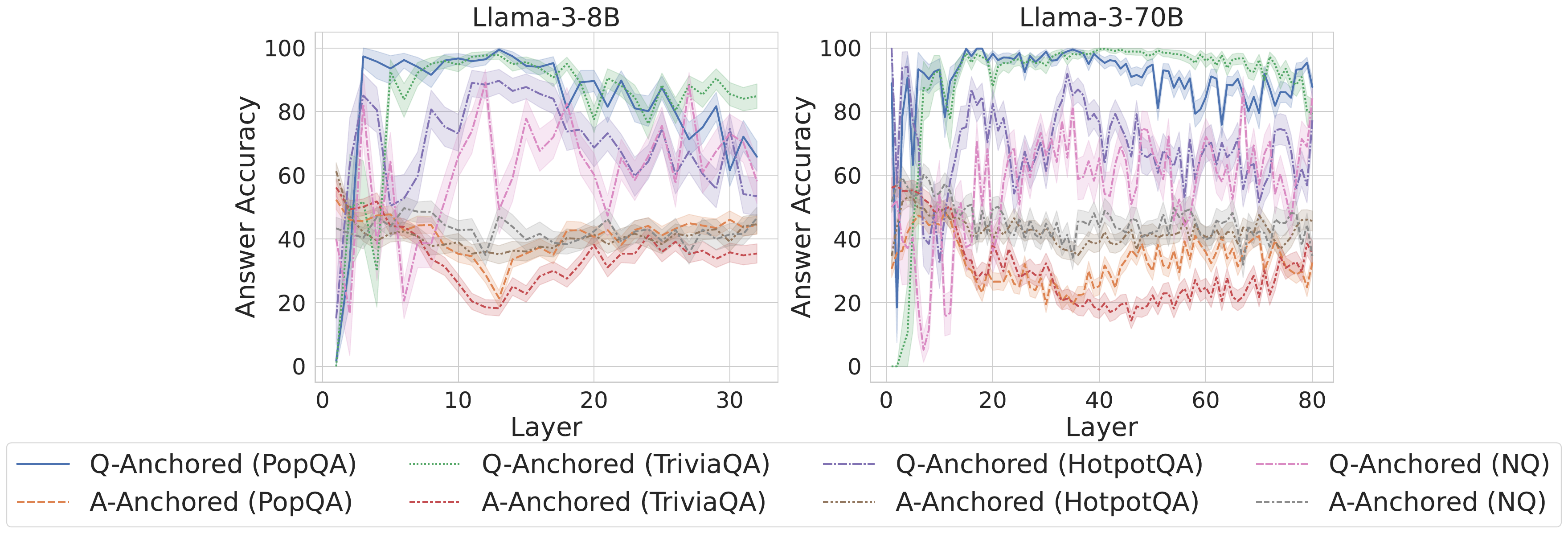}
\\[5ex]
\includegraphics[width=\textwidth]{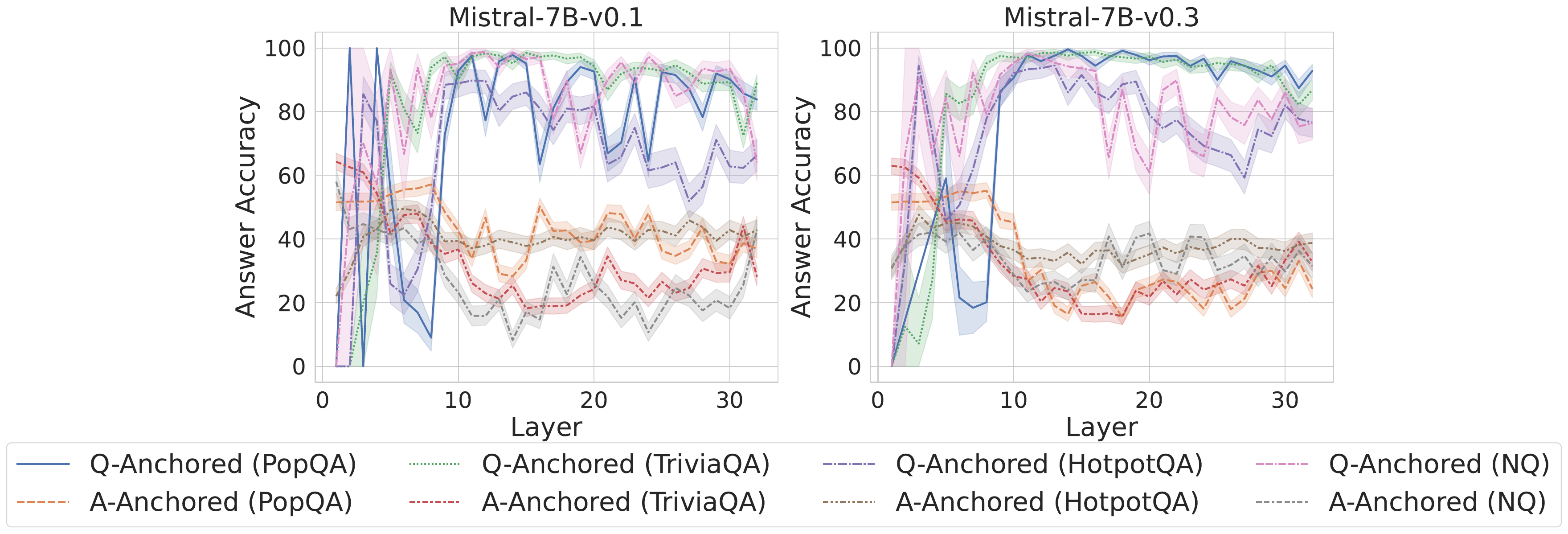}
\caption{Comparisons of answer accuracy between pathways, probing mlp activations of the final token.}
\label{fig:appendix_answer_acc_base_mlpact_-1}
\end{figure*}

\begin{figure*}[!htb]
\centering
\includegraphics[width=\textwidth]{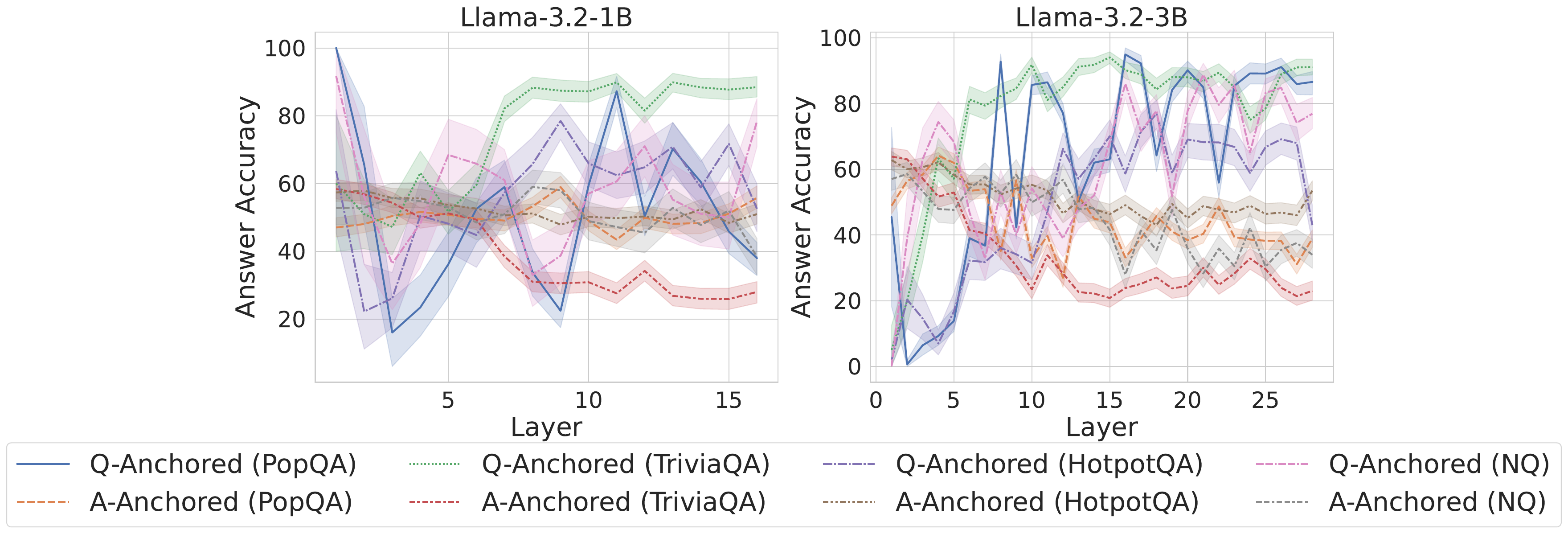}
\\[5ex]
\includegraphics[width=\textwidth]{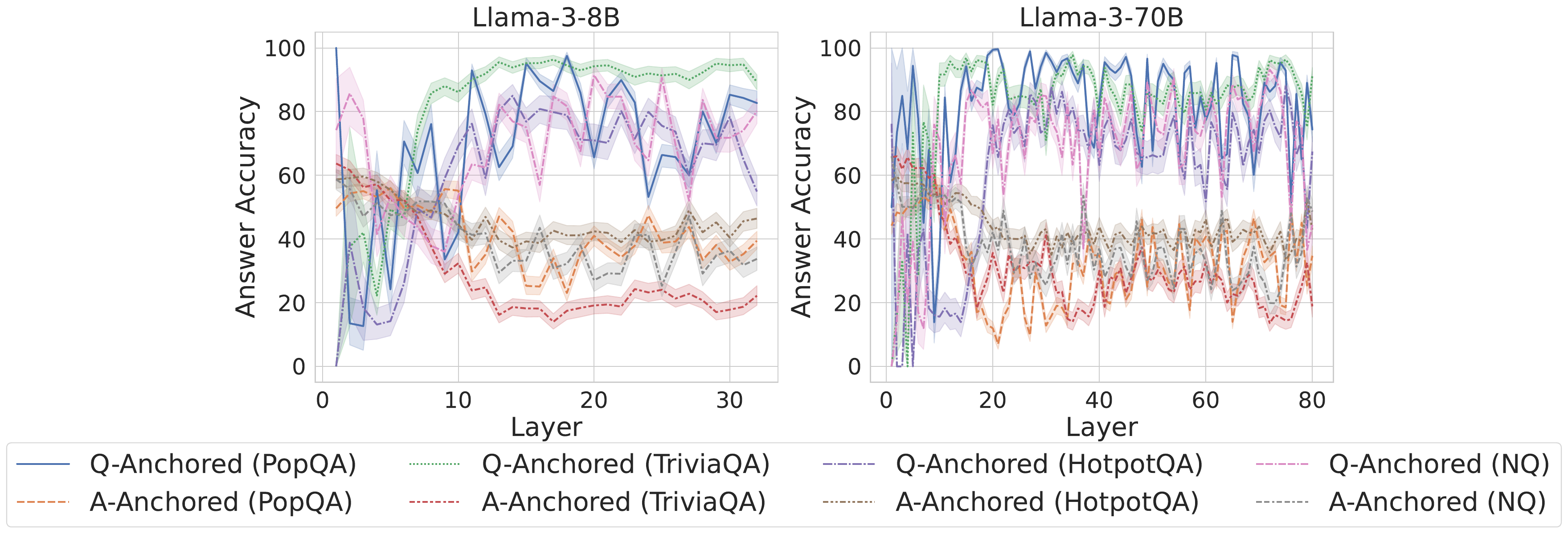}
\\[5ex]
\includegraphics[width=\textwidth]{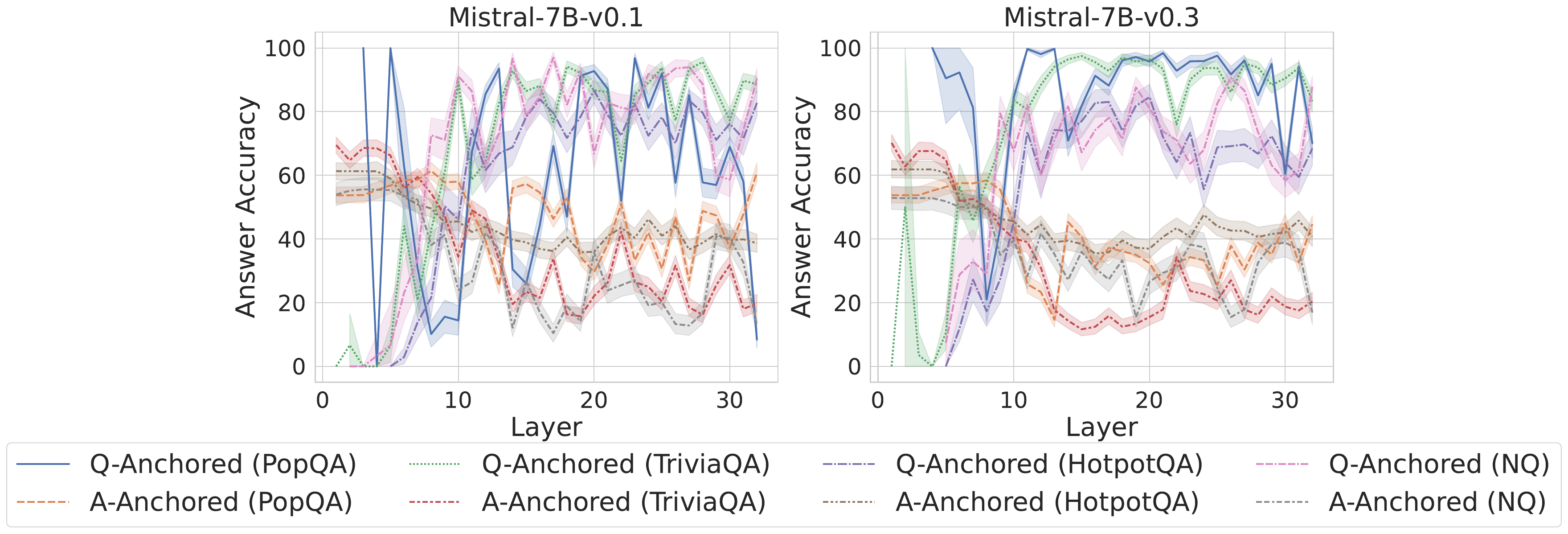}
\caption{Comparisons of answer accuracy between pathways, probing mlp activations of the token immediately preceding the exact answer tokens.}
\label{fig:appendix_answer_acc_base_mlpact_beforefirst}
\end{figure*}

\begin{figure*}[!htb]
\centering
\includegraphics[width=\textwidth]{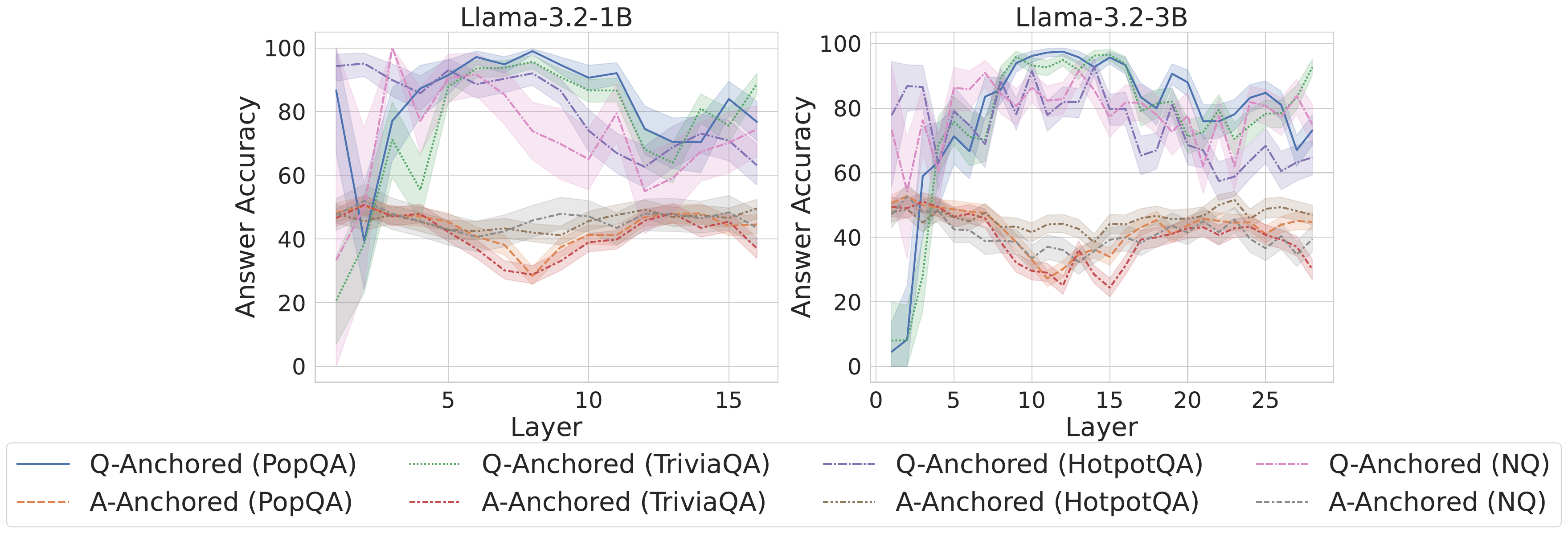}
\\[5ex]
\includegraphics[width=\textwidth]{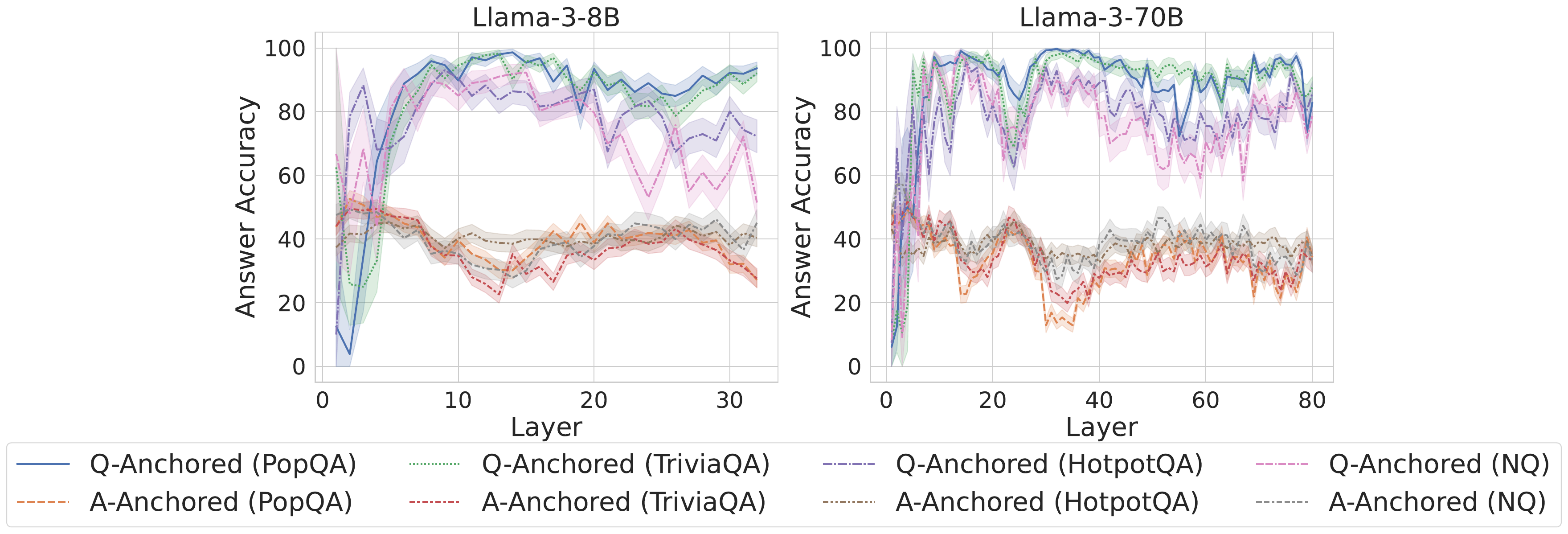}
\\[5ex]
\includegraphics[width=\textwidth]{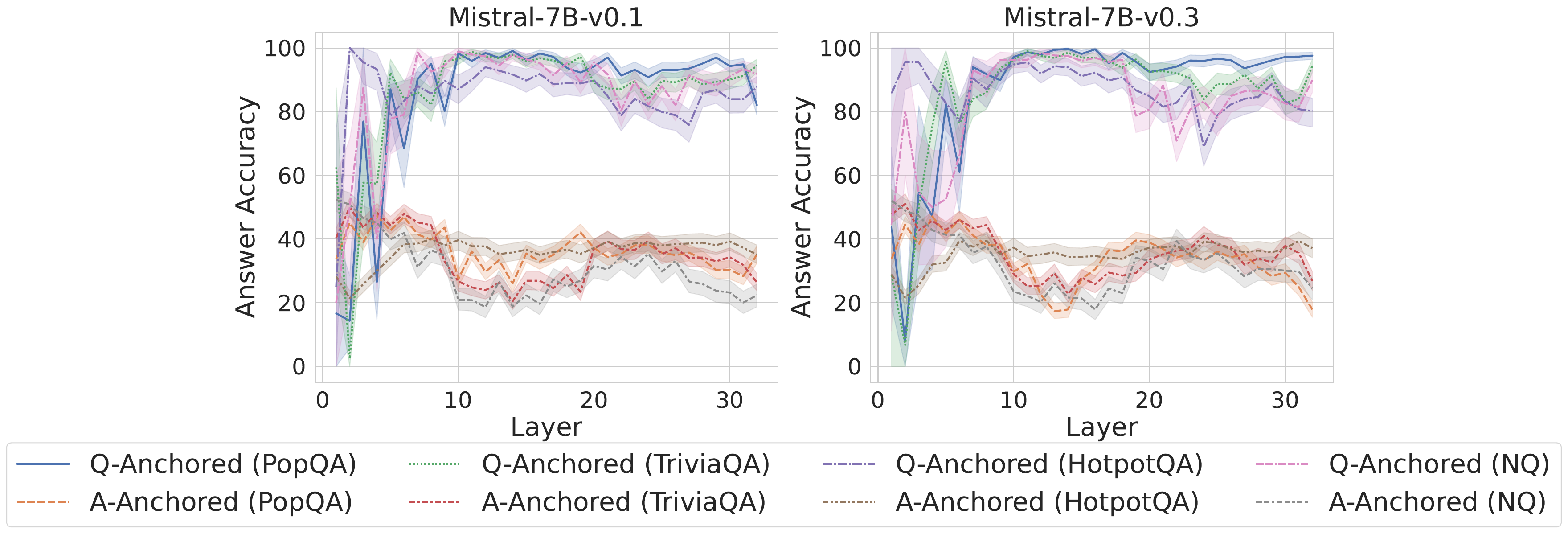}
\caption{Comparisons of answer accuracy between pathways, probing mlp activations of the last exact answer token.}
\label{fig:appendix_answer_acc_base_mlpact_exactans}
\end{figure*}

\clearpage
\section{I-Don't-Know Rate}
\label{sec:appendix_idk}

\begin{minipage}{\textwidth}
    \centering

    \includegraphics[width=\textwidth]{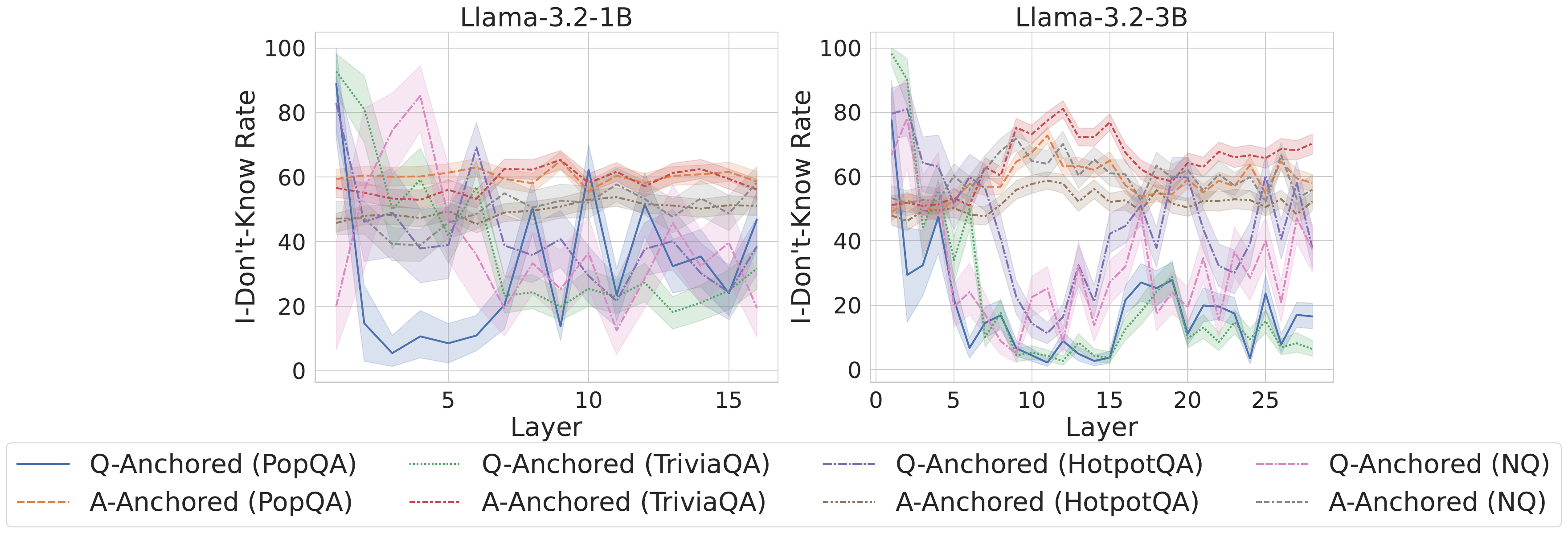}

    \vspace{4ex}

    \includegraphics[width=\textwidth]{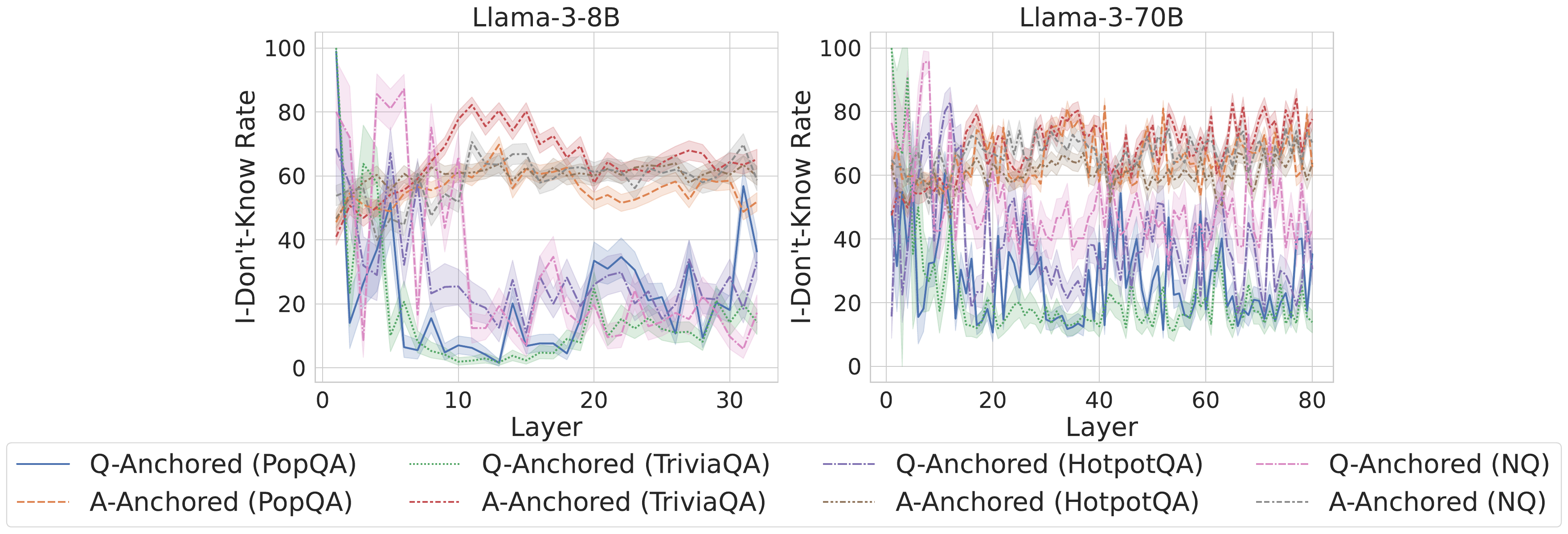}

    \vspace{4ex}

    \includegraphics[width=\textwidth]{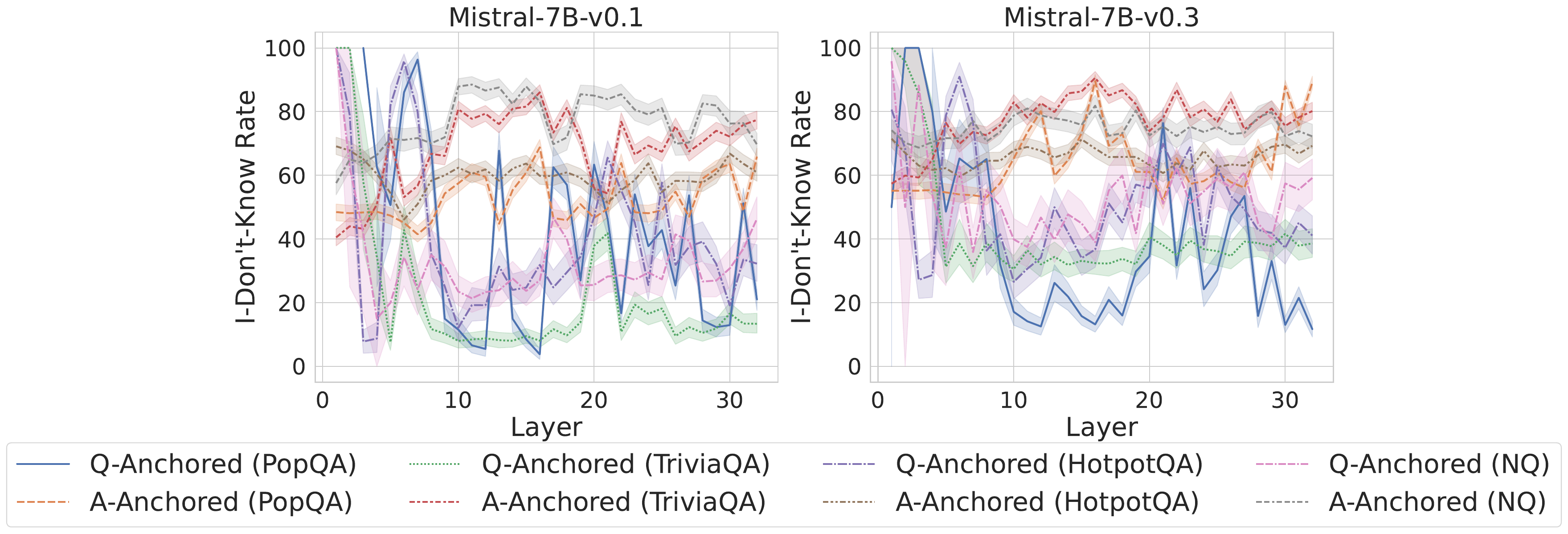}

    \vspace{2ex}

    \captionof{figure}{Comparisons of i-don't-know rate between pathways, probing attention activations of the final token.}
    \label{fig:appendix_idk_base_attnact_-1}
\end{minipage}

\begin{figure*}[!htb]
\centering
\includegraphics[width=\textwidth]{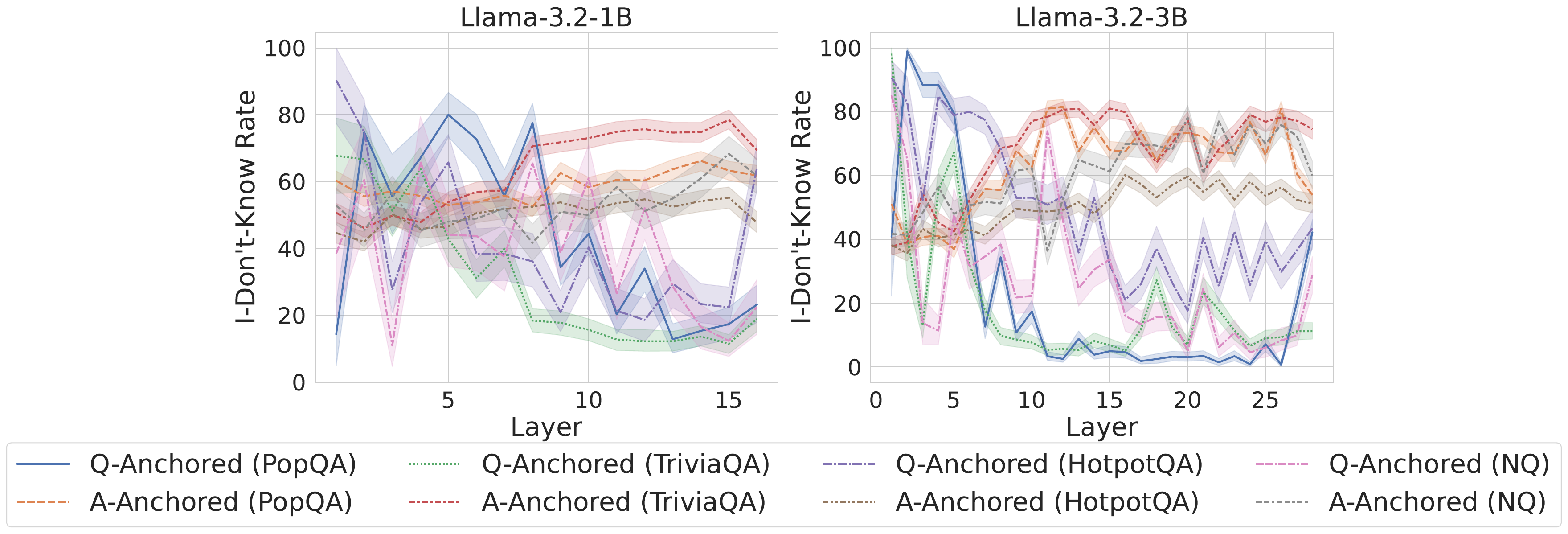}
\\[5ex]
\includegraphics[width=\textwidth]{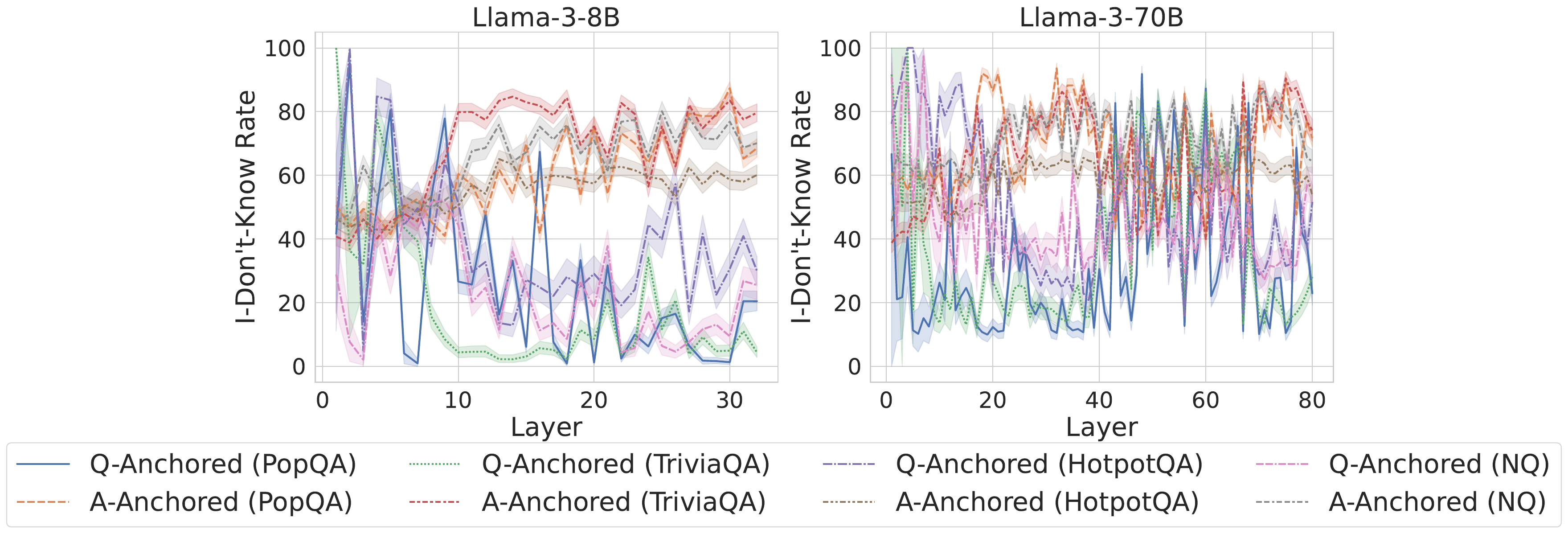}
\\[5ex]
\includegraphics[width=\textwidth]{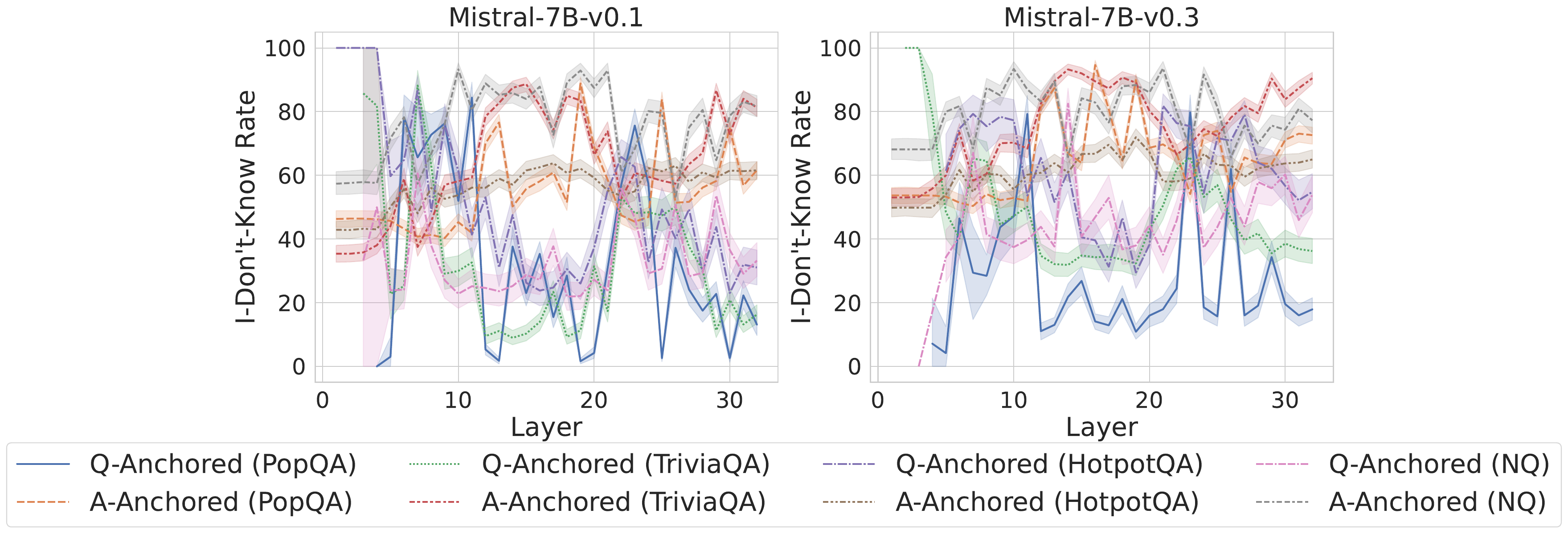}
\caption{Comparisons of i-don't-know rate between pathways, probing attention activations of the token immediately preceding the exact answer tokens.}
\label{fig:appendix_idk_base_attnact_beforefirst}
\end{figure*}

\begin{figure*}[!htb]
\centering
\includegraphics[width=\textwidth]{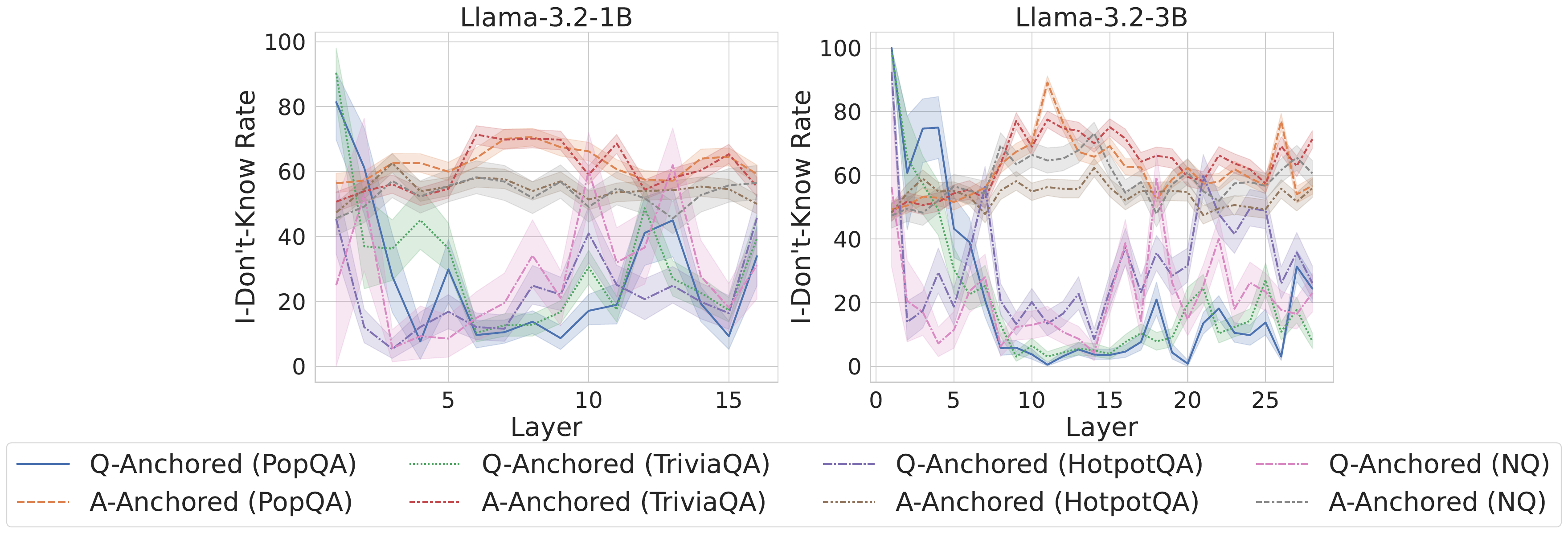}
\\[5ex]
\includegraphics[width=\textwidth]{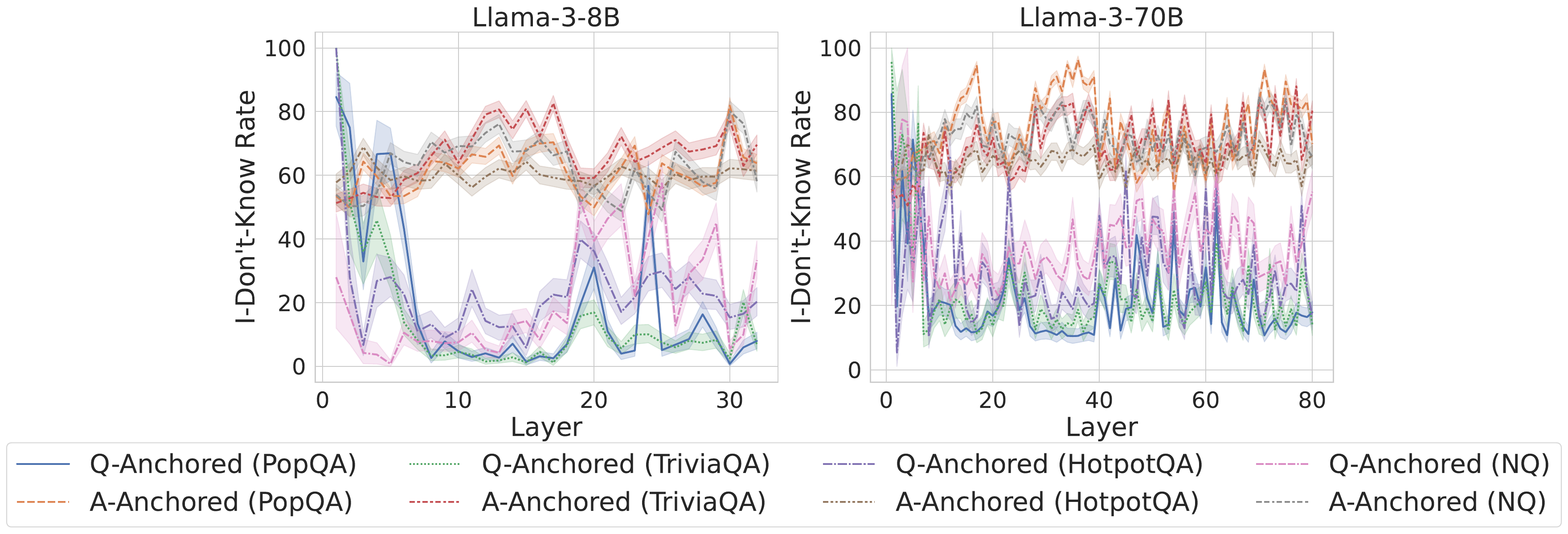}
\\[5ex]
\includegraphics[width=\textwidth]{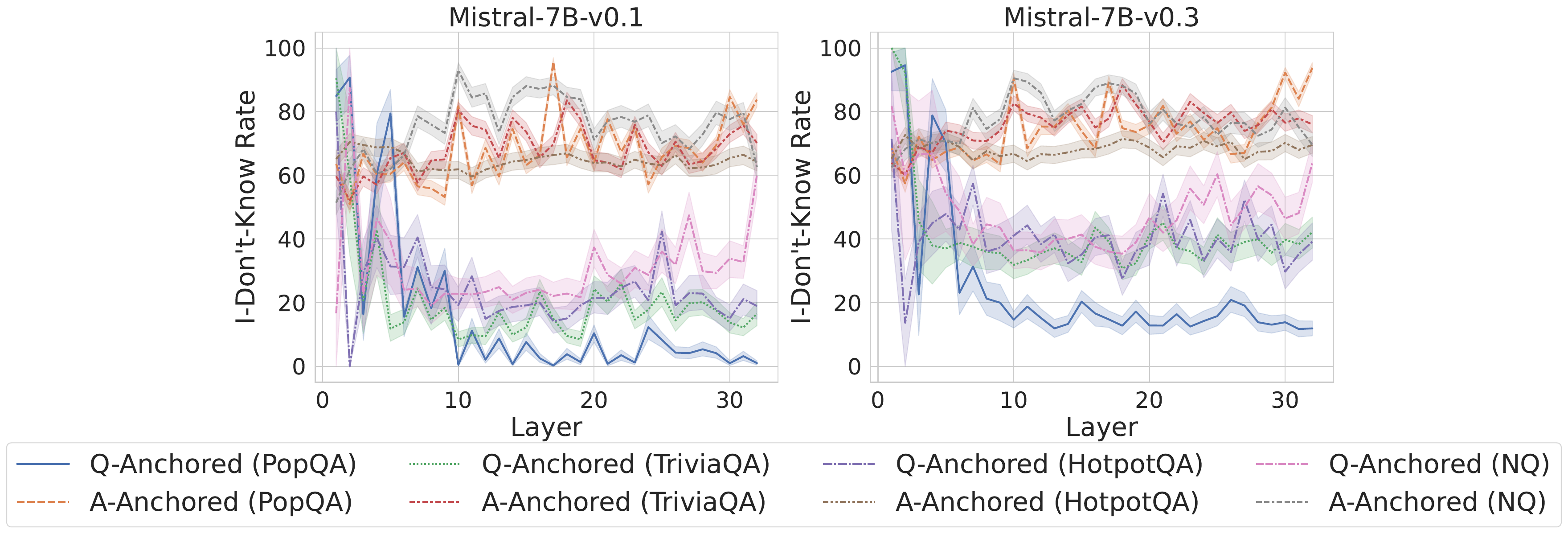}
\caption{Comparisons of i-don't-know rate between pathways, probing attention activations of the last exact answer token.}
\label{fig:appendix_idk_base_attnact_exactans}
\end{figure*}

\begin{figure*}[!htb]
\centering
\includegraphics[width=\textwidth]{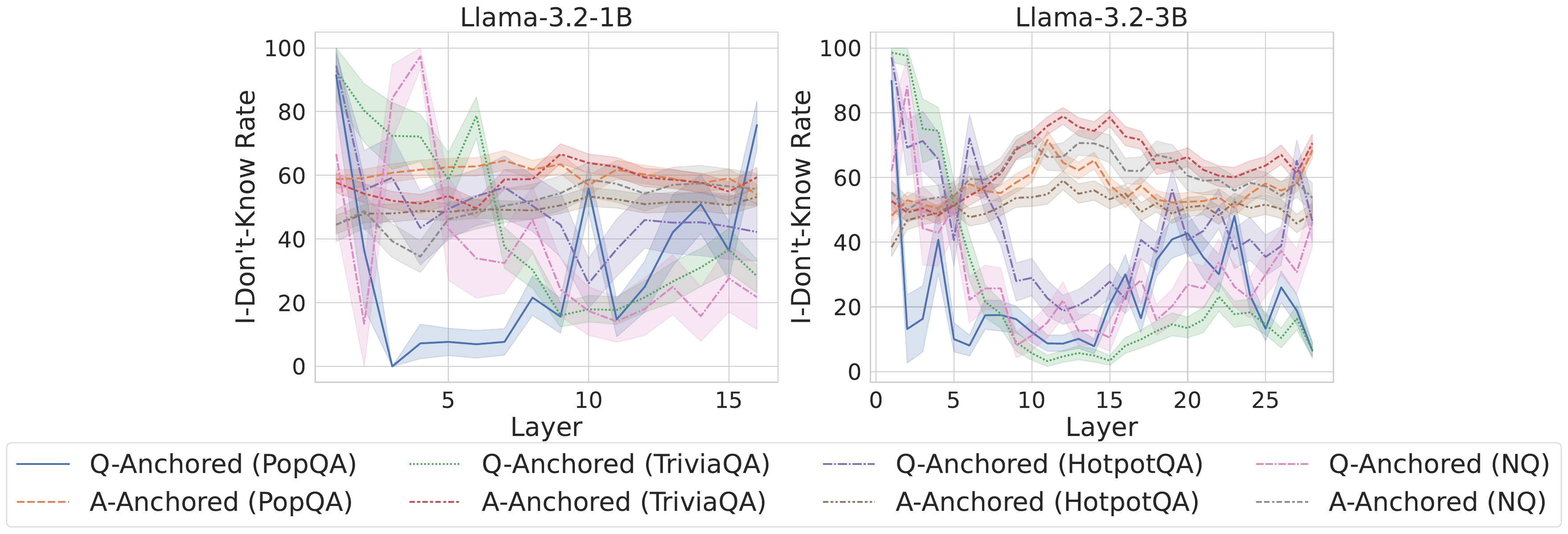}
\\[5ex]
\includegraphics[width=\textwidth]{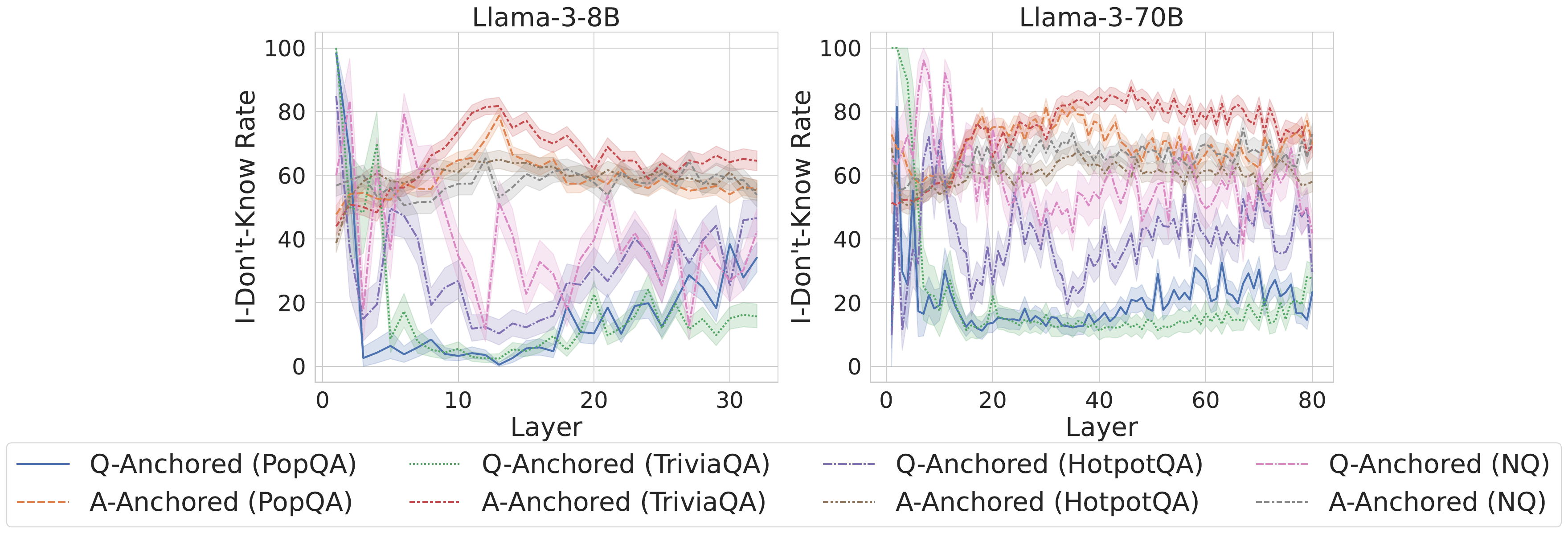}
\\[5ex]
\includegraphics[width=\textwidth]{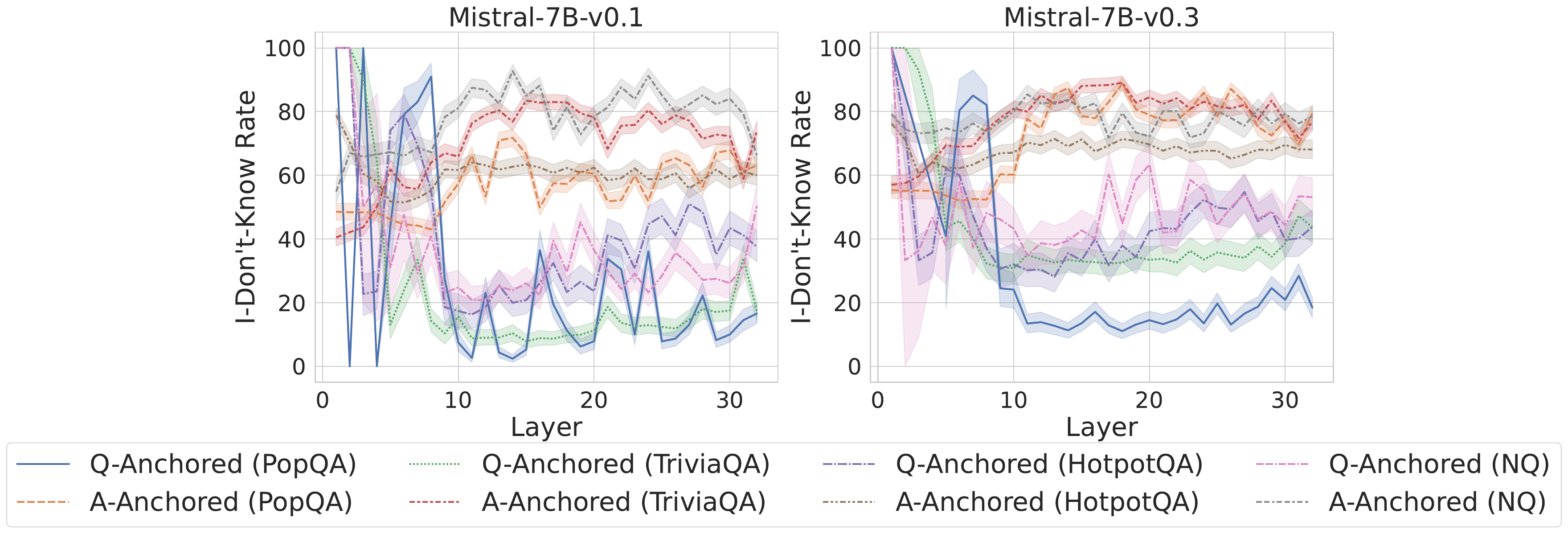}
\caption{Comparisons of i-don't-know rate between pathways, probing mlp activations of the final token.}
\label{fig:appendix_idk_base_mlpact_-1}
\end{figure*}

\begin{figure*}[!htb]
\centering
\includegraphics[width=\textwidth]{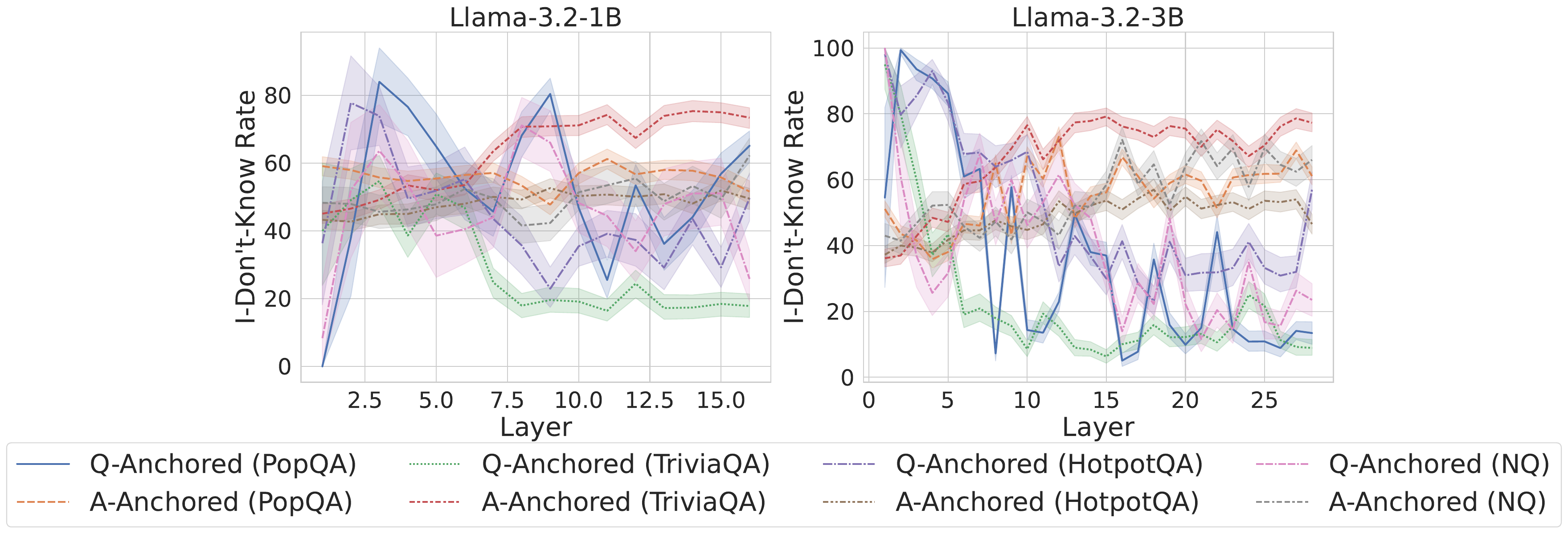}
\\[5ex]
\includegraphics[width=\textwidth]{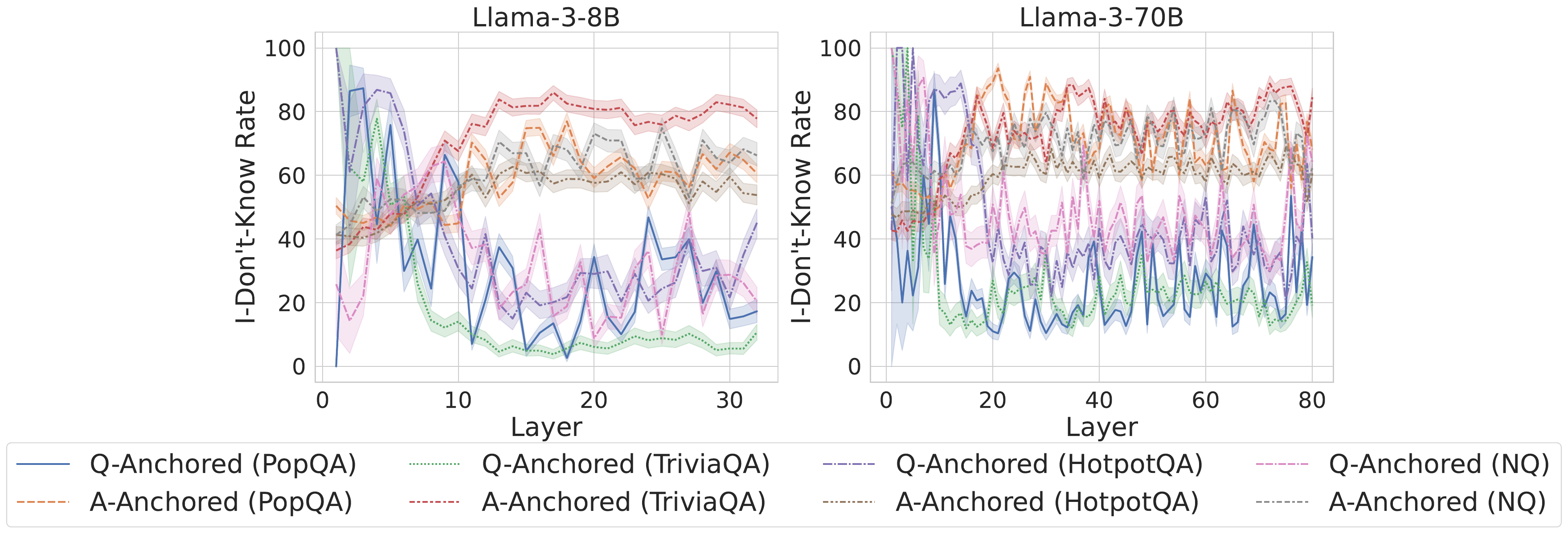}
\\[5ex]
\includegraphics[width=\textwidth]{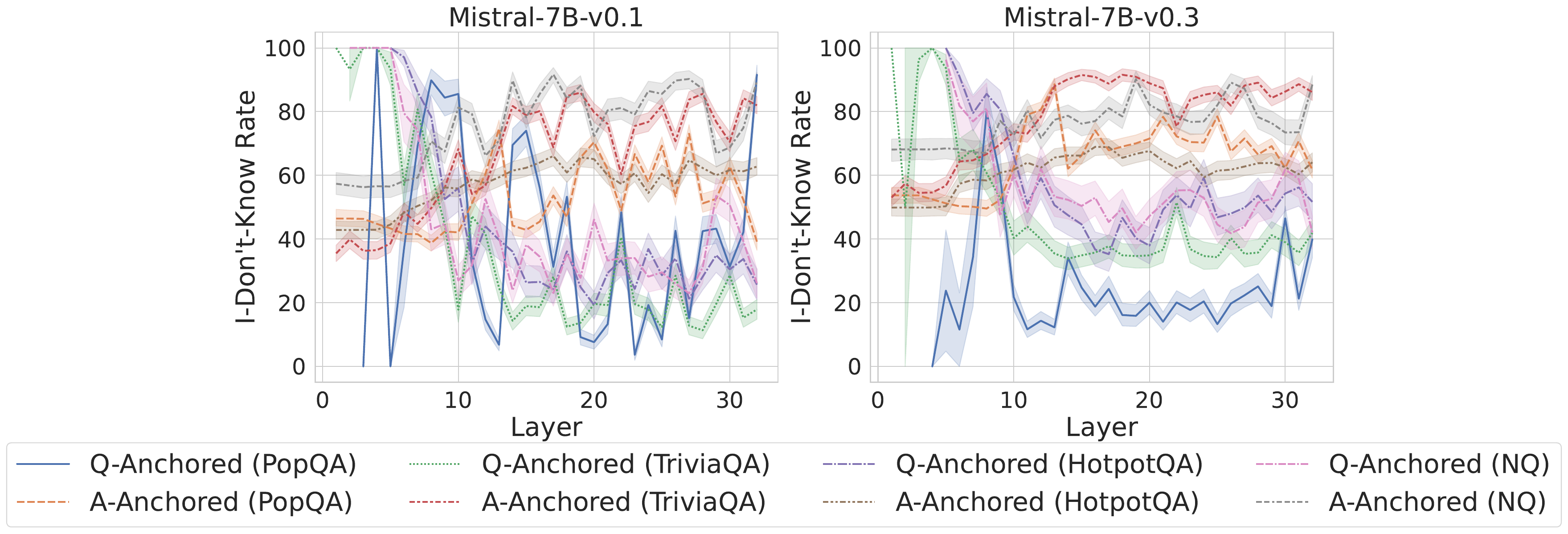}
\caption{Comparisons of i-don't-know rate between pathways, probing mlp activations of the token immediately preceding the exact answer tokens.}
\label{fig:appendix_idk_base_mlpact_beforefirst}
\end{figure*}

\begin{figure*}[!htb]
\centering
\includegraphics[width=\textwidth]{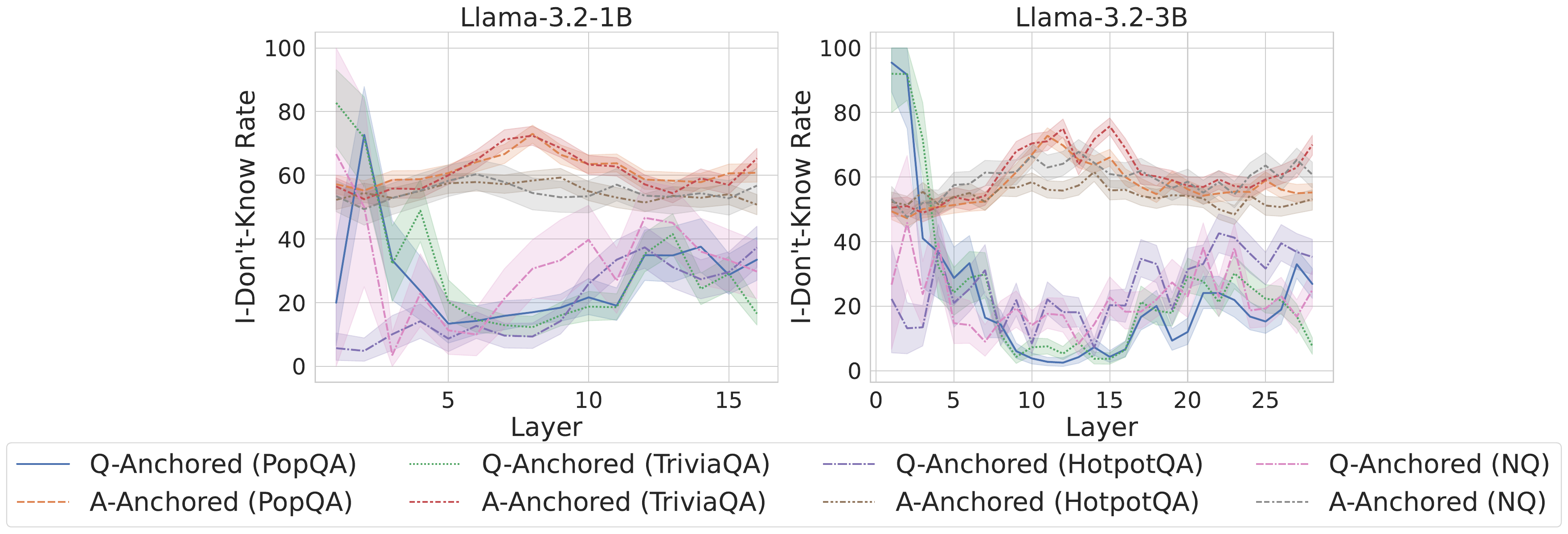}
\\[5ex]
\includegraphics[width=\textwidth]{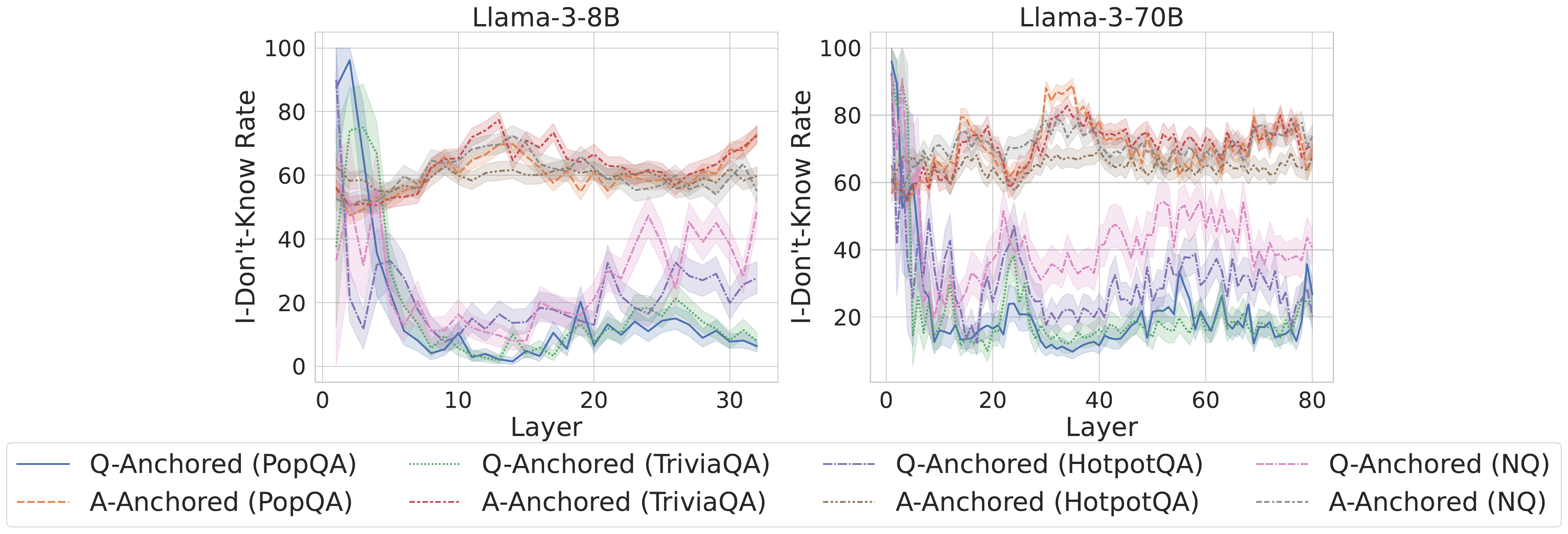}
\\[5ex]
\includegraphics[width=\textwidth]{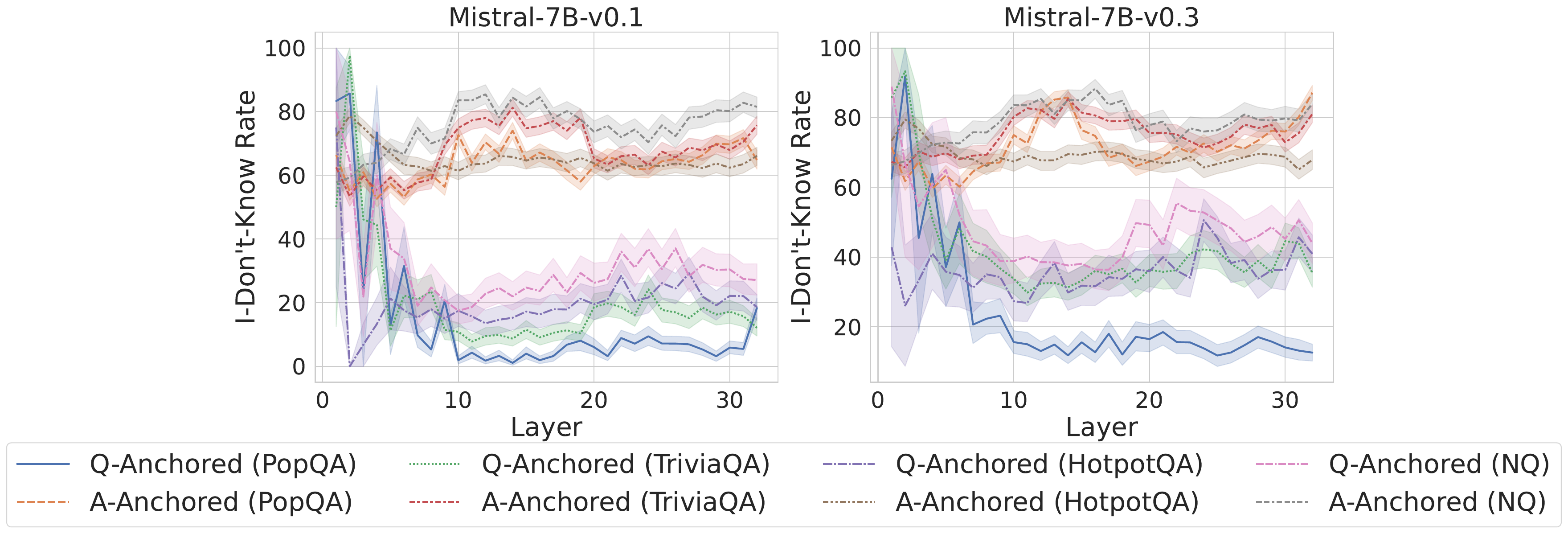}
\caption{Comparisons of i-don't-know rate between pathways, probing mlp activations of the last exact answer token.}
\label{fig:appendix_idk_base_mlpact_exactans}
\end{figure*}

\clearpage

\section{Pathway-Aware Detection}
\label{sec:appendix_applications}

\begin{table*}[!htb]
\centering
\begin{adjustbox}{valign=c, width=0.97\textwidth}
\begin{tabular}{lcccccccc}
\toprule
\multirow{2}{*}{Method} & \multicolumn{4}{c}{LLama-3.2-1B} & \multicolumn{4}{c}{LLama-3.2-3B} \\
\cmidrule(lr){2-5} \cmidrule(lr){6-9}
 & PopQA & TriviaQA & HotpotQA  & NQ & PopQA & TriviaQA & HotpotQA  & NQ\\
\midrule
P(True) & 60.00 & 49.65 & 43.34 & 52.83 & 54.58 & 51.76 & 47.73 & 53.78 \\
Logits-mean & 74.89 & 60.24 & 60.18 & 49.92 & 73.47 & 63.46 & 60.35 & 54.89 \\
Logits-max & 58.56 & 52.37 & 52.29 & 46.19 & 56.03 & 54.33 & 48.65 & 48.88 \\
Logits-min & 78.66 & 62.37 & 67.14 & 51.20 & 80.92 & 69.60 & 71.11 & 58.24 \\
Scores-mean & 72.91 & 61.13 & 62.16 & 64.67 & 67.99 & 61.96 & 64.91 & 61.71 \\
Scores-max & 69.33 & 59.74 & 61.29 & 64.08 & 63.34 & 61.92 & 61.09 & 57.56 \\
Scores-min & 64.84 & 55.93 & 59.28 & 55.81 & 61.51 & 56.76 & 63.95 & 57.43 \\
Probing Baseline & 94.25 & 77.17 & 90.25 & 74.83 & 90.96 & 76.61 & 86.54 & 74.20 \\
\rowcolor{mygray} MoP-RandomGate & 83.69 & 69.20 & 84.11 & 68.76 & 79.69 & 72.38 & 75.13 & 67.11 \\
\rowcolor{mygray} MoP-VanillaExperts & 93.86 & 78.63 & 90.91 & 75.73 & 90.98 & 77.68 & 86.41 & 75.30 \\
\rowcolor{mygray} MoP & 95.85 & 80.07 & 91.51 & 79.19 & 92.74 & 78.72 & 88.16 & 78.14 \\
\rowcolor{mygray} PR & \textbf{96.18} & \textbf{84.22} & \textbf{92.80} & \textbf{86.45} & \textbf{95.70} & \textbf{80.66} & \textbf{90.66} & \textbf{81.91} \\
\bottomrule
\end{tabular}
\end{adjustbox}
\caption{Comparison of hallucination detection performance (AUC) on LLama-3.2-1B and LLama-3.2-3B.}
\label{tab:appendix: Pathway-Aware Hallucination Detection on LLama-3.2-1B and LLama-3.2-3B}
\end{table*}

\begin{table*}[!htb]
\centering
\begin{adjustbox}{valign=c, width=0.97\textwidth}
\begin{tabular}{lcccccccc}
\toprule
\multirow{2}{*}{Method} & \multicolumn{4}{c}{LLama-3-8B} & \multicolumn{4}{c}{LLama-3-70B} \\
\cmidrule(lr){2-5} \cmidrule(lr){6-9}
 & PopQA & TriviaQA & HotpotQA  & NQ & PopQA & TriviaQA & HotpotQA  & NQ\\
\midrule
P(True) & 55.85 & 49.92 & 52.14 & 53.27 & 54.83 & 50.96 & 49.39 & 51.18 \\
Logits-mean & 74.52 & 60.39 & 51.94 & 52.63 & 67.81 & 52.40 & 50.45 & 48.28 \\
Logits-max & 58.08 & 52.20 & 46.40 & 47.89 & 56.21 & 48.16 & 43.42 & 45.33 \\
Logits-min & 85.36 & 70.89 & 61.28 & 56.50 & 79.96 & 61.53 & 62.63 & 52.16 \\
Scores-mean & 62.87 & 62.09 & 62.06 & 60.32 & 56.81 & 60.70 & 60.91 & 58.05 \\
Scores-max & 56.62 & 60.24 & 59.85 & 56.06 & 55.15 & 59.60 & 57.32 & 51.93 \\
Scores-min & 60.99 & 58.27 & 60.33 & 57.68 & 58.77 & 58.22 & 64.06 & 58.05 \\
Probing Baseline & 88.71 & 77.58 & 82.23 & 70.20 & 86.88 & 81.59 & 84.45 & 74.39 \\
\rowcolor{mygray} MoP-RandomGate & 75.52 & 69.17 & 79.88 & 66.56 & 67.96 & 70.56 & 72.16 & 66.28\\
\rowcolor{mygray} MoP-VanillaExperts & 89.11 & 78.73 & 84.57 & 71.21 & 86.04 & 82.47 & 82.48 & 73.85  \\
\rowcolor{mygray} MoP & 92.11 & 81.18 & 85.45 & 74.64 & 88.54 & 84.12 & 86.65 & 76.12 \\
\rowcolor{mygray} PR & \textbf{94.01} & \textbf{83.13} & \textbf{87.81} & \textbf{79.10} & \textbf{90.08} & \textbf{84.21} & \textbf{87.69} & \textbf{78.24} \\
\bottomrule
\end{tabular}
\end{adjustbox}
\caption{Comparison of hallucination detection performance (AUC) on LLama-3-8B and LLama-3-70B.}
\label{tab:appendix: Pathway-Aware Hallucination Detection on LLama-3-8B and LLama-3-70B}
\end{table*}

\begin{table*}[!htb]
\centering
\begin{adjustbox}{valign=c, width=0.97\textwidth}
\begin{tabular}{lcccccccc}
\toprule
\multirow{2}{*}{Method} & \multicolumn{4}{c}{Mistral-7B-v0.1} & \multicolumn{4}{c}{Mistral-7B-v0.3} \\
\cmidrule(lr){2-5} \cmidrule(lr){6-9}
 & PopQA & TriviaQA & HotpotQA  & NQ & PopQA & TriviaQA & HotpotQA  & NQ\\
\midrule
P(True) & 48.78 & 50.43 & 51.94 & 55.52 & 45.49 & 47.61 & 57.87 & 52.79 \\
Logits-mean & 69.09 & 64.95 & 54.47 & 59.41 & 69.52 & 66.76 & 55.45 & 57.88 \\
Logits-max & 54.37 & 54.76 & 46.74 & 56.45 & 54.34 & 55.24 & 48.39 & 54.37 \\
Logits-min & 86.02 & 76.56 & 68.06 & 53.73 & 87.05 & 77.33 & 68.08 & 54.40 \\
Scores-mean & 59.00 & 59.61 & 64.18 & 57.60 & 58.84 & 60.22 & 63.28 & 60.05 \\
Scores-max & 51.71 & 56.58 & 63.29 & 55.82 & 53.00 & 55.55 & 63.13 & 57.73 \\
Scores-min & 60.00 & 57.48 & 61.17 & 48.51 & 60.59 & 57.84 & 59.85 & 50.76 \\
Probing Baseline & 89.61 & 78.43 & 83.76 & 74.10 & 87.39 & 81.74 & 83.19 & 73.60 \\
\rowcolor{mygray} MoP-RandomGate & 80.50 & 68.27 & 74.51 & 68.05 & 79.81 & 70.88 & 72.23 & 61.19 \\
\rowcolor{mygray} MoP-VanillaExperts & 89.82 & 79.51 & 83.54 & 74.78 & 88.53 & 80.93 & 82.93 & 73.77 \\
\rowcolor{mygray} MoP & 92.44 & 84.03 & 84.63 & 76.38 & 91.66 & 83.57 & 85.82 & 76.87 \\
\rowcolor{mygray} PR & \textbf{94.72} & \textbf{84.66} & \textbf{89.04} & \textbf{80.92} & \textbf{93.09} & \textbf{84.36} & \textbf{89.03} & \textbf{79.09} \\
\bottomrule
\end{tabular}
\end{adjustbox}
\caption{Comparison of hallucination detection performance (AUC) on Mistral-7B-v0.1 and Mistral-7B-v0.3.}
\label{tab:appendix: Pathway-Aware Hallucination Detection on Mistral-7B-v0.1 and Mistral-7B-v0.3}
\end{table*}

\twocolumn
\end{document}